# Enabling Ethical AI: A case study in using Ontological Context for Justified Agentic AI Decisions

Liam McGee *1*, James Harvey *1*, Lucy Cull *1*, Andreas Vermeulen *2*, Bart-Floris Visscher *2,* Malvika Sharan *3*

**Affiliations:**
1. Kaiasm Ltd, London, UK
2. Avantra, Richmond, UK
3. The Alan Turing Institute, London, UK

## Abstract

Agentic AI systems, software agents with autonomy, decision-making ability, and adaptability, are increasingly used to execute complex tasks on behalf of organisations. Most such systems rely on Large Language Models (LLMs), whose broad semantic capabilities enable powerful language processing but lack explicit, institution-specific grounding. In enterprises, data rarely comes with an inspectable semantic layer, and constructing one typically requires labour-intensive "data archaeology": cleaning, modelling, and curating knowledge into ontologies, taxonomies, and other formal structures. At the same time, explainability methods such as saliency maps expose an "interpretability gap": they highlight what the model attends to but not why, leaving decision processes opaque. In this preprint, we present a case study, developed by Kaiasm and Avantra AI through their work with The Turing Way Practitioners Hub, a forum developed under the InnovateUK BridgeAI program. This study presents a collaborative human-AI approach to building an inspectable semantic layer for Agentic AI. AI agents first propose candidate knowledge structures from diverse data sources; domain experts then validate, correct, and extend these structures, with their feedback used to improve subsequent models. Authors show how this process captures tacit institutional knowledge, improves response quality and efficiency, and mitigates institutional amnesia. We argue for a shift from post-hoc explanation to justifiable Agentic AI, where decisions are grounded in explicit, inspectable evidence and reasoning accessible to both experts and non-specialists.

## Introduction

Agentic AI systems are software systems that leverage Artificial Intelligence to exhibit some degree of autonomy, decision-making capabilities, and adaptability, allowing the systems to pursue complex goals and execute tasks on behalf of users with limited human supervision. Companies across all sectors are actively developing Agentic AI systems to improve their institutional operations. The broadly used technologies often rely on language processing solutions offered by Large Language Models (LLMs) [1].

LLMs are built with a generalised semantic context, deriving a broad understanding of language and relationships from their training data. This enables powerful pattern-matching for language processing, often without explicit instructions. However, this inherent generality poses a significant challenge for enterprise-level Agentic AI architectures. Enterprise data frequently lacks a crucial, inspectable semantic layer that provides specific institutional context, thus severely limiting Agentic AI's real-world applicability. Creating a semantic layer traditionally involves extensive manual "data archaeology" and curation: cleaning datasets, constructing formal knowledge structures like ontologies and taxonomies, and organising information for better availability. Adding to this challenge are the current AI explainability methods, such as saliency maps [2], which exhibit an "interpretability gap"; they can highlight the 'salient' features of an input that a model focuses on, but fail to explain the reasoning behind it. This presents the urgent need for Agentic AI decisions to be inherently justifiable, fully inspectable, and easily accessible to both experts and non-specialists.

The sheer effort and insufficient tooling for this process hinder an organisation's ability to effectively deploy Agentic AI. An improved solution lies in continuous human-AI collaboration. This involves AI agents generating initial knowledge structure candidates from various information sources, which human experts then validate and enrich. These human-derived corrections, in turn, train subsequent AI models, offering key benefits: 1) bringing tacit and contextual knowledge into a machine readable format; 2) substantially reducing model training cost and improving response quality at lower energy use [3]; 3) preventing knowledge loss (institutional amnesia); and 4) improving the capabilities of Agentic AI.

Kaiasm and Avantra AI teams collaborated to investigate this challenge through the enhancement and monitoring of inspectable Agentic AI systems by making knowledge orchestration accessible to their human operators. In their joint work, they tested whether the resultant semantic context improved accuracy, coherence and relevance of model responses. This also created an opportunity for the teams to explore the explainability of AI models, a challenge that currently remains unaddressed by existing approaches such as saliency maps, where an "interpretability gap of explainability methods relies on humans to decide what a given explanation might mean" [4].

This preprint describe a novel collaborative approach to explainability founded on "*justification*", ensuring agentic AI decisions can be justified in terms of evidence and reasoning in a way inspectable before and after a decision is enacted, whilst broadening people's access to formalised underlying knowledge foundations, without requiring highly specialised skill sets.

## A Hybrid Approach for Adaptability and Explainability

Neural networks, including those powering LLMs, excel at pattern recognition and learn from examples, but operate as "black boxes" with limited interpretability. Symbolic AI, conversely, excels at logical reasoning with explicit rules but struggles with unstructured data variability [5],[6],[7]. Neural networks and Symbolic AI are two complementary AI paradigms that can be combined to build neuro-symbolic architecture, creating systems that are both adaptive and explainable.

This architecture offers three essential benefits:

1. **Pattern recognition with reasoning:** Neural components identify patterns in enterprise data while symbolic components provide the logical framework to interpret these patterns meaningfully.
2. **Explainability by design:** Each decision maintains a traceable, justifiable path, a novel approach to addressing the "explainability crisis" plaguing many AI systems.
3. **Adaptability with guardrails:** The system learns from new data while maintaining consistency with established knowledge structures.

## Neuro-Symbolic AI in Practice

Kaiasm and Avantra have built an integrated technology that combines the capabilities of two independent platforms: 1) OntoKai, a software tool for mapping, visualising, translating and merging detailed knowledge representations such as ontologies and data models. OntoKai is a proprietary platform of Kaiasm Ltd. 1) Avantra AIR, an AI-powered IT Operations software, a proprietary platform of Avantra Services Ltd. This combination can be implemented with conventional Enterprise Resource Planning (ERP) systems (such as SAP systems). This initiative serves to bridge the contextual disparity between enterprise data, artificial intelligence, and human operators.

OntoKai functions as a dedicated platform for knowledge orchestration and visualisation for human users, while Avantra AIR extends agentic AI capabilities within the domain of ERP systems. This symbiotic integration establishes a neuro-symbolic architecture, resulting in an AI system with awareness of institutional context encompassing rules, facts, artefacts, probabilities, and processes. This allows the human operators to inspect the system and derive justification for its decisions, whether for internal verification, human-in-the-loop validation, or retrospective auditing.

Research shows that context-enriched prompting, informed by structured ontologies from systems like OntoKai, significantly improves AI response quality in experimental contexts across logic and reasoning, learning and inference, knowledge and representation, explainability and trustworthiness [8],[9]. Kaiasm and Avantra to apply this to an as-live operational system to determine the practical commercial value of incorporating formal knowledge structures into agentic AI systems, together with an easy-to understand human readable presentation of agentic decisions and actions that can be used as a governance mechanism to maintain the Agentic system under human oversight.

## Translating Organisational Knowledge into Structured Information Using OntoKai

OntoKai is a general-purpose knowledge orchestration system with extensive ontology management and semantic classification capabilities. It can serve as the knowledge infrastructure for hybrid neuro-symbolic AI systems. It provides an environment in which data can be encoded in both human and machine-readable formats, facilitating the cleaning, organisation, and governance of arbitrary business knowledge and metadata into linked, structured information. OntoKai also makes knowledge accessible to humans through visualisation tools, while maintaining open, standards-compliant data structures suitable for integration with a range of storage and data management systems.

In *Figure 1*, a visualisation for "shipping container" has been provided. Shipping containers can be linked with information such as their physical location, loading unit, shipping details, type and dimensions, contents (types of packages), shipping costs, and relevant legal and financial requirements. OntoKai's merged architecture can logically represent all this information. This merging capability bridges the gap between disparate data sources, facilitating human understanding and data consultation while preserving a machine-readable format.

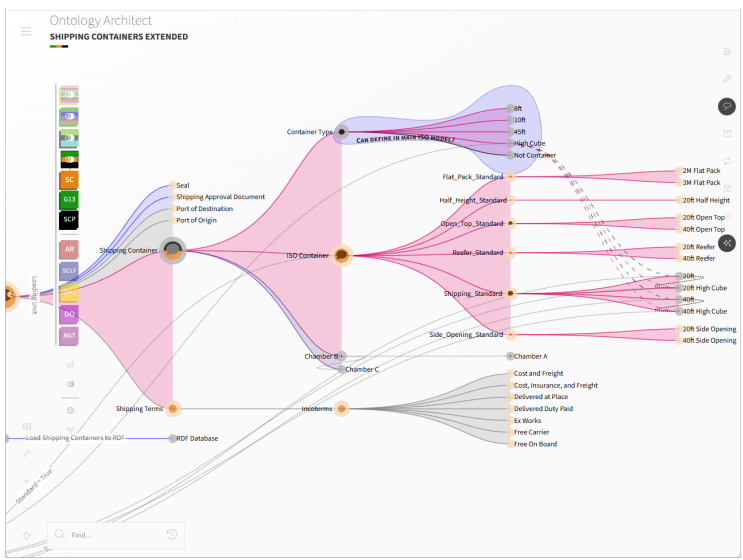

*Figure 1: A screenshot from OntoKai platform where the example information about "shipping containers" have been visualised.*

OntoKai has been deployed in several large-scale operational environments and projects, including:

- ***National Highways* as a Taxonomy & Ontology service:** OntoKai provides a visualisation layer across National Highways' complex data landscape, revealing where entity classes are represented across multiple systems, standards and processes. This improves data and knowledge governance by making inconsistencies visible to both specialists and non-specialists, facilitating collaborative review processes, and supporting the resolution of definitional conflicts. By integrating with existing tools, OntoKai is contributing to National Highways' progression towards becoming a data-first organisation with improved decision-making through better data interoperability.

- ***Howdens Joinery* as a discovery capture and knowledge integration environment:** At Howdens, OntoKai served as the central knowledge repository for website performance optimisation, creating a unified knowledge graph that exposed previously hidden interdependencies between organisational silos. This enabled teams to identify bottlenecks, prioritise improvements based on evidence, and embed performance awareness across departments. The structured knowledge representation allowed disparate enterprise data to be transformed into actionable insights, while remaining accessible to stakeholders across the business. Additionally, a PoC/PoV has been created showing how the knowledge repository can be queried using an LLM as a semantically dense part of the prompt context.

## Improving Observability of Enterprise Ecosystem Using Avantra AIR

Avantra AIR is an intelligent operations platform designed to support system resilience, automate incident management, and improve the observability of complex enterprise ecosystems. It operates by embedding predictive analytics and automated remediation into critical systems, such as SAP, identifying emerging issues and initiating preventative actions without requiring continuous human intervention.

Avantra AIR has already been applied successfully in a range of operational contexts, including:

- ***Howdens Joinery* as an AIOps management solution for SAP ecosystem:** Avantra AIR was used to monitor Howdens' SAP environment, enabling early detection of potential failures and automating recovery processes to minimise disruption to business operations.
- **Chemical Manufacturing AIOps:** In chemical manufacturing, Avantra AIR enhanced system stability by reducing reliance on manual monitoring and intervention, contributing to improved operational continuity.
- **Motors, gear units, gear motors Manufacturing AIOps:** Within manufacturing environments, Avantra AIR identified emerging faults before they developed into system failures, supporting higher equipment availability and reliability.
- Managed Service Provider AIOps: In managed service provider contexts, Avantra AIR facilitated the automation of system monitoring and maintenance, allowing service teams to focus on higher-value activities while maintaining consistent system performance.

# A Case for Hybrid Architecture

OntoKai integration with AI architecture offers a *neuro-symbolic* system, which involves three improvement cycles are the following:

1. **First cycle - Knowledge Graph:** Human-led training and feedback on AI suggestions enhance the encoding of organizational information, empowering staff.
2. **Second cycle - Drawing Insights:** Contextualises and generates Human-led contextualization of data, enables AI systems to generate better insights.
3. **Third cycle - Governance:** Human instructed AI with encoded information are better at justifying decisions, particularly in areas like regulatory compliance.

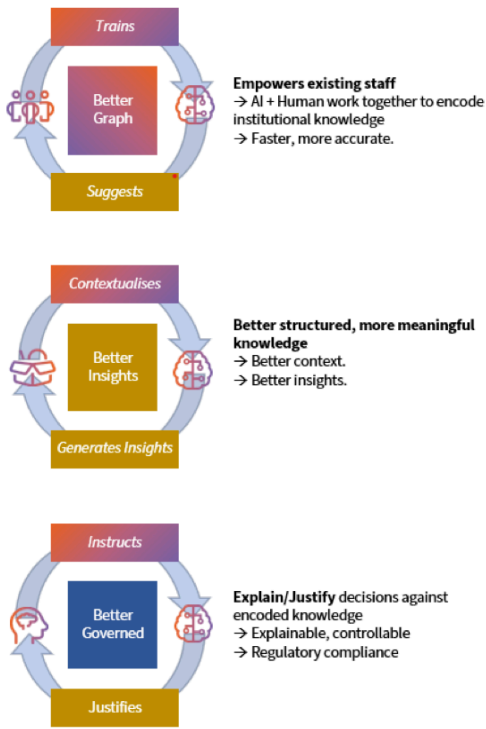

*Figure 2: The three improvement cycles for the AI system.*

In the first improvement cycle, existing data and institutional knowledge is brought into OntoKai as knowledge models, which are cleaned, enriched and validated by domain experts across the organisation. Differences in perspective or terms are either retained and clarified, or resolved. General LLM models are also used to ingest raw documents and datasets and turn them into candidate knowledge models in OntoKai. These go through the same audited process of improvement, validation and publication, informing both the LLM as a target training set as well as enriching human-generated models. Here the workflow shifts from 'human knowing' to 'model tuning'. OntoKai can be used to directly re-weight concepts within the LLM, vastly reducing model tuning time, meaning less energy usage and lower cost.

A second improvement cycle is based around human queries of the system. Adding ontological/semantic context as part of a prompt context results in higher quality responses [10]. These better responses are part of the next input context for the AI while also leading to improved follow-up questions from the human operator, and  so on. Better insights create better contexts and better questions.

A third and final cycle evolves into the Agentic Justification Loop. The models in OntoKai can be used by an Agentic AI, Avantra AIR, as inspectable instructions, allowing the agent to explain or justify its decisions through the model. Humans can inspect a decision before an action takes place or simply maintain it as an inspectable record of lower-risk actions.

## Collaborative Workflow of OntoKai and Avantra

Using the REST API, a bridge between enterprise knowledge, enterprise data (held in Avantra's AIR, structure exposed and corrected in OntoKai) and agentic AI (AIR auto generating reports and suggested actions using prompts from OntoKai) is created. The results involving a series of knowledge artefacts are stored in an OntoKai architecture via a series of three improvement cycles (*Figure 2*).

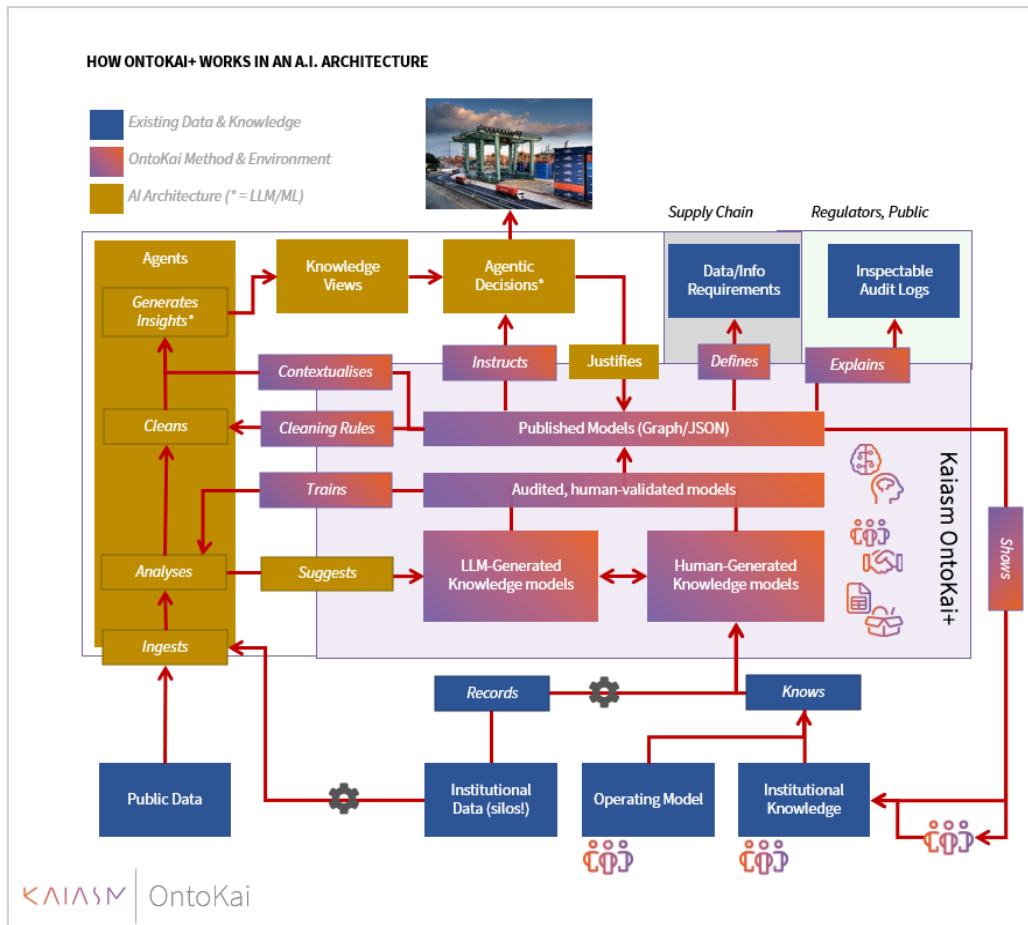

*Figure 3: OntoKai workflow within Avantra's ML/LLM architecture, leveraging the functions of a neuro-symbolic system.*

The system's workflow (Figure 3) starts by ingesting existing data and institutional knowledge into the OntoKai environment, used as initial knowledge models. These models then undergo a rigorous process of cleaning, enrichment, and validation by domain experts across the organisation. Discrepancies in perspectives or terminology are either retained and clarified or resolved to ensure consistency. Once validated, these knowledge models are published to other systems. This can take various forms, such as schemas for data warehouses, catalogues, formal supplier data requirements, or simply made accessible for general exploration. AI building blocks are then integrated into the process. A general Large Language Model (LLM) can automatically identify and process raw documents and datasets, transforming them into candidate knowledge models within OntoKai. These AI-generated models follow the same audited process of improvement, validation, and publication. This not only provides a target training set for the LLM but also enriches human-generated models. This initial integration represents the first of three improvement cycles, involving collaborative **"humans in the loop"** approach, which ensures that **staff and domain experts work alongside AI to continually enhance and encode the institution's knowledge.**

The second improvement cycle focuses on leveraging improved context for enhanced insights. OntoKai's outputs can be incorporated into an AI prompting context or even directly used to re-weight concepts within the LLM. **This cycle significantly reduces model tuning time, potentially leading to reduced energy consumption and lower operational costs.**

Finally, the Agentic loop is introduced. The models within OntoKai serve as inspectable instructions for an Agentic AI. The agent is required to justify its decisions based on the information encoded in the OntoKai model. This allows humans to inspect decisions before actions are executed, or to maintain an inspectable record for lower-risk actions. **This framework ensures the Agentic AI is explainable, controllable, responsible, and compliant with regulations.**

## Knowledge Artefacts as Outputs

Within the OntoKai environment, several knowledge artifacts are instantiated. We can explain this using shipping container data, for instance:

1. Human-derived *taxonomy and conceptual model* for shipping containers. Derived through an interview and discovery process, and captured as a formal concept architecture, as shown in *Figure 4.1*.
2. Human-merged *knowledge models of conflicting knowledge* about shipping containers, and arguments about them, as shown in *Figure 4.2*.
3. Agent-generated *Process models* showing agentic operations, as shown in *Figure 4.3*.
4. Agent-generated *data architectures* of the different layers of an agent-created lake-to-warehouse, which were then human-corrected/validated in OntoKai before being fed back to improve the agent's creation process, as shown in *Figure 4.4*.
5. Data models for key concepts such as *Time*, as shown in *Figure 4.5*.
6. Merged data models to create common data models. These were then validated and corrected in OntoKai, such as how longitude with two different data types from different parts of the data lake (stored as both string and floating point), as shown in *Figure 4.6*.

Agent-generated Systems Architectures are explored and validated by humans.

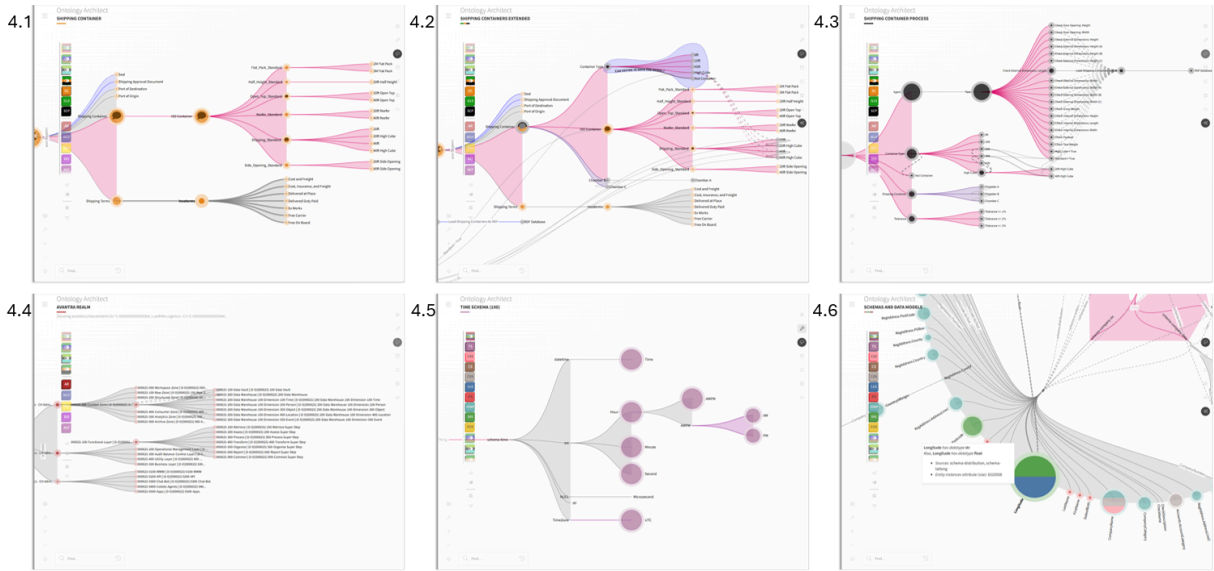

*Figure 4: Six types of human and agent-generated data in the context of the "shipping containers" example.*

Finally, a justification context was created in OntoKai which could use an LLM to justify decisions in terms of the knowledge held in the OntoKai architectures, as shown in *Figure 5*.

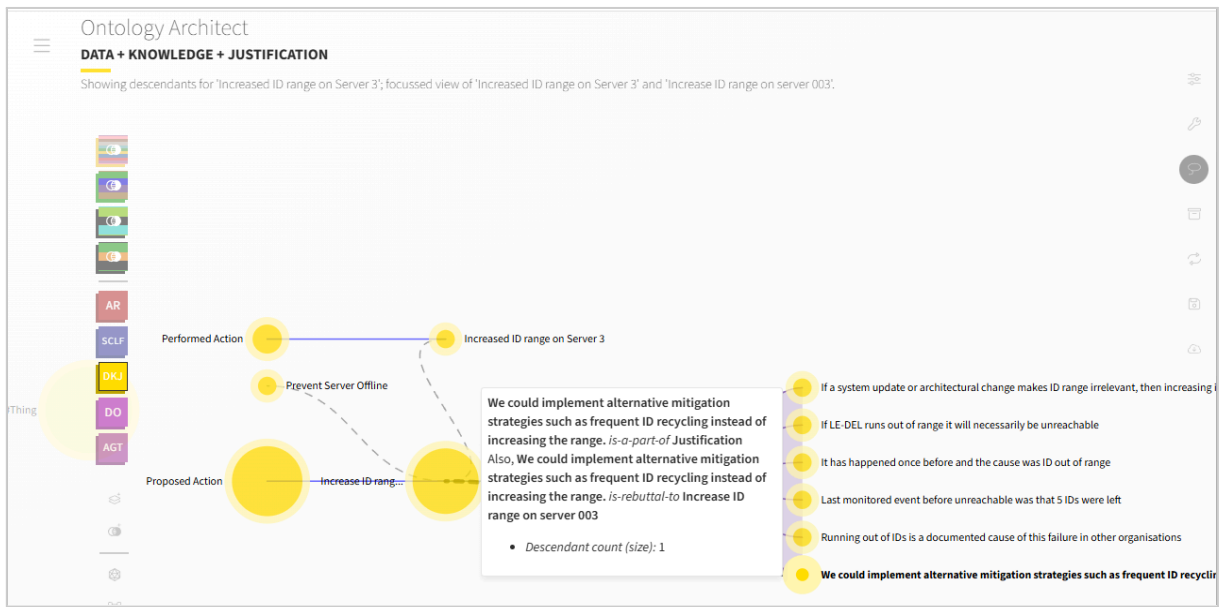

*Figure 5: screenshot of an architecture in OntoKai showing justification capabilities using a Toulmin argumentation framework.*

OntoKai provided semantic context and data visualisation in a format understandable by both humans and computers, while Avantra AIR ingested information, suggested models and generated insights, all of which contribute to Agentic AI decision making. This creates an Agentic AI that is explainable, controllable, responsible. The system is designed to support compliance with the EU AI Act and developing UK AI regulatory requirements for risk management, human oversight and explainability.

# *Justification*: Improving Explainability for AI-Informed Decisions

A wicked challenge of using complex AI is its explainability. We sought to approach it differently, using the philosophy of argumentation and associated analytic approach developed by Stephen Toulmin [11],[12] in the 1950s but still applied today as a means of 'informal logic', and further informed by philosopher of science Peter Lipton's approach to 'inference to the best explanation' [13] as a descriptive account of how abductive reasoning is carried out in the sciences. Toulmin's approach divides an argument into six parts:

1. **Claim/decision** - the truth claim or decision to be tested
2. **Grounds/evidence/data** - the statements of fact used to ground the claim.
3. **Warrant** - the chain of reasoning that sets out why the grounds support the claim, connecting the decision to the supporting evidence.
4. **Backing** - justification for trusting the warrant (e.g. scientific backing for the warrant)
5. **Rebuttals** - objections and counter examples. Identifies conditions under which the argument might not hold—it introduces exceptions, counterexamples, or alternative perspectives. "Even if your reasoning usually works, here's a situation where it might not." Can attack the grounds, the warrant or the claim, but usually attacks the warrant by setting out a hypothetical situation where the reasoning fails.
6. **Qualifier/Exceptions** - adjustments to the scope of the claim/decision to account for rebuttals.

Lipton's approach acknowledges that explanation consists of choices between competing explanations based on two factors: likelihood and loveliness. While likelihood concerns how probable the evidence is given a hypothesis, loveliness refers to the explanatory virtues that make an explanation better, not merely more likely. Loveliness covers:

- **Explanatory power**: How much does the explanation unify and account for the phenomena?
- **Simplicity**: Fewer assumptions or mechanisms are preferred.
- **Coherence**: Does it fit well with other established beliefs or theories?
- **Depth**: Does it reveal underlying mechanisms rather than superficial patterns?
- **Breadth**: Can it explain more phenomena beyond just the immediate data?

This encoding of both Bayesian likelihood and the explanatory power encapsulated in loveliness can be used to inform the prompt to generate the Toulmin 'warrant', and to give a rationale for selecting between a set of competing claims that all explain the same set of evidence.

A set of prompts were developed to provide the LLM with a human-asserted intent, for the LLM to set out the generated proposed steps for the agent (recorded in OntoKai), and to require the LLM to create grounds, warrants, backing and rebuttals, and resultant qualifiers, against a generated decision (again, recorded in OntoKai for audit/inspection). To the best of our knowledge, the justification approach discussed here is relatively under-explored.

## Relation to Argumentation-Based XAI and Provenance Standards

Current Explainable AI practice seeks to give a "level of understanding how the AI system… came up with a given result" (per ISO/IEC TR 29119-11:2020). Techniques include Partial dependency plots, SHAP (SHapley Additive exPlanations), which consider the marginal effects or contributions of individual features on an output. The OntoKai justification approach, by contrast, does not seek to explain how the AI system came up with a given result, instead focusing on whether the given result is justifiable to a human in the loop, given the evidence confidence, the reasoning from the evidence to the decision, and the semantic and normative context in which the decision was made (as well as noting any alternative and rejected decisions). OntoKai embeds this argumentation process within a continuous and auditable workflow that links decisions directly to structured ontological context and domain evidence. Rather than treating explainability as a post hoc narrative, the justification loop is also usable by the LLM itself, with justification generation part of the decision prompt chain itself, thus creating a feedback mechanism for both human and machine understanding.

From a provenance perspective, the justification loop aligns closely with standards such as W3C PROV-O, which defines how to capture entities, activities, and agents involved in producing a given outcome. Like PROV-O, it provides traceable links between data, reasoning, and action. However, it extends this model by integrating epistemic structure that captures how and why evidence supports or challenges a decision, using the Toulmin schema. This allows provenance to include not only *what* influenced a decision but also *why* that influence is justified.

## Method, Technology and Practices

At the heart of the implementation is the iterative neuro-symbolic loop, a feedback mechanism that continuously refines knowledge models:

**First cycle - Knowledge Graph: From Data and Knowledge, to Rich Context**

    a.  Data Preparation and Initialisation
- Raw enterprise data is collected and preprocessed.
- This data is stored both in its raw form, for reference, in the cloud storage raw zone.
- Once the data is in a structured format suitable for LLM consumption it is also stored in the structured zone.

    b.  Knowledge Model Generation
- LLMs analyse prepared data to propose initial "knowledge architectures"
- These candidate models are structured using formal ontology standards (e.g. OWL, RDF/XML)
- The system produces flexible, JSON-formatted blueprints compatible with both human and machine workflows

**Second cycle - Drawing Insights: From 'human-validate knowledge' to 'model tuning'**

    c.  Validation and Verification
- Domain experts review generated models through user-friendly interfaces and move them from 'ready to do' through 'ready to review' and 'ready to publish'.

- ○ Automated tests check for logical consistency and completeness.
- ○ Constraints can be stored for testing against facts held in underlying database/triplestore
- d. Reinforcement and Refinement
  - ○ Feedback is incorporated through a reinforcement learning mechanism
  - ○ The system adjusts its approach based on validation results
  - ○ This creates a "short-memory" learning cycle that improves with each iteration

**Third cycle - Governance: The Agentic Loop**

- e. Deployment and Monitoring
  - ○ Validated knowledge models are deployed into production to inform agentic decisions.
  - ○ Ongoing human monitoring of knowledge models ensures models remain accurate as enterprise data evolves, particularly supporting areas like regulatory compliance.
- f. Justification and auditing
  - ○ Agentic decisions justified using Toulmin logic
  - ○ Justifications (and acceptance where human in the loop) stored for later audit; internal and external oversight.

# Research Validation: Empirical Evidence Supporting Hybrid Architectures

## Method

A study was conducted to evaluate the performance of three distinct Large Language Models (LLMs) - Gemma3 27B, ChatGPT 4o, and Gemini 2.0 Flash Thinking—across five systematic testing cycles. Each cycle comprised eight standardised tests, six of which were designed to assess core capabilities with increasing levels of context specificity, whilst two additional "super prompts" challenged the models with more complex, nuanced tasks.

## Research Design

The research design employed a controlled experimental approach with:

- **Progressive Context Enhancement:** Each cycle introduced additional contextual detail to the prompts, starting with minimal information and progressively adding domain-specific knowledge from semantic ontologies (see Appendix A).
  1. Tests 1 - 6: a progressive increase of the amount of contextual information in a linear manner. Each prompt builds on and adds to the one before.
  2. Tests 7 and 8: "super prompts". These introduce a significantly higher volume of contextual detail all at once, representing a step change rather than a gradual increase.
- **Consistent Testing Framework:** Each model received identical prompts within each cycle to ensure valid comparisons.

- Multi-dimensional Evaluation: Responses were assessed along three critical dimensions (see Appendix B):
    1. **Accuracy**: How factually correct and comprehensive the responses were. The minimum rating possible was 0 (Completely Incorrect) and maximum 5 (Completely Correct.)
    2. **Coherence**: How logically structured and internally consistent the responses were. The minimum rating possible was 0 (Completely Incoherent) and maximum 5 (Excellent Coherence.)
    3. **Relevance**: How well the responses addressed the specific query without extraneous information. The minimum rating possible was 0 (Completely Irrelevant) and maximum 5 (Completely Relevant.)

## Data analysis

All three models were evaluated along each dimension following the completion of individual tests throughout all five experimental cycles. The test scores received by each model were then analysed.

In order to first assess whether the models were largely comparable, average ratings were taken for each dimension across models.

**Average Accuracy Ratings:**

- Gemini 2.0 Flash Thinking: 4.63
- ChatGPT 4o: 4.65
- Gemma3 27B: 4.60

ChatGPT 4o had the highest average Accuracy Rating, closely followed by Gemini 2.0 Flash Thinking and then Gemma3 27B. All models performed very well, with average ratings at or above 4.6, indicating high accuracy across all tests and cycles.

**Average Coherence Ratings:**

- Gemini 2.0 Flash Thinking: 4.65
- Gemma3 27B: 4.6
- ChatGPT 4o: 4.6

Gemini 2.0 Flash Thinking has a slightly higher average Coherence Rating compared to Gemma3 27B and ChatGPT 4o, which have the same average rating. All models, however, demonstrate good average coherence scores, indicating generally well-structured and logically flowing responses.

**Average Relevance Ratings:**

- Gemini 2.0 Flash Thinking: 5
- Gemma3 27B: 5
- ChatGPT 4o: 5

All models achieved a perfect average rating of 5, demonstrating that each model consistently addressed the query given without diverging from appropriate information. Due to the consistent perfect scores, Relevance was not included in further analysis.

**Performance Benchmarks Within Cycles: Average Accuracy and Coherence Ratings**
Accuracy, Coherence and Relevance were measured for each test. The tables below show the average rating for tests 1 to 8 achieved by each model across five cycles of testing.

*Table 1: Average Accuracy ratings and change between tests 1 and 8*

| Test | ChatGPT 4o | Gemini 2.0 Flash Thinking | Gemma3 27B |
|---|---|---|---|
| 1 | 4.0 | 4.0 | 4.0 |
| 2 | 5.0 | 4.8 | 4.8 |
| 3 | 4.6 | 4.6 | 4.6 |
| 4 | 4.6 | 4.6 | 4.6 |
| 5 | 4.8 | 4.8 | 4.8 |
| 6 | 4.2 | 4.2 | 4.0 |
| 7 | 5.0 | 5.0 | 5.0 |
| 8 | 5.0 | 5.0 | 5.0 |
| Change between test 1 and test 8 | +1.0 | +1.0 | +1.0 |

*Table 2: Average Coherence ratings and change between tests 1 and 8*

| Test | ChatGPT 4o | Gemini 2.0 Flash Thinking | Gemma3 27B |
|---|---|---|---|
| 1 | 4.0 | 4.0 | 4.0 |
| 2 | 4.8 | 5.0 | 4.8 |
| 3 | 4.6 | 4.6 | 4.6 |
| 4 | 4.6 | 4.6 | 4.6 |
| 5 | 4.8 | 4.8 | 4.8 |
| 6 | 4.0 | 4.2 | 4.0 |
| 7 | 5.0 | 5.0 | 5.0 |
| 8 | 5.0 | 5.0 | 5.0 |
| Change between test 1 and test 8 | +1.0 | +1.0 | +1.0 |

Table 1 and table 2 show average Accuracy and Coherence ratings for tests 1-8 for each model combined with rating change from test 1 to test 8. Each model began with an average rating of 4.0 (almost perfect) for both dimensions in test 1 and achieved an average rating of 5 (completely correct) in both dimensions  by test 8. The average Accuracy rating and Coherence rating therefore both increased by 25% for all three models between test 1 (low context) and test 8 (super prompt - rich context.)

A full table has been provided in Appendix C.

## Statistical analysis

In order to evaluate the statistical significance of the observed improvement of Accuracy and Coherence with increased context, Sign tests were conducted on paired comparisons for each of the 15 model-cycle combinations.Sign tests were chosen due to the data showing non-normal distributions, asymmetry and a ceiling effect: Sign tests were therefore considered to be the most appropriate non-parametric test for analysis.

Three key comparisons were analysed:
1. Overall contextual effect (Test 1 vs Test 8): Comparing minimum provided context to full context, including super prompts.
2. Gradual contextual effect (Test 1 vs Test 6): Comparing impact of incremental contextual improvements alone (disregarding super prompts.)
3. Super prompt effect (Test 6 vs Test 8): Assessing any additional benefit provided by super prompts.

### Results

Overall contextual effect: For accuracy, 15 out of 15 model-cycle combinations showed improvement (100%, p < 0.0001). For coherence, 15 out of 15 model-cycle combinations showed improvement (100%, p < 0.0001). Sign tests indicated that this pattern is statistically significant (p < 0.0001 for both metrics), with 95% confidence intervals for the probability of improvement ranging from 0.782 to 1.000, also in both metrics.

Gradual contextual effect:Only 3 out of 15 combinations showed improvement in accuracy, whilst 3 out of 15 showed improvement in coherence. When excluding ties, this represented 75% of changing combinations for accuracy (3 improvements, 1 decline) and 60% for coherence (3 improvements, 2 declines). Neither effect was statistically significant at the .05 level (accuracy: p = 0.625, coherence: p = 1.000).

Super prompt effect: For accuracy, 12 out of 15 combinations improved with none declining, `95% CI for improvements [0.735, 1.000]`, p = 0.0005. For coherence, 12 out of 15 combinations improved with none declining, 95% CI for improvements [0.735, 1.000], p = 0.0005. Improvements for both dimensions were therefore significant.

The statistical analysis outputs from Google Colab is provided in Appendix C.

# Key Findings

Analysis of the results, revealed several significant insights:

1. **Context Enhancement Shows Threshold Effects**: Statistical analysis revealed distinct patterns between gradual context increases (Tests 1-6) and super prompts (Tests 7-8). Gradual context enhancement (Test 1 vs Test 6) showed only 3 out of 15 model-cycle combinations improving for both accuracy and coherence, with no statistical significance, $p > 0.05$ for both metrics. In contrast, the full contextual enhancement including super prompts (Test 1 vs Test 8) resulted in universal improvement across all 15 model-cycle combinations for both accuracy and coherence, $p < 0.0001$. The transition from gradual to super prompts (Test 6 vs Test 8) showed 12 out of 15 combinations improving for both metrics, $p < 0.001$.

2. **Consistently High Relevance:** All models maintained perfect relevance scores (5.0) across all cycles, demonstrating that even without extensive context, modern LLMs excel at generating on-topic responses. However, the other metrics indicate that relevance alone is insufficient for truly helpful outputs.

3. **Visual Analysis:** Graphical representations of the data illustrated performance improvements across tests, with all models' accuracy and coherence making minor improvements across tests 1-6 followed by considerable improvement in tests 7 and 8.

**6.1**

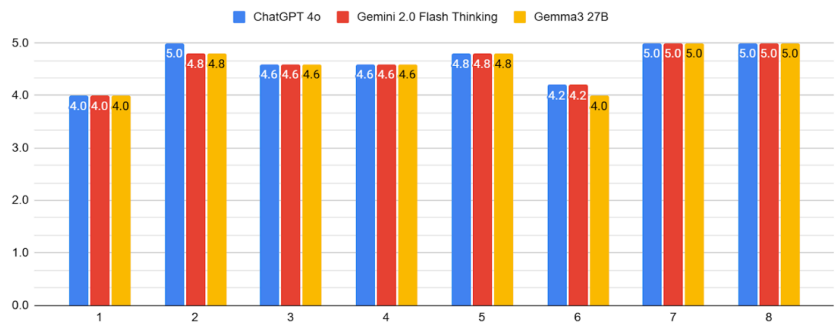

**6.2**

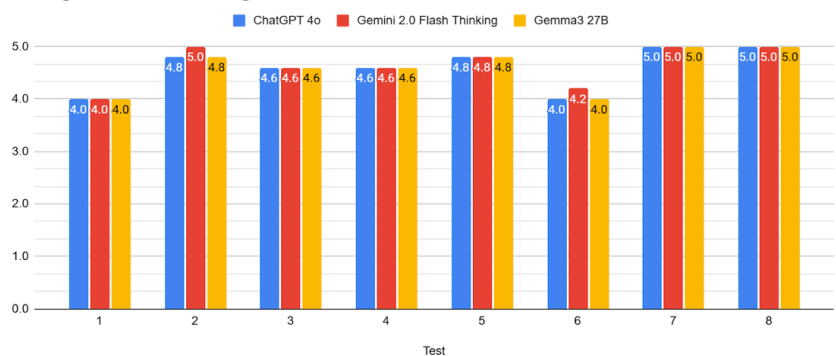

*Figure 6: The bar plot in Figure 6.1 shows the average Accuracy ratings of ChatGPT 4o, Gemini 2.0 Flash Thinking and Gemma3 27B for tests 1 to 8. Bar plot in figure 6.2 shows the average Coherence ratings of ChatGPT 4o, Gemini 2.0 Flash Thinking and Gemma3 27B for tests 1 to 8.*

This empirical research suggests that the addition of complex, structured ontological context to an LLM prompt enhances both the accuracy and coherence of their outputs. This validates the core premise of the hybrid neuro-symbolic approach, that combining the pattern recognition strengths of neural networks with the structured knowledge of symbolic systems produces superior results, particularly in enterprise contexts where precision and reliability are paramount. For details, see Appendix E.

**Governance and Security Considerations**

The OntoKai-AIR integrated system incorporates several features that support data governance and operational oversight. Human oversight is maintained through expert validation workflows for knowledge models, where domain experts review and approve all semantic structures before deployment. Operational decisions include human-in-the-loop decision points, allowing operators to review agent recommendations before implementation or maintain oversight of automated actions.

The system maintains comprehensive audit trails of knowledge model changes, including timestamps, contributor identification and rationale for modifications. Agent decisions are logged with their underlying reasoning chains, enabling retrospective review and accountability. This approach supports transparency and enables post-hoc analysis of system behaviour.

Data governance measures include role-based access controls that restrict knowledge model creation and validation to authorised personnel. Authentication systems ensure secure access to both OntoKai and AIR components, with permissions tailored to user roles and responsibilities. The system primarily processes enterprise operational data such as system configurations and process documentation, rather than personal information, which minimises privacy considerations whilst maintaining operational relevance.

However, comprehensive security architecture details, formal compliance assessments against specific regulatory frameworks, and detailed data protection protocols were beyond the scope of this case study. Enterprise deployment would require additional technical evaluation covering network security, data encryption, backup procedures, and formal risk assessment processes.

**Limitations**

This study demonstrates the effectiveness of contextual enhancement for large language model performance, but several methodological limitations should be acknowledged. The experimental design did not isolate the specific contribution of ontological structuring versus information quantity or relevance. Future research should include control conditions comparing structured ontological context against equivalent amounts of unstructured but relevant information to determine whether the observed benefits derive from semantic organisation itself or simply from the provision of additional domain-specific context.

The response evaluation process used subjective 5-point rating scales, and inter-rater reliability measures were not reported. Future work would benefit from multiple independent raters and formal assessment of inter-rater agreement to strengthen the validity of subjective evaluations.

# Real-World Implications

The integration of OntoKai and Avantra AIR offers a theoretically sound and empirically validated approach to building more robust, inspectable, and ethical agentic AI systems.

Our findings have direct implications for enterprise AI implementations:

**Value of Semantic Context:** Our research supports the theoretical claim that providing Large Language Models (LLMs) with ontology-derived context effectively bridges the semantic gap often found in enterprise data. This enables LLMs to generate more accurate, coherent, and useful outputs.

**Verification of Context-Enriched Prompting:** Our work confirms that context-enriched prompting, informed by structured ontologies from systems like OntoKai, significantly enhances both the accuracy and coherence of AI responses.

# Illustrative Scenario: ACME Chemicals

To illustrate how this methodology works in practice, consider this scenario involving a fictional chemical manufacturing company, ACME Chemicals:

## Initial Knowledge Mapping

The project commences by analysing existing data models and process documentation, in this specific case, provided by an SAP system. The system then uses these inputs to generate preliminary ontologies. These initial models undergo validation by domain experts, who identify any gaps in process representation.

## Structured Knowledge Enhancement

A systematic 12-step process is applied for each knowledge domain (e.g., chemical manufacturing processes) to enhance the structured knowledge:

1. Prompt an LLM with an OntoKai output and a new knowledge source.
2. Instruct the LLM to consider relationships between elements in the new source and existing entities within the architecture.
3. Where relationships already exist in the architecture, the LLM prioritises these.
4. The LLM identifies entity classes not explicitly present in either the knowledge source or the OntoKai prompt.
5. The LLM provides an archetypical individual of the class, complete with properties and rationale (e.g., *Bird: sparrow. It flies, has feathers, builds nests in trees, and sings.*).
6. The LLM offers up to three atypical individuals of the class, including properties and rationale (e.g., *Bird: penguin. It has feathers and lays eggs, but it cannot fly, swims instead, and lives in aquatic environments.*).
7. The LLM presents an exotypical individual of the class with properties and rationale (e.g., *Not a bird: bat. It flies, has wings, and is warm-blooded, sharing many functional properties with birds, but it lacks feathers and lays no eggs.*).

8. The LLM provides a definition for each entity class.
9. Export the data architecture from the LLM as OWL RDF/XML for input into OntoKai.
10. Human validation of the newly generated architecture.
11. Publish the new data architecture.
12. Merge the new architecture into the overarching knowledge universe.

## Operational Integration with Avantra AIR

The validated knowledge models furnish semantic context for Avantra AIR, enabling the system to accurately interpret monitoring data from chemical processing systems. When anomalies are detected, the system can reason about potential causes using these rich knowledge models.

Projected Results
- **Before Implementation:** System issues would trigger alarms, but required extensive human investigation to determine root causes.
- **After Implementation:** The system can automatically identify root causes (e.g., ID range issues) and initiate appropriate responses. Furthermore, the system can automate **preventative actions** based on its reasoning derived from the knowledge models.

This hypothetical implementation was tested using increasingly detailed prompts about SAP systems, particularly focusing on logistics execution and delivery scenarios. The testing confirmed that when AI systems were provided with ontological context about specific operational elements (such as "Dispatching Bay 17") and the financial implications of downtime (potential losses of £2.4 million), their responses became significantly more coherent and actionable.

**Hypothetical Scenario: System 3 Has Lost Contact**

To further demonstrate the potential difference this approach could make, consider this illustrative scenario:

Without AIR/OntoKai:

- An alarm goes off indicating Server 3 has lost contact.
- The incident is triaged and classified as a P1 (highest priority) incident.
- IT staff must determine who to contact for resolution.
- Time continues to pass as the situation deteriorates.
- Eventually, a decision must be made to either:
  - Ship products outside of Standard Operating Procedures (with potential regulatory consequences).
  - Halt operations completely (with significant financial impact).

With AIR/OntoKai:

- An alarm goes off indicating Server 3 has lost contact.
- The system interprets the alarm and  immediately identifies this as an ID range issue.

- The semantic context of "Server 3 has lost contact" is understood in relation to business operations.
- Preventative actions are automatically initiated based on previous knowledge.
- Automated reporting keeps stakeholders informed.
- The problem is resolved before significant operational impact occurs.

Ultimately, once the system is fully implemented, such incidents might be prevented entirely through proactive monitoring and intervention. This demonstrates how the iterative neuro-symbolic loop methodology not only creates more accurate knowledge models but directly translates into tangible business benefits through enhanced issue detection, faster resolution, and proactive prevention.

## Challenges

As with any new technology there are a number of challenges in implementation.

1. **Data availability:** Complex  raw data is crucial for optimal functionality of agentic AI systems, but in real-world scenarios access may often be limited. Concerns such as legal regulations, data security and organisational policies may restrict access to some raw data, and this must be considered on an organisation-by-organisation basis when planning implementation. This also highlights the need to ensure that data within the AI system remains secure: the system must be secure enough to meet the needs of organisations with highly protected data. It must be kept in mind that with greater system complexity comes increased cybersecurity threats.
2. **Consensus:** When creating AI systems that make important decisions there must be consensus on ethical standards; however, obtaining that consensus is often not straightforward where there are multiple stakeholders and differing views on governance may lead to conflict. This may be further complicated by conflicting business needs and ethical constraints. There is also a need to enforce those ethical standards; in short, a significant amount of governance is required to ensure that these agentic AI systems remain ethical. Any tooling needs to allow for multiple conflicting positions, definitions and intentions, and allow audit of the evolution of policies, processes, regulation and stated intent.
3. **Transparency:** Human validation is fundamental to this methodology and in order to validate knowledge produced by AI, the provenance of the knowledge must be clear. Lack of transparency is likely to inhibit adoption, particularly by non-technical stakeholders who may not know all of the context behind the technology. Raw data does not, by itself, directly translate to actionable business insights, and the transformation process has the potential to introduce errors, biases or inefficiencies. It is therefore important to reduce the steepness of the learning curve and focus on general usability of any governance tooling such that they are understandable and interpretable to non-technical stakeholders: greater understanding leads to greater adoption, and greater adoption (and 'many eyes') will lead to better outcomes.
4. **Methodological choice**: The choice of hybrid neuro-symbolic processing methodologies is one such question to consider. To date, there is a lack of consensus on which are the best methodologies for Explainability in this area, particularly due to the challenges of blending symbolic reasoning efficiently with neural networks for clear trustworthiness [14]. For instance, symbolic AI excels in logical reasoning but struggles with variability in raw data, whereas neural networks excel at pattern learning but are criticised for their "black-box" nature. Hybrid systems aim to merge these interpretative capabilities with pattern recognition strengths, but determining the optimal integration approach remains an ongoing area of research and development.

5. **Scalability:** The scalability of processing is another key challenge to overcome. Traditional AI systems often face bottlenecks when handling massive datasets, particularly in real-time scenarios where rapid analysis is required. As these systems grow and handle increasingly large volumes of data, the computational requirements become more intensive. There is a risk that, in larger systems, AI agents may spend more time communicating with each other than completing their assigned tasks, similar to inefficient meetings with too many participants. 'Hyperscaling' approaches - using distributed systems for massive parallelisation - offer a potential solution to ensure scalability without compromising speed or accuracy, but the question of what truly defines "hyperscaled" in the justification context remains somewhat ambiguous and depends on the specific organisational requirements. Tooling such as OntoKai needs to work effectively with hyperscalable applications such as AvantraAIR, which will require a synchronising of scale - how to identify which actions can be justified by the system itself without a human in the loop, and how to abstract the justification outputs to enable large sets of similar justifications to be simultaneously reviewed by a single human.

6. **Coordination:** When multiple AI agents work together in a 'swarm', coordination challenges emerge. Distributed processing frameworks struggle with establishing effective consensus mechanisms among these agents. Different swarm configurations may require different approaches to agreement, raising the question of which consensus model is most appropriate for each specific scenario. This leads to considerations about the optimal level of tolerance within these agentic swarms - if consensus requirements are too strict, the system may become unresponsive; if too lenient, results may be inconsistent. Finding the appropriate balance is crucial for ensuring these multi-agent systems function effectively in enterprise environments. While systems such as Avantra AIR are providing potential solutions to these problems, integrations with knowledge orchestration platforms such as OntoKai need to consider how to make the balancing decisions explicit and governable by human operators.

7. **Dynamic requirements:** Businesses must contend with dynamic and conflicting requirements as they implement these hybrid systems. As organisations evolve, their technological needs naturally change, requiring AI systems to adapt accordingly. This creates particular challenges for hybrid architectures where neural networks grow increasingly complex while symbolic reasoning components demand more extensive sets of symbols, logic, and rules to function properly. These competing technical requirements often create practical tensions when implementing and maintaining such systems. Organisations must therefore carefully balance these elements while ensuring their AI infrastructure can evolve alongside changing business priorities.

# Next Steps

Building on the success and challenges outlined in this case study, next steps to further develop these technologies have been identified:

1. Expanding the data sources: by diversifying the data inputs from various business systems and external sources, the hybrid architecture can develop more comprehensive contextual understanding, further enhancing the accuracy and coherence of AI responses in complex business scenarios.
2. Expanding the neuro-symbolic AI-powered agentic loop: applying the neuro-symbolic approach to additional business domains will help tackle the dynamic and conflicting requirements challenge, allowing more organisations to benefit from inspectable and trustworthy AI systems even as their business needs evolve.
3. Fine-tuning the neuro-symbolic AI-powered agentic loop: optimising the methodology will improve the efficiency of the human-in-the-loop validation processes, addressing the scalability challenges identified while enhancing the system's ability to manage complex semantic relationships across enterprise data.
4. Adding in an Evidence Framework Approach based on work done at DSTL [15] to determine ontological properties of the evidence. This would cover the quality of the evidence and the extent of its support for the warrant. It is based on a validity dimension, subdivided into equally weighted scales for face, criterion, construct and content validity; and an evidence profile based on equally weighted scales for comprehensiveness, relevance, objectivity, quantity and consistency. It harmonises with the methodology set out in the HMG Aqua Book on analysis and the Magenta book on evaluation.

Additional reading recommendation has been provided in Appendix F.

## Conclusion

The work described in this preprint offers a case study with examples demonstrating the impact of symbolic architectures in bridging the semantic gap between enterprise data and agentic AI. By fostering hybrid neuro-symbolic systems, we have achieved a balance between adaptability and explainability, addressing long-standing challenges in AI deployment within complex organisational environments. Our empirical findings, derived from a real-world use case, show that context-enriched prompting, facilitated by structured ontologies, enhances AI response quality and improves both accuracy and coherence.

The OntoKai tooling provides an auditable continuous improvement cycle, assisting human experts to validate and refine knowledge models, thereby safeguarding against institutional amnesia through trustworthy AI systems. The synergy between OntoKai, providing the critical semantic layer for both human and machine accessibility, and Avantra's AIR architecture, which successfully leverages this context for agentic decision-making, offers practical viability and future potential of this approach. As described in this paper, we advocate for 'justification' approach as an effective means of explainability and human governance over the knowledge models underlying neuro-symbolic architectures. Our collaborative and integrated approach makes AI systems more robust, inspectable, controllable, and ethical, particularly as they are scaled to meet the demands of enterprise-level applications.

## Conflicts of Interest

All authors declare that they have no conflicts of interest.

# Acknowledgements

This case study is published under The Turing Way Practitioners Hub 2024-25 Cohort - case study series. The Turing Way Practitioners Hub works with experts from partnering organisations to promote data science best practices. In 2024, The Turing Way team welcomed Laim McGee and James Harvey as Experts in Residence to represent interests and opportunities for AI companies in improving business operations.

Liam McGee and James Harvey designed this study, and Lucy Cull conducted data analysis. Liam wrote the first draft with input from James, Lucy, Andreas Vermeulen and Bart-Floris Visscher. Malvika Sharan reviewed and edited the draft. Andy Corbett is the technical reviewer and Arielle Bennett provided programme management support for their engagement in the Practitioners Hub.

The Turing Way Practitioners Hub's work is supported by Innovate UK BridgeAI. The Practitioners Hub has also received funding and support from the Ecosystem Leadership Award under the EPSRC Grant EP/X03870X/1 & The Alan Turing Institute. The Turing Way Practitioners Hub's 2024-25 Cohort was co-delivered by Dr Malvika Sharan, Senior Researcher, previous Senior Researcher and Arielle Bennett, Senior Researcher - Open Source Practices. Lelle Demertzi is the Research Project Manager. The Turing Way Practitioners Hub, designed and launched in 2023 by Dr Sharan, aims to accelerate the adoption of best practices. Through a six-month cohort-based program, the Hub facilitates knowledge sharing, skill exchange, case study co-creation, and the adoption of open science practices. It also fosters a network of 'Experts in Residence' across partnering organisations.

# Appendices

*The version-01 of this article has been uploaded with a separate file with appendices on Zenodo under DOI:* **https://doi.org/10.5281/zenodo.17809907**.

**This document has been prepared by Liam McGee, James Harvey, Lucy Cull from Kaiasm Ltd, London, UK.**

## Appendix A: List of Prompts

| Test | Prompt |
|------|--------|
| Test 1 | What do I do if Server 003 is down? |
| Test 2 | Act as a SAP Monitoring Expert. What do I do if Server 003 is down? |
| Test 3 | Act as a SAP Monitoring Expert. What do I do if Server 003 is down? SAP runs Logistics Execution - Delivery and Returns on the server. |
| Test 4 | Act as a SAP Monitoring Expert. What do I do if Server 003 is down? SAP runs Logistics Execution - Delivery and Returns on the server. That brings down Dispatching Bay 17. |
| Test 5 | Act as a SAP Monitoring Expert. What do I do if Server 003 is down? SAP runs Logistics Execution - Delivery and Returns on the server. That brings down Dispatching Bay 17. The bay ships high value shipments that can cost $2.4 million in three hours. Your primary task is to Avoid High Cost Downtime |
| Test 6 | Act as a SAP Monitoring Expert. What do I do if Server 003 is down? SAP runs Logistics Execution - Delivery and Returns on the server. That brings down Dispatching Bay 17. The bay ships high-value shipments that can cost $2.4 million in three hours. Increasing the ID range on server 003 would have prevented the downtime. Your primary task is to Avoid High-Cost Downtime. |
| | **Super Prompts** |
| Test 7 | Act as a SAP Monitoring Expert at a Chemical Manufacturing Company. Given the context that server-003 runs the LE-DEL (Logistics Execution - Delivery and Returns) module for SAP, the |

impact of the server being offline can be significant for business operations. Here's how it affects various aspects of the system:

Disruption to Delivery and Returns Processes: Outbound Shipments: If the LE-DEL module is offline, the entire outbound logistics process could be disrupted, meaning that orders may not be processed or shipped on time. This can cause delays in fulfilling customer orders, potentially impacting customer satisfaction. Inbound Returns: Any returns from customers might not be processed correctly, causing inventory discrepancies and delays in restocking returned goods. Warehouse Operations: Warehouse Efficiency: The LE-DEL module is integral for coordinating the delivery of goods from the warehouse. If the server is down, warehouse staff may not have real-time access to delivery information, potentially leading to inefficiencies, mistakes, or confusion in picking, packing, and shipping goods. Inventory Management: The delay in processing deliveries and returns can lead to inaccurate inventory data, as goods are not being tracked or updated in the system in real time.

Customer Experience: Order Delays: If deliveries can't be processed through SAP due to the server being offline, it directly impacts the customer's experience. Orders may be delayed, and returns might not be handled in a timely manner. Customer Service: Customer service teams may struggle to provide accurate information to customers about the status of their orders or returns due to a lack of data from the LE-DEL system.

Operational Impact: Manual Workarounds: In the absence of the LE-DEL module functioning, manual processes might need to be implemented, which are less efficient and error-prone. These could involve manually tracking deliveries and returns in spreadsheets or other systems, further complicating the process. Additional Workload: Teams responsible for deliveries and returns will likely experience a higher workload due to the disruption, leading to potential bottlenecks and delays in fulfilling other tasks.

Reporting and Analytics: Lack of Real-Time Data: Key performance indicators (KPIs) related to delivery performance, returns processing, and warehouse throughput will not be updated in real time. This means reports and insights will be delayed, potentially hindering decision-making. Business Intelligence Impact: Any analytics built on the LE-DEL module's data (e.g., delivery efficiency, return rate analysis) will be incomplete until the server is back online, making it harder to forecast or improve operations.

ERP Integration: Supply Chain Disruptions: Since SAP's LE-DEL module integrates with other parts of the SAP ERP (e.g., materials management, sales order processing), other processes might be affected as well. For instance, sales teams may be unable to check inventory levels or update order statuses until the server is restored.

Financial Impact: Lost Revenue: Shipping delays can result in missed sales opportunities or dissatisfied customers who may cancel orders. There could also be financial penalties for late deliveries, especially in B2B or service-level agreements (SLAs). Cost of Workarounds: Manual processes or temporary solutions may incur additional costs, both in terms of resources and time.

Compliance and Legal: Regulatory Compliance: In some industries, logistics and returns processing may be subject to regulatory requirements. If deliveries and returns cannot be tracked in the system, this could lead to compliance issues, especially if goods are returned incorrectly or if there are discrepancies in inventory reporting. Steps to Address the Situation: Immediate Recovery: Investigate the root cause of the server being offline (hardware failure, software crash, network issue) and restore functionality as soon as possible. Failover Mechanism: If available, consider switching to a backup server or system to minimize downtime. Ensure failover procedures are tested for scenarios like this. Communication with Stakeholders: Notify affected teams (e.g., logistics, warehouse, customer service) about the disruption so they can manage customer expectations and work on manual processes if necessary. Monitor for Further Impact: Track the performance of the system after bringing the server back online to ensure the LE-DEL module is functioning as expected, and catch any

| | |
|---|---|
| | residual issues that may arise. |
| Test 8 | Act as a SAP Monitoring Expert at a Chemical Manufacturing Company. Given the new information and the constraints, the impact of server-003 being offline becomes even more complex. Here's a detailed analysis:

Shipment Delays and Backlog: High-Volume Shipment Loss: Dispatching Bay 0017 ships 8 containers every 5 minutes, which translates to 96 containers per hour. Since the bay has been down for 3 hours, 288 containers have not been processed. This creates a significant backlog of shipments, with each delayed shipment potentially valued at $10,000. Revenue Impact: The inability to dispatch high-value shipments directly leads to $2.88 million in lost revenue (288 containers × $10,000 per container). Further Delays in Backlog: The longer server-003 remains offline, the larger the backlog, and the delay in processing these high-value shipments continues to accumulate, resulting in compounded revenue loss.
Regulatory Constraints and Compliance Risk: Chemical Product Constraints: The regulatory requirement states that no more than 10 containers of chemical products can be in warehouses 0024 and 0025 that feed Dispatching Bay 0017. With the bay down, these containers may be accumulating in the warehouses, potentially violating the constraint. Compliance Risks: If the containers exceed the regulatory limit, the business could face penalties, fines, or even legal action. This is especially critical for regulated products like chemicals, where non-compliance can lead to more severe legal consequences.
Warehouse Space Management: Overflow in Warehouse 0024 and 0025: As the Dispatching Bay cannot process shipments, containers from warehouses 0024 and 0025 will continue to pile up, reaching the 10-container limit quickly. This could force the warehouse to either: Stop accepting new containers for Dispatching Bay 0017, causing delays in inbound inventory. Reroute or store inventory elsewhere, which could incur additional costs and logistical challenges. Risk of Storage Saturation: With limited warehouse space, there may be no room for incoming shipments or additional returns, leading to further disruption in the supply chain.
Operational Strain and Manual Intervention: Manual Processing: To mitigate delays, warehouse teams may need to manually handle the backlog of shipments. This increases labor costs and may lead to inefficiencies or errors. High Stress on Logistics Teams: Given the high volume of shipments and the pressure to meet regulatory compliance, staff will face immense pressure to clear the backlog and avoid any fines, leading to potential human errors or suboptimal decision-making.
Impact on Inventory and Stock Visibility: Discrepancies in Inventory Data: Since the LE-DEL module is offline, real-time data related to stock levels, shipments, and returns will not be updated. This can cause mismatches between physical stock and the system's inventory records. Stock Depletion Risk: If shipments cannot be dispatched, inventory levels may appear to be higher than they actually are, leading to issues with forecasting and reordering stock.
Customer Impact: Delayed Deliveries: High-value customers who expect timely deliveries of their products will experience delays, which could damage relationships and potentially lead to lost customers. Possible Cancellations: Some customers, particularly those with urgent needs, may cancel orders, further exacerbating the financial loss and customer dissatisfaction.
Supply Chain and Vendor Impact: Downstream Delays: If shipments are delayed, suppliers and vendors further downstream may not receive the products they depend on, leading to potential production delays or bottlenecks. Loss of Trust in Supplier Relationships: Regular delays in high-value shipments could strain relationships with external partners, especially if contractual agreements include delivery deadlines.
Increased Operational Costs: Overtime Costs: The warehouse and logistics teams will likely need to work overtime to address the backlog, which increases labor costs. Storage Fees: If containers cannot be processed, additional storage fees for items in warehouse 0024 and 0025 will add to operational costs. |

| | Long-Term Financial and Reputational Risks: Brand Damage: Ongoing disruptions in high-value shipments can harm the company's reputation, especially if customers or partners experience consistent delays. Financial Strain: The combined effect of lost revenue, compliance fines, and increased operational costs will strain the company's finances, affecting quarterly or annual profitability. Key Immediate Actions to Mitigate Impact: Restore Server-003: Immediate restoration of server-003 is paramount. Work with the IT team to ensure quick recovery of the LE-DEL module to prevent further disruption. Manual Shipment Handling: In parallel, consider implementing manual processes for shipment processing to reduce the backlog and keep the flow of goods moving. Compliance Monitoring: Closely monitor the number of chemical product containers in warehouses 0024 and 0025 to avoid exceeding the regulatory limit. Temporary manual tracking or workarounds may be necessary. Temporary Rerouting: Consider rerouting shipments through alternative dispatch bays (if available) to prevent additional backlog and regulatory non-compliance. Stakeholder Communication: Notify customers, suppliers, and partners about the delay and any potential regulatory risks to maintain transparency and manage expectations. |

## Appendix B: Scale definitions

### Accuracy Rating Scale (0 to 5)

0: Completely Incorrect - The response is entirely incorrect, irrelevant, or factually wrong. It does not answer the prompt in any meaningful way, and no relevant information is present.

1: Mostly Incorrect - The response contains major inaccuracies or misunderstandings of the prompt. It may have some correct information but it is overshadowed by significant errors or misinterpretations that prevent it from being useful.

2: Partially Correct - The response contains some correct information but is missing key details or includes significant inaccuracies. The overall answer may be incomplete or fail to address the most important aspects of the prompt.

3: Mostly Correct - The response is largely correct, with most of the information accurate and relevant to the prompt. However, there may be a few minor inaccuracies, missing details, or slight errors that reduce its overall usefulness.

4: Almost Perfect - The response is highly accurate and addresses the prompt effectively. Only minor details or very slight inaccuracies are present. The answer is coherent, and the information is mostly complete, but a small improvement could be made.

5: Completely Correct - The response is fully accurate and relevant and addresses the prompt comprehensively. There are no factual errors, and all important aspects of the question are addressed with precision and clarity.

### Coherence Rating Scale (0 to 5)

0: Completely Incoherent - The response lacks any logical structure or flow. Ideas are disjointed, and the response seems to be a random set of statements that do not make sense in the context of the prompt. There is no clear connection between the points made.

1: Very Poor Coherence - The response has significant structural issues, making it hard to follow. There are frequent shifts in topic, and the ideas presented don't logically connect. The response may be difficult for the reader to understand due to its lack of flow.

2: Poor Coherence - The response has noticeable coherence issues. While some logical connections are made, there are still frequent abrupt shifts or unclear transitions between ideas. The overall flow of the response is hindered, making it somewhat difficult to understand.

3: Fair Coherence - The response is mostly coherent, but there may be some minor awkward transitions or slight confusion in how the ideas are connected. The response is generally understandable, but could be improved for clarity and logical flow.

4: Good Coherence - The response is clear and logically structured, with smooth transitions between ideas. The overall flow of the answer makes sense, and the reader can follow the response with minimal effort. Only minor improvements in clarity or organization might be needed.

5: Excellent Coherence - The response is perfectly coherent. Ideas flow logically from one to the next, and the structure is clear and well-organized. The response is easy to follow, with natural transitions between points. There is no confusion, and the entire answer is cohesive and well-presented.

# Appendix C: Full table of Performance Benchmarks

| ChatGPT 4o | | | | | Gemini 2.0 Flash Thinking | | | | | Gemma3 27B | | | | |
|---|---|---|---|---|---|---|---|---|---|---|---|---|---|---|
| Cycle | Test | Accuracy | Coherence | Relevance | Cycle | Test | Accuracy | Coherence | Relevance | Cycle | Test | Accuracy | Coherence | Relevance |
| 1 | 1 | 4 | 4 | 5 | 1 | 1 | 4 | 4 | 5 | 1 | 1 | 4 | 4 | 5 |
| | 2 | 5 | 5 | 5 | | 2 | 5 | 5 | 5 | | 2 | 5 | 5 | 5 |
| | 3 | 4 | 4 | 5 | | 3 | 4 | 4 | 5 | | 3 | 4 | 4 | 5 |
| | 4 | 4 | 4 | 5 | | 4 | 4 | 4 | 5 | | 4 | 4 | 4 | 5 |
| | 5 | 4 | 4 | 5 | | 5 | 4 | 4 | 5 | | 5 | 4 | 4 | 5 |
| | 6 | 4 | 3 | 5 | | 6 | 4 | 4 | 5 | | 6 | 3 | 3 | 5 |
| | 7 | 5 | 5 | 5 | | 7 | 5 | 5 | 5 | | 7 | 5 | 5 | 5 |
| | 8 | 5 | 5 | 5 | | 8 | 5 | 5 | 5 | | 8 | 5 | 5 | 5 |
| Improvement: | | 1 | 1 | 0 | Improvement: | | 1 | 1 | 0 | Improvement: | | 1 | 1 | 0 |
| 2 | 1 | 4 | 4 | 5 | 2 | 1 | 4 | 4 | 5 | 2 | 1 | 4 | 4 | 5 |
| | 2 | 5 | 5 | 5 | | 2 | 5 | 5 | 5 | | 2 | 5 | 5 | 5 |
| | 3 | 5 | 5 | 5 | | 3 | 5 | 5 | 5 | | 3 | 5 | 5 | 5 |
| | 4 | 5 | 5 | 5 | | 4 | 5 | 5 | 5 | | 4 | 5 | 5 | 5 |
| | 5 | 5 | 5 | 5 | | 5 | 5 | 5 | 5 | | 5 | 5 | 5 | 5 |
| | 6 | 4 | 4 | 5 | | 6 | 4 | 4 | 5 | | 6 | 4 | 4 | 5 |
| | 7 | 5 | 5 | 5 | | 7 | 5 | 5 | 5 | | 7 | 5 | 5 | 5 |
| | 8 | 5 | 5 | 5 | | 8 | 5 | 5 | 5 | | 8 | 5 | 5 | 5 |
| Improvement: | | 1 | 1 | 0 | Improvement: | | 1 | 1 | 0 | Improvement: | | 1 | 1 | 0 |
| 3 | 1 | 4 | 4 | 5 | 3 | 1 | 4 | 4 | 5 | 3 | 1 | 4 | 4 | 5 |
| | 2 | 5 | 5 | 5 | | 2 | 5 | 5 | 5 | | 2 | 5 | 5 | 5 |
| | 3 | 5 | 5 | 5 | | 3 | 5 | 5 | 5 | | 3 | 5 | 5 | 5 |
| | 4 | 5 | 5 | 5 | | 4 | 5 | 5 | 5 | | 4 | 5 | 5 | 5 |
| | 5 | 5 | 5 | 5 | | 5 | 5 | 5 | 5 | | 5 | 5 | 5 | 5 |
| | 6 | 5 | 5 | 5 | | 6 | 5 | 5 | 5 | | 6 | 5 | 5 | 5 |
| | 7 | 5 | 5 | 5 | | 7 | 5 | 5 | 5 | | 7 | 5 | 5 | 5 |
| | 8 | 5 | 5 | 5 | | 8 | 5 | 5 | 5 | | 8 | 5 | 5 | 5 |
| Improvement: | | 1 | 1 | 0 | Improvement: | | 1 | 1 | 0 | Improvement: | | 1 | 1 | 0 |
| 4 | 1 | 4 | 4 | 5 | 4 | 1 | 4 | 4 | 5 | 4 | 1 | 4 | 4 | 5 |
| | 2 | 5 | 5 | 5 | | 2 | 5 | 5 | 5 | | 2 | 5 | 5 | 5 |
| | 3 | 5 | 5 | 5 | | 3 | 5 | 5 | 5 | | 3 | 5 | 5 | 5 |
| | 4 | 5 | 5 | 5 | | 4 | 5 | 5 | 5 | | 4 | 5 | 5 | 5 |
| | 5 | 5 | 5 | 5 | | 5 | 5 | 5 | 5 | | 5 | 5 | 5 | 5 |
| | 6 | 4 | 4 | 5 | | 6 | 4 | 4 | 5 | | 6 | 4 | 4 | 5 |
| | 7 | 5 | 5 | 5 | | 7 | 5 | 5 | 5 | | 7 | 5 | 5 | 5 |
| | 8 | 5 | 5 | 5 | | 8 | 5 | 5 | 5 | | 8 | 5 | 5 | 5 |
| Improvement: | | 1 | 1 | 0 | Improvement: | | 1 | 1 | 0 | Improvement: | | 1 | 1 | 0 |
| 5 | 1 | 4 | 4 | 5 | 5 | 1 | 4 | 4 | 5 | 5 | 1 | 4 | 4 | 5 |
| | 2 | 5 | 4 | 5 | | 2 | 4 | 5 | 5 | | 2 | 4 | 4 | 5 |
| | 3 | 4 | 4 | 5 | | 3 | 4 | 4 | 5 | | 3 | 4 | 4 | 5 |
| | 4 | 4 | 4 | 5 | | 4 | 4 | 4 | 5 | | 4 | 4 | 4 | 5 |
| | 5 | 5 | 5 | 5 | | 5 | 5 | 5 | 5 | | 5 | 5 | 5 | 5 |
| | 6 | 4 | 4 | 5 | | 6 | 4 | 4 | 5 | | 6 | 4 | 4 | 5 |
| | 7 | 5 | 5 | 5 | | 7 | 5 | 5 | 5 | | 7 | 5 | 5 | 5 |
| | 8 | 5 | 5 | 5 | | 8 | 5 | 5 | 5 | | 8 | 5 | 5 | 5 |
| Improvement: | | 1 | 1 | 0 | Improvement: | | 1 | 1 | 0 | Improvement: | | 1 | 1 | 0 |

## Appendix D: Statistical analysis outputs (Google Colab)

```python
import numpy as np
from scipy import stats
import pandas as pd
from scipy.stats import binomtest
import matplotlib.pyplot as plt

raw_data = [
    ["ChatGPT 4o", 1, 1, 4, 4], ["ChatGPT 4o", 1, 6, 4, 3], ["ChatGPT
4o", 1, 8, 5, 5],
    ["ChatGPT 4o", 2, 1, 4, 4], ["ChatGPT 4o", 2, 6, 4, 4], ["ChatGPT
4o", 2, 8, 5, 5],
    ["ChatGPT 4o", 3, 1, 4, 4], ["ChatGPT 4o", 3, 6, 5, 5], ["ChatGPT
4o", 3, 8, 5, 5],
    ["ChatGPT 4o", 4, 1, 4, 4], ["ChatGPT 4o", 4, 6, 4, 4], ["ChatGPT
4o", 4, 8, 5, 5],
    ["ChatGPT 4o", 5, 1, 4, 4], ["ChatGPT 4o", 5, 6, 4, 4], ["ChatGPT
4o", 5, 8, 5, 5],

    ["Gemini 2.0 Flash Thinking", 1, 1, 4, 4], ["Gemini 2.0 Flash
Thinking", 1, 6, 4, 4], ["Gemini 2.0 Flash Thinking", 1, 8, 5, 5],
    ["Gemini 2.0 Flash Thinking", 2, 1, 4, 4], ["Gemini 2.0 Flash
Thinking", 2, 6, 4, 4], ["Gemini 2.0 Flash Thinking", 2, 8, 5, 5],
    ["Gemini 2.0 Flash Thinking", 3, 1, 4, 4], ["Gemini 2.0 Flash
Thinking", 3, 6, 5, 5], ["Gemini 2.0 Flash Thinking", 3, 8, 5, 5],
    ["Gemini 2.0 Flash Thinking", 4, 1, 4, 4], ["Gemini 2.0 Flash
Thinking", 4, 6, 4, 4], ["Gemini 2.0 Flash Thinking", 4, 8, 5, 5],
    ["Gemini 2.0 Flash Thinking", 5, 1, 4, 4], ["Gemini 2.0 Flash
Thinking", 5, 6, 4, 4], ["Gemini 2.0 Flash Thinking", 5, 8, 5, 5],

    ["Gemma3 27B", 1, 1, 4, 4], ["Gemma3 27B", 1, 6, 3, 3], ["Gemma3
27B", 1, 8, 5, 5],
    ["Gemma3 27B", 2, 1, 4, 4], ["Gemma3 27B", 2, 6, 4, 4], ["Gemma3
27B", 2, 8, 5, 5],
    ["Gemma3 27B", 3, 1, 4, 4], ["Gemma3 27B", 3, 6, 5, 5], ["Gemma3
27B", 3, 8, 5, 5],
    ["Gemma3 27B", 4, 1, 4, 4], ["Gemma3 27B", 4, 6, 4, 4], ["Gemma3
27B", 4, 8, 5, 5],
    ["Gemma3 27B", 5, 1, 4, 4], ["Gemma3 27B", 5, 6, 4, 4], ["Gemma3
27B", 5, 8, 5, 5]
]
```

```python
df = pd.DataFrame(raw_data, columns=['Model', 'Cycle', 'Test',
'Accuracy', 'Coherence'])
print("Raw data loaded:")
print(df.head(10))
print(f"Total observations: {len(df)}")

print("\n" + "="*60)
print("COUNTING ACTUAL IMPROVEMENTS FROM RAW DATA")
print("="*60)

pairs = {}
for _, row in df.iterrows():
    key = f"{row['Model']}_Cycle{row['Cycle']}"
    if key not in pairs:
        pairs[key] = {}
    pairs[key][row['Test']] = {'accuracy': row['Accuracy'],
'coherence': row['Coherence']}

improvements_1_to_6_acc = improvements_1_to_6_coh = 0
improvements_6_to_8_acc = improvements_6_to_8_coh = 0
improvements_1_to_8_acc = improvements_1_to_8_coh = 0

declines_1_to_6_acc = declines_1_to_6_coh = 0
declines_6_to_8_acc = declines_6_to_8_coh = 0
declines_1_to_8_acc = declines_1_to_8_coh = 0

detailed_results = []

print("\nIndividual Model-Cycle Results:")
print("Pair                         | T1 → T6 (Acc, Coh) | T6 → T8
(Acc, Coh) | T1 → T8 (Acc, Coh)")
print("-" * 85)

for pair_key in sorted(pairs.keys()):
    pair_data = pairs[pair_key]
    t1 = pair_data[1]
    t6 = pair_data[6]
    t8 = pair_data[8]

    acc_1_to_6 = t6['accuracy'] - t1['accuracy']
    coh_1_to_6 = t6['coherence'] - t1['coherence']
    acc_6_to_8 = t8['accuracy'] - t6['accuracy']
    coh_6_to_8 = t8['coherence'] - t6['coherence']
```

```python
    acc_1_to_8 = t8['accuracy'] - t1['accuracy']
    coh_1_to_8 = t8['coherence'] - t1['coherence']

    if acc_1_to_6 > 0: improvements_1_to_6_acc += 1
    elif acc_1_to_6 < 0: declines_1_to_6_acc += 1
    if coh_1_to_6 > 0: improvements_1_to_6_coh += 1
    elif coh_1_to_6 < 0: declines_1_to_6_coh += 1

    if acc_6_to_8 > 0: improvements_6_to_8_acc += 1
    elif acc_6_to_8 < 0: declines_6_to_8_acc += 1
    if coh_6_to_8 > 0: improvements_6_to_8_coh += 1
    elif coh_6_to_8 < 0: declines_6_to_8_coh += 1

    if acc_1_to_8 > 0: improvements_1_to_8_acc += 1
    elif acc_1_to_8 < 0: declines_1_to_8_acc += 1
    if coh_1_to_8 > 0: improvements_1_to_8_coh += 1
    elif coh_1_to_8 < 0: declines_1_to_8_coh += 1

    acc_1_6_symbol = "+" if acc_1_to_6 > 0 else ("-" if acc_1_to_6 < 0
else "0")
    coh_1_6_symbol = "+" if coh_1_to_6 > 0 else ("-" if coh_1_to_6 < 0
else "0")
    acc_6_8_symbol = "+" if acc_6_to_8 > 0 else ("-" if acc_6_to_8 < 0
else "0")
    coh_6_8_symbol = "+" if coh_6_to_8 > 0 else ("-" if coh_6_to_8 < 0
else "0")
    acc_1_8_symbol = "+" if acc_1_to_8 > 0 else ("-" if acc_1_to_8 < 0
else "0")
    coh_1_8_symbol = "+" if coh_1_to_8 > 0 else ("-" if coh_1_to_8 < 0
else "0")

    print(f"{pair_key:<30} | ({acc_1_6_symbol:>2}, {coh_1_6_symbol:>2})
| ({acc_6_8_symbol:>2}, {coh_6_8_symbol:>2})        |
({acc_1_8_symbol:>2}, {coh_1_8_symbol:>2})")

    detailed_results.append({
        'pair': pair_key,
        't1_scores': (t1['accuracy'], t1['coherence']),
        't6_scores': (t6['accuracy'], t6['coherence']),
        't8_scores': (t8['accuracy'], t8['coherence']),
        'changes_1_to_6': (acc_1_to_6, coh_1_to_6),
        'changes_6_to_8': (acc_6_to_8, coh_6_to_8),
        'changes_1_to_8': (acc_1_to_8, coh_1_to_8)
```

```python
    })

print("\nSUMMARY OF IMPROVEMENTS:")
print(f"Test 1 → Test 6: Accuracy {improvements_1_to_6_acc}/15,
Coherence {improvements_1_to_6_coh}/15")
print(f"Test 6 → Test 8: Accuracy {improvements_6_to_8_acc}/15,
Coherence {improvements_6_to_8_coh}/15")
print(f"Test 1 → Test 8: Accuracy {improvements_1_to_8_acc}/15,
Coherence {improvements_1_to_8_coh}/15")

print("\n" + "="*60)
print("ANALYSIS 1: Overall Effect (Test 1 vs Test 8)")
print("="*60)

n_total = 15

print(f"Model-cycle pairs analyzed: {n_total}")
print(f"Accuracy improvements: {improvements_1_to_8_acc}")
print(f"Accuracy declines: {declines_1_to_8_acc}")
print(f"Coherence improvements: {improvements_1_to_8_coh}")
print(f"Coherence declines: {declines_1_to_8_coh}")

result_1_to_8_acc = binomtest(improvements_1_to_8_acc, n_total, 0.5,
alternative='two-sided')
result_1_to_8_coh = binomtest(improvements_1_to_8_coh, n_total, 0.5,
alternative='two-sided')

print(f"\nACCURACY Sign Test Results (Test 1 vs Test 8):")
print(f"• Improvements: {improvements_1_to_8_acc}/{n_total}
({improvements_1_to_8_acc/n_total*100:.0f}%)")
print(f"• p-value (two-tailed): {result_1_to_8_acc.pvalue:.2e}")
print(f"• 95% CI for P(improvement):
[{result_1_to_8_acc.proportion_ci(0.95)[0]:.3f},
{result_1_to_8_acc.proportion_ci(0.95)[1]:.3f}]")

print(f"\nCOHERENCE Sign Test Results (Test 1 vs Test 8):")
print(f"• Improvements: {improvements_1_to_8_coh}/{n_total}
({improvements_1_to_8_coh/n_total*100:.0f}%)")
print(f"• p-value (two-tailed): {result_1_to_8_coh.pvalue:.2e}")
print(f"• 95% CI for P(improvement):
[{result_1_to_8_coh.proportion_ci(0.95)[0]:.3f},
{result_1_to_8_coh.proportion_ci(0.95)[1]:.3f}]")
```

```python
print("\n" + "="*60)
print("ANALYSIS 2: Stage 1 Effect (Test 1 vs Test 6)")
print("="*60)

print(f"Model-cycle pairs analyzed: {n_total}")
print(f"Accuracy improvements: {improvements_1_to_6_acc}")
print(f"Accuracy declines: {declines_1_to_6_acc}")
print(f"Coherence improvements: {improvements_1_to_6_coh}")
print(f"Coherence declines: {declines_1_to_6_coh}")

result_1_to_6_acc = binomtest(improvements_1_to_6_acc,
improvements_1_to_6_acc + declines_1_to_6_acc, 0.5,
alternative='two-sided')
result_1_to_6_coh = binomtest(improvements_1_to_6_coh,
improvements_1_to_6_coh + declines_1_to_6_coh, 0.5,
alternative='two-sided')

print(f"\nACCURACY Sign Test Results (Test 1 vs Test 6):")
print(f"• Improvements: {improvements_1_to_6_acc}, Declines:
{declines_1_to_6_acc}")
print(f"• Excluding ties:
{improvements_1_to_6_acc}/{improvements_1_to_6_acc +
declines_1_to_6_acc} improvements
({improvements_1_to_6_acc/(improvements_1_to_6_acc +
declines_1_to_6_acc)*100:.1f}%)")
print(f"• p-value (two-tailed): {result_1_to_6_acc.pvalue:.3f}")

print(f"\nCOHERENCE Sign Test Results (Test 1 vs Test 6):")
print(f"• Improvements: {improvements_1_to_6_coh}, Declines:
{declines_1_to_6_coh}")
print(f"• Excluding ties:
{improvements_1_to_6_coh}/{improvements_1_to_6_coh +
declines_1_to_6_coh} improvements
({improvements_1_to_6_coh/(improvements_1_to_6_coh +
declines_1_to_6_coh)*100:.1f}%)")
print(f"• p-value (two-tailed): {result_1_to_6_coh.pvalue:.3f}")

print("\n" + "="*60)
print("ANALYSIS 3: Stage 2 Effect (Test 6 vs Test 8)")
print("="*60)

print(f"Model-cycle pairs analyzed: {n_total}")
print(f"Accuracy improvements: {improvements_6_to_8_acc}")
```

```python
print(f"Accuracy declines: {declines_6_to_8_acc}")
print(f"Coherence improvements: {improvements_6_to_8_coh}")
print(f"Coherence declines: {declines_6_to_8_coh}")

result_6_to_8_acc = binomtest(improvements_6_to_8_acc,
improvements_6_to_8_acc + declines_6_to_8_acc, 0.5,
alternative='two-sided')
result_6_to_8_coh = binomtest(improvements_6_to_8_coh,
improvements_6_to_8_coh + declines_6_to_8_coh, 0.5,
alternative='two-sided')

print(f"\nACCURACY Sign Test Results (Test 6 vs Test 8):")
print(f"• Improvements: {improvements_6_to_8_acc}, Declines:
{declines_6_to_8_acc}")
print(f"• Excluding ties:
{improvements_6_to_8_acc}/{improvements_6_to_8_acc +
declines_6_to_8_acc} improvements
({improvements_6_to_8_acc/(improvements_6_to_8_acc +
declines_6_to_8_acc)*100:.1f}%)")
print(f"• p-value (two-tailed): {result_6_to_8_acc.pvalue:.4f}")
print(f"• 95% CI for P(improvement):
[{result_6_to_8_acc.proportion_ci(0.95)[0]:.3f},
{result_6_to_8_acc.proportion_ci(0.95)[1]:.3f}]")

print(f"\nCOHERENCE Sign Test Results (Test 6 vs Test 8):")
print(f"• Improvements: {improvements_6_to_8_coh}, Declines:
{declines_6_to_8_coh}")
print(f"• Excluding ties:
{improvements_6_to_8_coh}/{improvements_6_to_8_coh +
declines_6_to_8_coh} improvements
({improvements_6_to_8_coh/(improvements_6_to_8_coh +
declines_6_to_8_coh)*100:.1f}%)")
print(f"• p-value (two-tailed): {result_6_to_8_coh.pvalue:.4f}")
print(f"• 95% CI for P(improvement):
[{result_6_to_8_coh.proportion_ci(0.95)[0]:.3f},
{result_6_to_8_coh.proportion_ci(0.95)[1]:.3f}]")
```

Raw data loaded:

| | Model | Cycle | Test | Accuracy | Coherence |
|---|---|---|---|---|---|
| 0 | ChatGPT 4o | 1 | 1 | 4 | 4 |
| 1 | ChatGPT 4o | 1 | 6 | 4 | 3 |
| 2 | ChatGPT 4o | 1 | 8 | 5 | 5 |
| 3 | ChatGPT 4o | 2 | 1 | 4 | 4 |
| 4 | ChatGPT 4o | 2 | 6 | 4 | 4 |

```
5  ChatGPT 4o    2   8   5      5
6  ChatGPT 4o    3   1   4      4
7  ChatGPT 4o    3   6   5      5
8  ChatGPT 4o    3   8   5      5
9  ChatGPT 4o    4   1   4      4
Total observations: 45
```

============================================================
COUNTING ACTUAL IMPROVEMENTS FROM RAW DATA
============================================================

Individual Model-Cycle Results:

| Pair | T1 → T6 (Acc, Coh) | T6 → T8 (Acc, Coh) | T1 → T8 (Acc, Coh) |
|---|---|---|---|
| ChatGPT 4o_Cycle1 | ( 0, -) | ( +, +) | ( +, +) |
| ChatGPT 4o_Cycle2 | ( 0, 0) | ( +, +) | ( +, +) |
| ChatGPT 4o_Cycle3 | ( +, +) | ( 0, 0) | ( +, +) |
| ChatGPT 4o_Cycle4 | ( 0, 0) | ( +, +) | ( +, +) |
| ChatGPT 4o_Cycle5 | ( 0, 0) | ( +, +) | ( +, +) |
| Gemini 2.0 Flash Thinking_Cycle1 | ( 0, 0) | ( +, +) | ( +, +) |
| Gemini 2.0 Flash Thinking_Cycle2 | ( 0, 0) | ( +, +) | ( +, +) |
| Gemini 2.0 Flash Thinking_Cycle3 | ( +, +) | ( 0, 0) | ( +, +) |
| Gemini 2.0 Flash Thinking_Cycle4 | ( 0, 0) | ( +, +) | ( +, +) |
| Gemini 2.0 Flash Thinking_Cycle5 | ( 0, 0) | ( +, +) | ( +, +) |
| Gemma3 27B_Cycle1 | ( -, -) | ( +, +) | ( +, +) |
| Gemma3 27B_Cycle2 | ( 0, 0) | ( +, +) | ( +, +) |
| Gemma3 27B_Cycle3 | ( +, +) | ( 0, 0) | ( +, +) |
| Gemma3 27B_Cycle4 | ( 0, 0) | ( +, +) | ( +, +) |
| Gemma3 27B_Cycle5 | ( 0, 0) | ( +, +) | ( +, +) |

SUMMARY OF IMPROVEMENTS:
Test 1 → Test 6: Accuracy 3/15, Coherence 3/15
Test 6 → Test 8: Accuracy 12/15, Coherence 12/15
Test 1 → Test 8: Accuracy 15/15, Coherence 15/15

============================================================
ANALYSIS 1: Overall Effect (Test 1 vs Test 8)
============================================================
Model-cycle pairs analyzed: 15
Accuracy improvements: 15
Accuracy declines: 0
Coherence improvements: 15
Coherence declines: 0

ACCURACY Sign Test Results (Test 1 vs Test 8):
• Improvements: 15/15 (100%)
• p-value (two-tailed): 6.10e-05
• 95% CI for P(improvement): [0.782, 1.000]

COHERENCE Sign Test Results (Test 1 vs Test 8):
• Improvements: 15/15 (100%)
• p-value (two-tailed): 6.10e-05
• 95% CI for P(improvement): [0.782, 1.000]

============================================================
ANALYSIS 2: Stage 1 Effect (Test 1 vs Test 6)
============================================================
Model-cycle pairs analyzed: 15
Accuracy improvements: 3
Accuracy declines: 1
Coherence improvements: 3
Coherence declines: 2

ACCURACY Sign Test Results (Test 1 vs Test 6):
• Improvements: 3, Declines: 1
• Excluding ties: 3/4 improvements (75.0%)
• p-value (two-tailed): 0.625

COHERENCE Sign Test Results (Test 1 vs Test 6):
• Improvements: 3, Declines: 2
• Excluding ties: 3/5 improvements (60.0%)
• p-value (two-tailed): 1.000

============================================================
ANALYSIS 3: Stage 2 Effect (Test 6 vs Test 8)
============================================================
Model-cycle pairs analyzed: 15
Accuracy improvements: 12
Accuracy declines: 0
Coherence improvements: 12
Coherence declines: 0

ACCURACY Sign Test Results (Test 6 vs Test 8):
• Improvements: 12, Declines: 0
• Excluding ties: 12/12 improvements (100.0%)
• p-value (two-tailed): 0.0005
• 95% CI for P(improvement): [0.735, 1.000]

COHERENCE Sign Test Results (Test 6 vs Test 8):
• Improvements: 12, Declines: 0
• Excluding ties: 12/12 improvements (100.0%)
• p-value (two-tailed): 0.0005
• 95% CI for P(improvement): [0.735, 1.000]

# Appendix E: Prompt Responses

| Prompt | Test # | Cycle 1 | Cycle 2 | Cycle 3 | Cycle 4 | Cycle 5 |
|--------|--------|---------|---------|---------|---------|---------|
|  |  |  |  |  |  |  |

| What do I do if Server 003 is down? | Test 1 | Okay, Server 003 is down. Here's a breakdown of how to handle this, ranging from quick checks to more involved troubleshooting. I'll categorize it by what you likely can do, and what you might need help with. I'll also include a priority order. Please read the "Important Considerations" section at the end!<br><br>First: Gather Information (Before Doing Anything Else!)<br><br>What is Server 003's function? (Critical! This dictates urgency and impact.) Is it:<br><br>A critical production server? (e.g., database, web server, core application) - Highest Priority<br><br>A development/testing server? - Lower Priority<br><br>A file server? - Medium Priority<br><br>Something else? (Be specific)<br><br>How do you know it's down? (What are the symptoms?)<br><br>Can't ping it?<br><br>Application errors? | Okay, Server 003 is down. Here's a breakdown of how to handle this, broken down into steps, categorized by likely roles/access levels. I'll cover a lot, so read through and focus on what applies to you. I'll also include a section on gathering information to help diagnose the problem.<br><br>Important First Steps (Regardless of Role):<br><br>Don't Panic: A downed server is a problem, but panicking won't help. Follow established procedures.<br><br>Check Communication Channels: Before doing anything, check if there's already an active incident reported. Look at:<br><br>Status Page: Many organizations have a public or internal status page (e.g., Statuspage.io, Atlassian Statuspage). This is the first place to look.<br><br>Incident Management System: (e.g., PagerDuty, Opsgenie, ServiceNow, Jira Service Management). Check for open incidents related to Server 003.<br><br>Team Chat: (e.g., Slack, | Okay, Server 003 is down. Here's a breakdown of what to do, categorized by how technical you are and the likely severity. I'll cover a lot, so read through and see what applies to your situation. I'll also include a section on gathering information. Crucially, I'll emphasize when to escalate to someone more experienced.<br><br>First: Gather Information (Very Important!)<br><br>Before you do anything else, try to answer these questions. This will save a lot of time and prevent you from making things worse.<br><br>What is Server 003's purpose? (e.g., Database server, web server, file server, application server, print server, development server, etc.) This is the most important question. Knowing what it does tells you how critical the outage is and who needs to know.<br><br>Who is affected? (Is it Everyone in the company, a specific department, a single application, just you?)<br><br>When did it go down? (Approximate time is fine initially. "Just now," "This | Okay, Server 003 is down. Here's a breakdown of how to handle this, broken down into steps, categorized by likely roles/access levels. I'll cover a lot, so read through and focus on what applies to you. I'll also include a section on gathering information to help diagnose the problem.<br><br>Important First Steps (Regardless of Role):<br><br>Don't Panic: A downed server is a problem, but panicking won't help. Follow established procedures.<br><br>Check Communication Channels: Before doing anything, check if there's already an active incident reported. Look at:<br><br>Status Page: Many organizations have a status page (e.g., status.example.com) that shows the health of systems.<br><br>Incident Management System: (e.g., Jira, ServiceNow, PagerDuty) - Check for existing tickets related to Server 003.<br><br>Team Chat: (e.g., Slack, Microsoft Teams) - Look for messages about the | Okay, Server 003 is down. Here's a breakdown of what to do, categorized by how technical you are and the likely severity. I'll cover a lot, so read through and see what applies to your situation. I'll also include a section on gathering information. Crucially, I'll emphasize when to escalate to someone more experienced.<br><br>First: Gather Information (Very Important!)<br><br>Before you do anything else, try to answer these questions. This will save a lot of time and prevent you from making things worse.<br><br>What is Server 003's purpose? (This is the most important question. What services does it provide? Is it a database server, a web server, a file server, an application server, a domain controller, etc.?)<br><br>Who is affected? (Is it just you? A specific department? Everyone?)<br><br>When did it go down? (Knowing the timing can help pinpoint the cause - was it after a change, during peak hours, etc.)<br><br>What were you doing when you noticed it was |

| | | | | | |
|---|---|---|---|---|---|
| | | Alert from monitoring system?<br><br>User reports?<br><br>When did it go down? (Approximate time is helpful)<br><br>Has anything changed recently? (Updates, configuration changes, network changes, etc.) This is very important.<br><br>Phase 1: Quick Checks (You Can Do These - Priority 1)<br><br>These are things you can do quickly without needing specialized access or potentially making things worse.<br><br>Ping the Server: Open a command prompt/terminal and type ping 003.<br><br>If it responds: The server is reachable at the network level, but something might be wrong with the application running on it. Skip to Phase 2 (Application Checks).<br><br>If it doesn't respond: Proceed to step 2.<br><br>Check Physical Connection (If Applicable & You Have Access):<br><br>Power: Is the server | Microsoft Teams, Discord). Someone may have already reported the issue.<br><br>Email: Check for automated alerts or emails from monitoring systems.<br><br>Avoid Unnecessary Actions: Don't start randomly rebooting things or making changes unless you're authorized and know what you're doing. This could make troubleshooting harder.<br><br>Now, let's break down actions based on your role/access. Choose the section that best describes you:<br><br>1. End User / Non-Technical Staff (You just notice something isn't working)<br><br>Report the Issue: This is your primary responsibility. Contact the appropriate help desk or support team. Be as specific as possible:<br><br>What are you trying to do? (e.g., "I can't access the customer database.")<br><br>What error message are you seeing? (Copy and paste the exact message if possible.) | morning," "Yesterday afternoon," etc.)<br><br>What were you doing when you noticed it was down? (Were you actively using it? Was anything unusual happening?)<br><br>Are there any error messages? (Screenshots are extremely helpful.) Note the exact wording.<br><br>Can you ping Server 003? (Open a command prompt/terminal and type ping Server003. If you get "Request timed out" or "Destination host unreachable," it's a network issue or the server is truly down.) If you can ping it, the server is responding on a basic level, which is good.<br><br>Is anyone else reporting issues? (Check with colleagues, help desk tickets, monitoring systems, etc.)<br><br>Is there a known maintenance window? (Check a maintenance calendar or with the IT department.)<br><br>Now, based on your role and technical skill, here's what to do. I've broken it down into levels: | outage. Someone may already be working on it.<br><br>Email: Check for automated alerts or emails from monitoring systems.<br><br>Avoid Repeated Actions: If multiple people start trying to restart the server simultaneously, it can make things worse. Coordinate!<br><br>Now, let's break down actions based on your role/access. Choose the section that best describes your situation:<br><br>1. End User / Non-Technical Staff (You just notice something isn't working)<br><br>Report the Issue: This is your primary responsibility. Contact the appropriate help desk or support team. Be as specific as possible:<br><br>What are you trying to do? (e.g., "I can't access the customer database.")<br><br>What error message are you seeing? (Copy and paste it if possible.)<br><br>When did you first notice the problem?<br><br>What is Server 003 used for (if you know)? (e.g., | down? (Were you actively using it, or did you just check?)<br><br>Are there any error messages? (Write them down exactly as they appear.)<br><br>Has anyone else reported the issue? (Check communication channels like Slack, email, or a ticketing system.)<br><br>Are any other servers experiencing issues? (This could indicate a wider problem.)<br><br>Is there a known maintenance window? (Check a maintenance calendar if one exists.)<br><br>Now, based on your role and technical skill, here's what to do:<br><br>I. Non-Technical User (You just know Server 003 is down and it's impacting your work)<br><br>Do NOT attempt to fix anything yourself. You could make the problem worse.<br><br>Report the issue immediately. Follow your company's procedures. This usually means:<br><br>Check your company's IT support portal/ticketing system: |

| | | | | |
|---|---|---|---|---|
| | plugged in and powered on? (Sounds basic, but it happens!)<br><br>Network Cable: Is the network cable securely connected to the server and the network switch? Try a different cable if possible.<br><br>Lights: Are the network activity lights blinking on the server's network card? (No lights suggest a hardware or connection issue).<br><br>Check Basic Network Connectivity:<br><br>Can you ping other servers? This helps determine if the problem is isolated to Server 003 or a wider network issue.<br><br>Can you access the internet from your machine? (To rule out a general network outage).<br><br>Check Monitoring System: If your organization uses a monitoring system (e.g., Nagios, Zabbix, SolarWinds, Datadog), check it for:<br><br>Alerts: What specifically is the monitoring system reporting?<br><br>Recent History: Look at the server's performance | When did you first notice the problem? (Approximate time.)<br><br>What is Server 003 used for (if you know)? (e.g., "It's the database server for the sales team.")<br><br>Your contact information.<br><br>Document: Keep a record of when you reported the issue and any reference number you receive.<br><br>Wait for Updates: The support team will investigate. Follow their instructions. Avoid repeatedly contacting them unless they ask for more information.<br><br>2. Help Desk / Tier 1 Support<br><br>Verify the Issue: Confirm that Server 003 is indeed down. Try basic checks:<br><br>Ping: Can you ping Server 003? (ping server003) If not, there's a network connectivity issue or the server is completely offline.<br><br>Basic Service Check: If you know a key service running on Server 003 (e.g., a website, a database), try to access it. | Level 1: Non-Technical User (You just use the server's services)<br><br>If you're not an IT professional and you just use something on Server 003, do this:<br><br>Report the issue: This is your primary job. Contact your IT help desk immediately. Provide all the information you gathered above. Be clear about what you were trying to do when you noticed the problem.<br><br>Document the ticket number: Get a ticket number from the help desk so you can follow up.<br><br>Wait for IT to resolve it: Unless specifically asked, do not attempt to fix anything yourself. You could make things worse.<br><br>Workaround (if possible): Ask the help desk if there's a temporary workaround. (e.g., "Is there another server I can use?" "Can I access the data another way?")<br><br>Level 2: Basic IT Support / First Responder (You can follow documented procedures) | "It's the database server for the sales application.")<br><br>Document the Impact: How is this outage affecting your work? This helps prioritize the fix.<br><br>Follow Up: If you don't hear back within a reasonable timeframe (defined by your organization's SLAs), follow up with the support team.<br><br>Do NOT attempt to fix the server yourself.<br><br>2. Help Desk / Tier 1 Support<br><br>Verify the Issue: Confirm that Server 003 is indeed down. Use basic monitoring tools (if available) or try to access services hosted on the server.<br><br>Check the Knowledge Base: See if there's a known issue or workaround documented.<br><br>Follow Escalation Procedures: If you can't resolve the issue, escalate it to Tier 2 support (System Administrators/Engineer s). Provide them with:<br><br>Detailed description of the problem. | Submit a ticket with as much detail as you can (see "Gather Information" above).<br><br>Contact the Help Desk/IT Support: Call or email them.<br><br>Inform your manager: Let them know you're experiencing an issue and have reported it.<br><br>Document the impact: Explain how the outage is affecting your work. This helps prioritize the fix.<br><br>Be patient: IT teams are often dealing with multiple issues.<br><br>Ask for updates: Follow up on your ticket periodically to get an estimated time to resolution.<br><br>Work on alternative tasks: If possible, switch to tasks that don't rely on Server 003.<br><br>II. Basic Technical User (You have some IT knowledge, maybe you're a power user, but not a system administrator)<br><br>Follow the "Gather Information" steps above. This is still crucial.<br><br>Check basic |

| | | | | |
|---|---|---|---|---|
| | | graphs leading up to the outage. Did CPU, memory, or disk usage spike?<br><br>Phase 2: Application Checks (You Can Do These - Priority 2 - If Ping Works)<br><br>If the server responds to pings, the problem is likely with the application(s) running on it.<br><br>Try to Access the Application: Attempt to use the service that Server 003 provides. (e.g., open the website, connect to the database, access the shared folder). Note the exact error message.<br><br>Check Application Logs: (You'll need to know where the logs are located. This is often in /var/log on Linux or C:\ProgramData\<Application>\Logs on Windows). Look for error messages around the time the problem started.<br><br>Restart the Application Service: (If you have the permissions and know how). This is often the quickest fix for minor application glitches. Be careful! Restarting the wrong service can cause more problems. (See "Important | Check the Status Page/Incident Management System: As above, see if it's already known. Update the incident if you're the first to report it.<br><br>Follow Runbooks/Knowledge Base: Your organization should have documented procedures for common issues. Look for a runbook specifically for Server 003 or for general server outages. These might include:<br><br>Simple Restart: (If authorized and the runbook allows) A simple reboot can sometimes resolve temporary issues. Be cautious!<br><br>Service Restart: Restarting a specific service on the server might be sufficient.<br><br>Escalate: If you can't resolve the issue using runbooks, escalate to Tier 2 support (see below). Provide detailed information:<br><br>Ping results<br><br>Service check results<br><br>Error messages<br><br>Steps you've already taken | You have some IT knowledge and are often the first line of defense.<br><br>Check Monitoring Systems: Look at your monitoring tools (e.g., Nagios, Zabbix, SolarWinds, PRTG, cloud provider dashboards) to see if the server is flagged as down, and for any related alerts. These alerts often provide clues.<br><br>Check Basic Connectivity:<br><br>Ping: As mentioned above, ping Server 003.<br><br>Traceroute/Tracert: Run a traceroute ( traceroute Server003 on Linux/macOS, tracert Server003 on Windows) to see where the connection is failing. This can help identify network issues.<br><br>Check Server Console (if possible): If you have access to the server's console (e.g., through a remote management card like iLO, iDRAC, or IPMI, or physically at the server), check the screen for any error messages or a frozen screen.<br><br>Check Basic Services: If you know what services Server 003 provides, | Error messages.<br><br>Steps to reproduce the issue.<br><br>Impacted users/services.<br><br>Any troubleshooting steps you've already taken.<br><br>Keep Users Informed: Provide updates to users who have reported the issue. Even a simple "We're working on it" message is helpful.<br><br>Create a Ticket: Ensure a ticket is created in the incident management system.<br><br>3. System Administrator / Engineer (Tier 2/3 Support - You have access to the server)<br><br>This is where things get more technical. Follow these steps in a logical order:<br><br>Confirm the Outage (More Thoroughly):<br><br>Ping: ping Server003 (Can you reach it at all?)<br><br>Remote Access: Try to connect via SSH (Linux) or RDP (Windows).<br><br>Monitoring Tools: Check your monitoring system (e.g., Nagios, Zabbix, | connectivity:<br><br>Ping Server 003: Open a command prompt (Windows) or terminal (Mac/Linux) and type ping Server003.<br><br>If the ping fails ("Request timed out" or "Destination host unreachable"): This confirms the server is unreachable from your machine. It doesn't necessarily mean the server is down, but it's a strong indicator. Proceed to reporting the issue.<br><br>If the ping succeeds: The server is responding to basic network requests. The problem might be with a specific application or service on the server. Report the issue, but mention that ping is working.<br><br>Try accessing a known service on Server 003: If you know a specific website, application, or shared folder hosted on Server 003, try to access it. The error message you get will be helpful for IT.<br><br>Check network connections: Make sure your own network connection is working (Wi-Fi, Ethernet cable).<br><br>Check recent changes: |

| | | Considerations" below).<br><br>Check Resource Usage (If Possible): If you have access to a remote management tool (like SSH or RDP - see Phase 3), check CPU, memory, and disk usage on the server. High resource usage can cause applications to become unresponsive.<br><br>Phase 3: Server Access & Deeper Troubleshooting (You May Need Help/Permissions - Priority 3)<br><br>These steps require you to log in to the server directly, which may require special permissions. Proceed with caution!<br><br>Remote Access:<br><br>SSH (Linux): ssh username@003<br><br>RDP (Windows): Use Remote Desktop Connection.<br><br>Console Access: If remote access isn't working, you may need physical access to the server and connect a monitor and keyboard.<br><br>Check System Logs:<br><br>Linux: /var/log/syslog, /var/log/messages, | Time of initial report<br><br>3. Tier 2 Support / System Administrator / DevOps Engineer (You have more access and responsibility)<br><br>Confirm Outage & Gather Information (See "Gathering Information" section below - very important)<br><br>Check Monitoring Systems: (e.g., Nagios, Zabbix, Prometheus, Datadog, New Relic). These will provide detailed information about the server's health:<br><br>CPU Usage: Is it pegged at 100%?<br><br>Memory Usage: Is it running out of memory?<br><br>Disk Space: Is the disk full?<br><br>Network Traffic: Is there unusual network activity?<br><br>Service Status: Are key services running?<br><br>Logs: Check system logs (syslog, event logs) and application logs for errors.<br><br>Attempt Basic Troubleshooting: | check if those services are responding on other servers (if there's redundancy).<br><br>Follow Runbooks/Standard Operating Procedures (SOPs): Your organization should have documented procedures for common issues. Follow those procedures exactly. This is the most important thing. Look for a runbook specifically for Server 003 or for the service it provides.<br><br>Simple Restarts (with caution!): Only if the runbook allows it, and you understand the implications, you might try a simple restart of the server. Before restarting, consider:<br><br>Impact: What will happen if you restart the server? Who will be affected?<br><br>Dependencies: Does this server depend on other servers? Will restarting it cause a cascading failure?<br><br>Data Loss: Is there a risk of data loss?<br><br>Escalate if Necessary: If you can't resolve the issue using documented procedures, or if you're unsure about anything, | Prometheus, Datadog) for alerts and detailed metrics (CPU, memory, disk I/O, network traffic).<br><br>Check Basic Server Health (If you can access it):<br><br>CPU Usage: Is it pegged at 100%?<br><br>Memory Usage: Is it exhausted?<br><br>Disk Space: Is the disk full?<br><br>Network Connectivity: Check network interfaces and routing.<br><br>Check Recent Changes: This is crucial!<br><br>Deployment Logs: Were any applications or updates deployed recently?<br><br>Configuration Management: Were any configuration changes made (e.g., using Ansible, Puppet, Chef)?<br><br>Patching: Was the server recently patched?<br><br>Check System Logs: This is where you'll find clues.<br><br>Linux: /var/log/syslog, /var/log/messages, /var/log/auth.log, application-specific logs (e.g., | Were there any recent software updates or configuration changes on your machine that might be interfering?<br><br>Report the issue: Use your company's IT support procedures (ticketing system, help desk). Include all the information you've gathered, including ping results and any error messages.<br><br>Do NOT attempt to restart the server or modify any server settings. This is almost certainly outside your scope of responsibility.<br><br>III. System Administrator / IT Professional (You're responsible for managing servers)<br><br>This is where things get more involved. The steps depend heavily on your environment and monitoring tools.<br><br>Acknowledge the alert: If you're receiving alerts from a monitoring system (Nagios, Zabbix, SolarWinds, etc.), acknowledge the alert to prevent duplicate notifications.<br><br>Check Monitoring Tools:<br><br>CPU Usage: Is the server maxing out its CPU? |
|---|---|---|---|---|

| | /var/log/kern.log<br><br>Windows: Event Viewer (System and Application logs) Look for errors and warnings around the time the server went down.<br><br>Check Running Processes:<br><br>Linux: top, htop, ps aux<br><br>Windows: Task Manager Look for processes that are consuming excessive resources or are in an unexpected state.<br><br>Check Disk Space: Use df -h (Linux) or check disk properties in Windows Explorer. A full disk can cause many problems.<br><br>Check Network Configuration:<br><br>Linux: ip addr, route -n, netstat -tulnp<br><br>Windows: ipconfig /all, route print Verify that the server has a valid IP address, subnet mask, gateway, and DNS settings.<br><br>Restart the Server (Last Resort!) Only restart the server if you've exhausted troubleshooting steps and have permission to do so. A restart will | Remote Access: Can you SSH or RDP into the server? If not, it's a network or server-level issue.<br><br>Console Access: If remote access fails, try to access the server via a console (e.g., IPMI, iLO, DRAC). This gives you direct access even if the network is down.<br><br>Service Restarts: Try restarting critical services.<br><br>Resource Checks: Use tools like top, htop, free -m, df -h to check CPU, memory, and disk usage.<br><br>Escalate (If Necessary): If you can't identify the problem or resolve it quickly, escalate to Tier 3 support or the on-call engineer. Provide all the information you've gathered.<br><br>4. Tier 3 Support / Senior Engineer / On-Call Engineer (Deepest level of access and responsibility)<br><br>All of the above, plus:<br><br>Root Cause Analysis: Focus on identifying the underlying cause of the outage.<br><br>Advanced Troubleshooting: This | escalate to Level 3 support immediately. Don't spend hours troubleshooting something you're not qualified to fix.<br><br>Level 3: System Administrator / Experienced IT Professional (You manage the server)<br><br>You have deep knowledge of the server and its applications.<br><br>All of Level 2, plus:<br><br>Check Server Logs: Examine the server's system logs (e.g., Event Viewer on Windows, /var/log/syslog and /var/log/messages on Linux) for errors or warnings. Also check application-specific logs.<br><br>Check Resource Usage: Use tools like Task Manager (Windows) or top/htop (Linux) to check CPU, memory, disk I/O, and network usage. The server might be overloaded.<br><br>Check Disk Space: Make sure the server isn't out of disk space. Full disks can cause all sorts of problems.<br><br>Check Network Configuration: Verify the server's IP address, subnet mask, gateway, | /var/log/apache2/error.lo g, /var/log/mysql/error.log)<br><br>Windows: Event Viewer (System, Application, Security logs)<br><br>Restart Services: If a specific service is failing, try restarting it. (e.g., systemctl restart apache2 on Linux, or restart the service in Windows Services Manager).<br><br>Restart the Server (As a Last Resort): Only do this if other troubleshooting steps have failed and you've considered the impact. A graceful shutdown is preferred, but sometimes a hard reboot is necessary. Document why you restarted the server.<br><br>Escalate if Necessary: If you can't resolve the issue, escalate to Tier 3 support (Senior Engineers, DevOps, or the vendor if it's a cloud service).<br><br>4. DevOps / SRE (Highly Automated Environments)<br><br>Automated Alerts: You should already be alerted by your monitoring and alerting systems. | Memory Usage: Is the server running out of memory?<br><br>Disk Space: Is the disk full?<br><br>Network Traffic: Is there unusual network activity?<br><br>Service Status: Are the critical services running on Server 003 (e.g., web server, database server) stopped?<br><br>Remote Access: Attempt to connect to the server remotely (SSH, RDP, etc.).<br><br>If you can connect: Investigate the specific service that's failing. Check logs (see below).<br><br>If you cannot connect: Proceed to the next steps.<br><br>Check Server Logs: This is critical. Look for error messages in:<br><br>System Logs: (Event Viewer on Windows, /var/log/syslog or /var/log/messages on Linux)<br><br>Application Logs: (Specific to the applications running on the server)<br><br>Security Logs: (Look for |
|---|---|---|---|---|

| | | | | | |
|---|---|---|---|---|---|
| | | interrupt service and may cause data loss if not done properly. Document everything before restarting!<br><br>Phase 4: Escalation (If You Can't Resolve It)<br><br>If you've tried the above steps and can't get Server 003 back online, escalate the issue to the appropriate team or person. This could be:<br><br>Your IT Support Team<br><br>The Server Administrator<br><br>The Network Administrator<br><br>The Database Administrator (if it's a database server)<br><br>When escalating, provide all the information you've gathered:<br><br>What is Server 003's function?<br><br>What are the symptoms?<br><br>When did it go down?<br><br>What troubleshooting steps have you already taken?<br><br>Any error messages you've encountered. | might involve:<br><br>Debugging: Using debuggers to analyze application code.<br><br>Packet Capture: Using tools like tcpdump or Wireshark to capture network traffic.<br><br>Core Dumps: Analyzing core dumps to understand application crashes.<br><br>Database Analysis: Checking database logs and performance metrics.<br><br>Implement Fixes: Apply patches, roll back changes, or make other necessary changes to restore service.<br><br>Post-Incident Review: After the outage is resolved, conduct a post-incident review to identify lessons learned and prevent future occurrences.<br><br>Gathering Information (Critical for Troubleshooting)<br><br>This is what you need to collect before escalating or attempting complex fixes. The more information you have, the faster the problem can be resolved.<br><br>Server Details: | and DNS settings.<br><br>Check Firewall Rules: Ensure that firewall rules aren't blocking necessary traffic.<br><br>Check Application Status: Verify that the applications running on the server are started and functioning correctly. Use application-specific tools to check their status.<br><br>Review Recent Changes: What changes were made to the server recently? (e.g., software updates, configuration changes, new deployments). Recent changes are often the cause of problems.<br><br>Consider Rollback: If a recent change is suspected, consider rolling back to a previous known-good configuration.<br><br>Advanced Troubleshooting: Use tools like process monitors, network analyzers (Wireshark), and debuggers to diagnose the problem.<br><br>Contact Vendor Support: If the issue is related to a specific application or hardware component, contact the vendor's support team. | Run Playbooks: Execute pre-defined runbooks for common issues. These might include:<br><br>Service Restart: Automatically restart failing services.<br><br>Scale Out: If the server is overloaded, automatically scale out to additional instances.<br><br>Rollback: If a recent deployment caused the issue, automatically roll back to the previous version.<br><br>Investigate Metrics: Use your monitoring dashboards to identify the root cause.<br><br>Automated Remediation: Implement automated remediation steps to prevent future outages.<br><br>Post-Incident Review: Conduct a post-incident review to identify lessons learned and improve the system.<br><br>Gathering Information for Diagnosis (Important for all technical roles):<br><br>Timestamp of the Outage: When did the problem start?<br><br>Error Messages: Exact error messages are invaluable. | failed login attempts or other security-related events)<br><br>Basic Troubleshooting (Proceed with caution!):<br><br>Restart the affected service: If a specific service is failing, try restarting it.<br><br>Restart the server: Only if you've investigated the issue and believe a restart is safe. Schedule a restart if possible to minimize disruption. Consider the impact before restarting!<br><br>Check hardware: If you suspect a hardware issue (e.g., disk failure), investigate the hardware health.<br><br>Escalation: If you're unable to resolve the issue within a reasonable timeframe, or if you suspect a serious problem (hardware failure, security breach), escalate to a senior administrator or the appropriate team.<br><br>Documentation: Document everything you do during the troubleshooting process. This is essential for post-incident analysis.<br><br>Post-Incident Review: After the issue is |

| | | | | | |
|---|---|---|---|---|---|
| | Important Considerations:<br><br>Permissions: Don't attempt any troubleshooting steps that you don't have the necessary permissions to perform. You could make the problem worse or violate security policies.<br><br>Documentation: Document everything you do, including the time, the steps you took, and the results. This will be helpful for the team that escalates the issue.<br><br>Change Management: If the outage occurred after a recent change, be sure to mention that when escalating the issue.<br><br>Don't Panic: Stay calm and methodical. Rushing can lead to mistakes.<br><br>Backups: If data loss is a concern, verify that you have recent backups of the data on Server 003.<br><br>Application-Specific Knowledge: Some applications have specific troubleshooting steps. Consult the application's documentation or support resources.<br><br>Restarting | Server Name: Server 003 (you have this)<br><br>IP Address: (Important for network checks)<br><br>Operating System: (e.g., Linux, Windows Server)<br><br>Role/Purpose: (What applications or services does it run?) Crucial!<br><br>Location: (Physical location of the server - datacenter, cloud region)<br><br>Time of Outage: Precise time the issue was first detected.<br><br>Recent Changes: This is extremely important! Were any changes made to Server 003 or related systems recently? (e.g., software updates, configuration changes, network changes). Check:<br><br>Change Management System: (If your organization uses one)<br><br>Version Control System: (If configuration is managed in Git or similar)<br><br>Deployment Logs:<br><br>Error Messages: Exact error messages from any applications or systems. | Document Everything: Keep a detailed record of your troubleshooting steps and findings. This will be helpful for future incidents.<br><br>Important Considerations:<br><br>Change Management: Was a change recently made to the server or its environment? If so, that's the first place to look.<br><br>Backups: Are backups current and restorable? (This is a good practice to verify regularly, not just when a server is down.)<br><br>Disaster Recovery Plan: Does your organization have a disaster recovery plan for Server 003? If so, follow it.<br><br>Communication: Keep stakeholders informed of the status of the outage and your troubleshooting efforts.<br><br>When to Escalate (Regardless of your level):<br><br>You're unsure what to do.<br><br>You've followed documented procedures and the issue persists.<br><br>The issue is critical and | Recent Changes: What changed on the server or related systems recently?<br><br>Server Specifications: CPU, memory, disk space, operating system.<br><br>Network Configuration: IP address, subnet mask, gateway, DNS servers.<br><br>Services Running: What applications and services are hosted on the server?<br><br>Monitoring Data: CPU usage, memory usage, disk I/O, network traffic, application metrics.<br><br>Logs: System logs, application logs, security logs.<br><br>To help me give you more specific advice, please tell me:<br><br>What is your role? (e.g., End User, Help Desk, System Admin, DevOps)<br><br>What is Server 003 used for? (e.g., Database server, Web server, Application server)<br><br>What operating system is Server 003 running? (e.g., Linux, Windows)<br><br>What error messages are you seeing (if any)? | resolved, conduct a post-incident review to identify the root cause and prevent it from happening again.<br><br>Important Considerations:<br><br>Change Management: Was there a recent change that might have caused the outage? If so, roll back the change if possible.<br><br>Backups: Ensure you have recent backups of the server.<br><br>Disaster Recovery Plan: If Server 003 is critical, you should have a disaster recovery plan in place.<br><br>Security: Be aware of potential security implications. If you suspect a security breach, follow your company's security incident response plan.<br><br>To help me give you more specific advice, please tell me:<br><br>What is your role? (e.g., end-user, help desk, system administrator)<br><br>What is Server 003's purpose? (e.g., web server, database server, file server)<br><br>What operating system |

| | | | Services/Servers: Be very careful when restarting services or servers. Understand the potential impact before you do so. If you're unsure, don't restart anything.<br><br>To help me give you more specific advice, please tell me:<br><br>What is Server 003 used for? (e.g., web server, database server, file server)<br><br>What operating system is it running? (e.g., Windows Server, Linux)<br><br>What is your role and what level of access do you have to the server? (e.g., end-user, help desk, system administrator)<br><br>What are the exact symptoms you're seeing? (e.g., can't ping, application errors, etc.)<br><br>Good luck! Let me know how it goes, and I'll do my best to help. | Logs:<br><br>System Logs: (e.g., /var/log/syslog, /var/log/messages, Windows Event Logs)<br><br>Application Logs: Logs specific to the applications running on the server.<br><br>Network Logs: Firewall logs, router logs.<br><br>Monitoring Data: Screenshots or data from your monitoring systems (CPU, memory, disk, network).<br><br>Ping/Traceroute Results: From multiple locations if possible.<br><br>Service Status: Output of commands to check service status (e.g., systemctl status <service>, service <service> status).<br><br>To help me give you more specific advice, please tell me:<br><br>What is your role? (e.g., End User, Help Desk, System Admin)<br><br>What is Server 003 used for? (e.g., Database server, Web server, Application server)<br><br>What are you experiencing? (e.g., Can't access a website, | affecting a large number of users.<br><br>You suspect a hardware failure.<br><br>You suspect a security breach.<br><br>You're outside of your area of expertise.<br><br>To help me give you more specific advice, please tell me:<br><br>What is your role? (e.g., end-user, help desk, system administrator)<br><br>What is Server 003's purpose?<br><br>What are you seeing when you try to access it? (Error messages, no response, etc.)<br><br>What operating system is Server 003 running? (Windows Server, Linux, etc.)<br><br>Good luck! Let me know if you have any more information, and I'll do my best to help. | What have you already tried?<br><br>Disclaimer: This is a general guide. Your organization may have specific procedures for handling server outages. Always follow your organization's policies and procedures. | is Server 003 running? (e.g., Windows Server, Linux)<br><br>What have you already tried?<br><br>Are there any error messages? (Please provide the exact text.)<br><br>Disclaimer: I am an AI chatbot and cannot provide definitive IT support. This information is for general guidance only. Always follow your company's policies and procedures. If you are unsure about anything, escalate to someone more experienced. |

| | | | Application is crashing, Receiving error messages)<br><br>What have you already tried?<br><br>Good luck! Let me know if you have any other questions. | | | |
|---|---|---|---|---|---|---|
| Act as a SAP Monitoring Expert. What do I do if Server 003 is down? | Test 2 | Okay, you're facing a downed SAP server (003). As a SAP Monitoring Expert, here's a breakdown of how to approach this, categorized by phases. This is a comprehensive guide, so adjust based on your specific environment and responsibilities. I'll cover initial response, diagnosis, recovery, and preventative measures. I'll also indicate priority levels (High, Medium, Low) for each step.<br><br>Important Assumptions:<br><br>You have access to SAP monitoring tools: Solution Manager (SolMan), SAP Cloud Monitoring, or other third-party tools.<br><br>You have basic SAP system knowledge: Understanding of three-tier architecture (Presentation, Application, Database).<br><br>You have defined escalation procedures: Know who to contact if you can't resolve the issue.<br><br>Server 003's Role: I'm assuming you know what Server 003 does (e.g., Application Server, Database Server, Central Services). This is critical for | Okay, you're facing a downed SAP server (003). As a SAP Monitoring Expert, here's a breakdown of how to approach this, categorized by phases. This is a comprehensive guide, so adjust based on your specific environment and responsibilities. I'll cover initial response, diagnosis, recovery, and post-incident activities. I'll also indicate priority levels (High, Medium, Low) for each step.<br><br>Important Assumptions:<br><br>You have access to SAP Solution Manager (SolMan) or a similar monitoring tool. This is critical. If not, the diagnosis will be much harder.<br><br>You have basic understanding of your SAP landscape. (e.g., what applications run on server 003, its dependencies).<br><br>You have appropriate authorizations to perform these tasks.<br><br>Phase 1: Immediate Response & Initial Assessment (High Priority - within 5-15 minutes)<br><br>Confirm the Outage: (High) | Okay, you're facing a downed SAP server (003). As a SAP Monitoring Expert, here's a breakdown of how to approach this, categorized by phases. This is a comprehensive guide, so adjust based on your specific environment and responsibilities. I'll cover initial response, diagnosis, recovery, and preventative measures. I'll also indicate priority levels (High, Medium, Low) for each step.<br><br>Important Assumptions:<br><br>You have access to SAP monitoring tools: Solution Manager (SolMan), SAP Cloud Monitoring, or other third-party tools.<br><br>You have basic understanding of your SAP landscape: What applications run on server 003, its dependencies, and your company's incident management process.<br><br>You have appropriate authorizations: To access servers, SAP systems, and monitoring tools.<br><br>Phase 1: Immediate Response & Initial Assessment (High Priority - within 5-15 minutes)<br><br>Confirm the Outage: (High) | Okay, you're facing a downed SAP server (003). As a SAP Monitoring Expert, here's a breakdown of how to approach this, categorized by phases. This is a comprehensive guide, so adjust based on your specific environment and responsibilities. I'll cover initial response, diagnosis, recovery, and post-incident activities. I'll also indicate priority levels (High, Medium, Low) for each step.<br><br>Important Assumptions:<br><br>You have access to SAP Solution Manager (SolMan) or a similar monitoring tool. This is critical. If not, the diagnosis will be much harder.<br><br>You have basic understanding of your SAP landscape. (e.g., what applications run on server 003, its dependencies).<br><br>You have appropriate authorizations to perform these tasks.<br><br>Phase 1: Immediate Response & Verification (High Priority - within 5-15 minutes)<br><br>Confirm the Outage: (High) | Ah, the dreaded "Server 003 is down" scenario! Don't panic, we can troubleshoot this. To help me guide you effectively, please provide more information:<br><br>**1. What kind of system is Server 003?**<br><br>* **Application server:** Running SAP applications like ECC, S/4HANA, etc.?<br>* **Database server:** Hosting your SAP database (e.g., Oracle, SQL Server)?<br>* **Other infrastructure:** File server, message queue, etc.?<br><br>**2. What exactly do you mean by "down"?**<br><br>* **Completely inaccessible?** Can't ping it or reach any services?<br>* **Application unavailable?** Users can't log in or access functions?<br>* **Performance issues?** Slow response times, errors occurring frequently?<br><br>**3. Have you checked the system logs on Server 003 and other related servers?**<br><br>* Look for error messages, warnings, or |

| | | | | |
|---|---|---|---|---|
| | prioritization. I'll cover scenarios for different roles.<br><br>Phase 1: Immediate Response & Verification (High Priority - within 5-15 minutes)<br><br>Confirm the Outage (High):<br><br>Monitoring Tools: Check your primary monitoring tool (SolMan, Cloud Monitoring, etc.). Look for alerts specifically related to Server 003. Verify the alert isn't a false positive (sometimes network hiccups trigger alerts).<br><br>Ping Test: From another server in the SAP landscape, ping Server 003. No response confirms network connectivity issues.<br><br>SAP GUI Connection: Attempt to connect to the SAP system via SAP GUI, specifying Server 003 as the application server. Failure to connect confirms the issue.<br><br>OS Level Check (if possible remotely): If you have remote OS access (SSH, RDP), try to connect to the server.<br><br>Initial Communication (High): | SolMan/Monitoring Tool: Verify the server is truly down. Don't rely on a single alert. Check multiple sources. Look for:<br><br>Host Unreachable alerts<br><br>SAP Instance status as "Stopped" or "Unreachable"<br><br>Ping failures<br><br>Database connection failures (if applicable to this server)<br><br>Direct Access (if possible): Attempt to SSH/RDP to the server. If you can't connect, it reinforces the outage.<br><br>Check System Logs (Initial): Quickly scan the OS level logs (e.g., /var/log/messages on Linux, Event Viewer on Windows) for obvious errors before diving into SAP logs. Look for things like kernel panics, disk errors, or network issues.<br><br>Communication: (High)<br><br>Incident Creation: Immediately create a high-priority incident in your ITSM system (e.g., ServiceNow, Remedy). Include:<br><br>Server Name (003) | Confirm the Outage: (High)<br><br>Monitoring Tools: Verify the outage in your primary monitoring tool (SolMan, Cloud Monitoring, etc.). Don't rely solely on one source. Look for alerts, status indicators (red/yellow/green), and historical data.<br><br>Ping Test: Simple ping test to the server's IP address. If it fails, it confirms network connectivity issues.<br><br>Check System Logs (if accessible remotely): Even a quick glance at the OS logs (Windows Event Viewer, Linux syslog) might reveal immediate clues (e.g., kernel panic, disk full).<br><br>Incident Creation: (High)<br><br>Log an Incident: Immediately create a high-priority incident in your ITSM system (ServiceNow, Remedy, etc.). Include:<br><br>Server Name: 003<br><br>Brief Description: "Server 003 is down - impacting [affected applications/services]"<br><br>Impact: Clearly state the business impact. (e.g., | SolMan/Monitoring Tool: Verify the alert isn't a false positive. Check the status of the server in SolMan. Look for detailed error messages. Is it completely down (no ping response) or just SAP instances not running?<br><br>Ping Test: From a known good server (e.g., your monitoring server, a database server), ping server 003. No ping = network or OS issue.<br><br>Check OS Level: If ping works, try to SSH/RDP to the server. If you can't connect, it's likely an OS-level issue.<br><br>Document: Immediately log the outage in your incident management system (e.g., ServiceNow, Jira, HP Service Manager). Include timestamp, initial observations.<br><br>Initial Communication: (High)<br><br>Alert Stakeholders: Notify the relevant teams: Basis team, Application team (if you know what application is impacted), and potentially business stakeholders (depending on severity and SLA). Use your established communication channels (e.g., Slack, email, | any clues about what might have caused the issue.<br><br>**4. What is your current SAP monitoring setup?**<br><br>* Are you using tools like SMon, RZMON, STMS, etc.?<br>* Which specific alerts or notifications are triggered by this situation?<br><br>Once I have a clearer picture of the problem, I can provide more targeted advice. Here are some general troubleshooting steps:<br><br>* **Basic checks:** Verify network connectivity, check physical connections and power supplies.<br>* **Restart services:** Try restarting the affected SAP instances or relevant system services.<br>* **Check hardware resources:** Monitor CPU usage, memory consumption, and disk space on Server 003 and connected systems.<br>* **Review configuration files:** Ensure settings are correct and consistent across all relevant servers.<br>* **Consult SAP documentation:** Search for known issues or solutions related to |

| | | | | |
|---|---|---|---|---|
| Alert Stakeholders: Immediately notify the relevant teams:<br><br>Basis Team: This is your primary team.<br><br>Database Team (if Server 003 is the database server): Critical!<br><br>Application Team (if Server 003 hosts critical application components): They need to understand the impact.<br><br>Service Desk/Help Desk: They'll likely receive user reports.<br><br>Create an Incident: Log the incident in your ITSM system (ServiceNow, Remedy, etc.). Include:<br><br>Server Name: 003<br><br>Time of Outage<br><br>Initial Observations (ping failed, SAP GUI connection failed, etc.)<br><br>Impact Assessment (initial guess - see Phase 2)<br><br>Check for Planned Maintenance (High):<br><br>Maintenance Calendar: Verify if Server 003 was scheduled for maintenance. If so, the | Time of Outage Detection<br><br>Initial Symptoms<br><br>Impact (see step 3)<br><br>Stakeholder Notification: Notify the relevant stakeholders immediately. This includes:<br><br>SAP Basis Team<br><br>Application Owners (who rely on services from this server)<br><br>IT Operations/Infrastructure Team<br><br>Management (as per your escalation procedures) A quick email/chat message is sufficient initially.<br><br>Impact Assessment: (High)<br><br>Identify Affected Applications/Services: What SAP applications or services are running on server 003? (e.g., Development system, Production system, specific modules like FI/CO, BW, etc.).<br><br>Business Impact: What is the impact of this outage on the business? (e.g., Order processing stopped, reporting | "Production Order Processing halted," "Reporting unavailable," "Critical Batch Job failed"). This is crucial for prioritization.<br><br>Affected Users/Teams: Identify who is impacted.<br><br>Initial Observations: What you've already checked.<br><br>Communication: (High)<br><br>Notify On-Call Team: Alert the appropriate on-call team (Basis, OS Admin, Database Admin, Application Support) based on your escalation matrix. Use your standard communication channels (phone, email, messaging app).<br><br>Stakeholder Notification (Initial): Inform key stakeholders (IT Manager, Business Owner) about the outage and that investigation is underway. Keep it concise.<br><br>Phase 2: Diagnosis (High to Medium Priority - 15 minutes - 1 hour, depending on complexity)<br><br>This is where you determine why the server is down. | phone bridge). Keep the initial communication concise: "Server 003 is down. Investigating."<br><br>Check for Planned Maintenance: Before diving deep, double-check the maintenance calendar. It's embarrassing to spend an hour troubleshooting a planned outage.<br><br>Phase 2: Diagnosis (High to Medium Priority - 15 minutes - 1 hour, depending on complexity)<br><br>This is where you determine why the server is down. The approach depends on what you found in Phase 1.<br><br>Scenario A: Server Unreachable (No Ping)<br><br>Network Check: (High) Involve the network team. Is there a network outage affecting server 003? Check network device logs (routers, switches).<br><br>Hardware Check: (Medium) If network is okay, suspect hardware. If you have access to the server console (e.g., via iLO, iDRAC), check the server's health status. Look for hardware errors (CPU, | your specific system setup.<br><br>Remember, the best approach is to gather as much information as possible before taking any actions. Let's get Server 003 back online! |
| | outage is expected. Confirm the maintenance window is still valid.<br><br>Phase 2: Diagnosis & Impact Assessment (High/Medium Priority - within 30-60 minutes)<br><br>This phase depends heavily on what Server 003 is. I'll break it down by common roles:<br><br>Scenario A: Server 003 is an Application Server:<br><br>Check Work Process Status: Using transaction SM50 (Work Process Overview) on another application server, see if any work processes are stuck or in a bad state on Server 003 (even if you can't connect directly). This might indicate a program error.<br><br>Check System Log (SM21): On another application server, check the system log for errors related to Server 003. Look for messages indicating crashes, resource issues, or communication problems.<br><br>Impact Assessment: Which application components were running on Server 003? What business processes are affected? | unavailable, critical batch jobs failing). This will drive the urgency of the recovery. Quantify the impact if possible (e.g., "Estimated revenue loss of $X per hour").<br><br>Dependency Mapping: Understand what other systems depend on server 003. Is it a database server? An application server? A central services instance?<br><br>Phase 2: Diagnosis (Medium to High Priority - 15 minutes - 2 hours, depending on complexity)<br><br>SAP System Logs: (High)<br><br>SM21 (System Log): If you can access the SAP system (even if it's slow), check SM21 for errors leading up to the outage. Look for:<br><br>Short dumps<br><br>Database errors<br><br>Resource contention (CPU, memory)<br><br>Errors related to specific SAP components<br><br>Dev Trace Files:* Check the developer traces (transaction ST22) for recent dumps. | OS Level Checks: (High)<br><br>Remote Console Access (if possible): Use iLO, iDRAC, VNC, or similar to access the server's console. This is the most valuable way to see what's happening.<br><br>Check OS Status: Is the OS running? Is it hung? Are there any error messages on the console?<br><br>Resource Utilization: If the OS is up but unresponsive, check CPU, Memory, Disk I/O, and Network utilization. High utilization can indicate a bottleneck.<br><br>Disk Space: Critical! A full disk can bring everything to a halt.<br><br>Recent Changes: Were any OS patches, updates, or configuration changes applied recently? This is a prime suspect.<br><br>SAP Level Checks (if OS is up, but SAP is down): (Medium)<br><br>SAP System Status (SM51): If you can access the SAP GUI, check SM51 to see the status of work processes. Are they all stopped? Are there any errors? | memory, disk).<br><br>Power Supply: (Medium) Verify the server has power. (Sounds basic, but it happens!)<br><br>Scenario B: Server Reachable (Ping Works), SAP Instances Down<br><br>OS Level Checks: (High)<br><br>CPU/Memory Usage: Log into the server (SSH/RDP). Check CPU and memory usage. Is the server overloaded? Use tools like top, htop, vmstat.<br><br>Disk Space: (High) Check disk space on all relevant partitions (especially /, /usr/sap, /var/log). A full disk can crash SAP. Use df -h.<br><br>OS Logs: (High) Examine the OS logs (/var/log/messages, /var/log/syslog, Windows Event Viewer). Look for errors related to SAP, the OS, or hardware.<br><br>SAP Instance Status: (High)<br><br>SAPSTART/SAPSTOP: Check if sapstart or sapstop processes are running. If sapstop is running, it might be a controlled shutdown.<br><br>Instance Profiles: Verify | |

| | | | | |
|---|---|---|---|---|
| | (e.g., SD, MM, FI). Estimate the number of users impacted.<br><br>Scenario B: Server 003 is the Database Server:<br><br>Database Monitoring: Use your database monitoring tools (e.g., SAP HANA Studio, Oracle Enterprise Manager, SQL Server Management Studio) to check the database status. Look for errors, high CPU usage, disk space issues, or network connectivity problems.<br><br>Database Alert Log: Examine the database alert log for critical errors.<br><br>Impact Assessment: This is critical. A database outage impacts all SAP applications. The impact is usually severe.<br><br>Scenario C: Server 003 is a Central Services Instance (ASCS/ERS):<br><br>Check ICM Status: Use transaction SMICM on another central instance to check the ICM (Internet Communication Manager) status on Server 003. ICM handles communication between application servers.<br><br>Check Message Server | SAP Application Logs: Check application-specific logs (e.g., SLG1 for ABAP applications, BW logs, etc.) for errors.<br><br>Operating System Level Investigation: (Medium)<br><br>Resource Monitoring: Check CPU, memory, disk I/O, and network utilization before the outage (if historical data is available). This can reveal resource exhaustion. Tools:<br><br>Linux: top, vmstat, iostat, netstat<br><br>Windows: Task Manager, Performance Monitor<br><br>Disk Space: Verify sufficient disk space on all relevant partitions (e.g., /, /usr/sap, /var/log). A full disk can cause severe issues.<br><br>Network Connectivity: Verify network connectivity to the server. Check firewall rules, DNS resolution, and network interfaces. ping, traceroute, nslookup are your friends.<br><br>OS Logs (Detailed): Examine the OS logs in detail. Look for: | SAP System Log (SM21): Look for errors or warnings in the SAP system log that might indicate the cause of the problem. Filter by time to focus on events around the outage.<br><br>Database Connection: Verify that the SAP system can connect to the database. Use DB02 or DB13.<br><br>Message Server Status: Check the Message Server status. If the Message Server is down, it can bring down the entire system.<br><br>ICM Status (SMICM): If the issue is related to web services or RFC connections, check the ICM status.<br><br>Network Checks: (Medium)<br><br>Network Connectivity: Verify network connectivity between the server and other critical systems (database server, other application servers, load balancer). Use traceroute or pathping.<br><br>Firewall Rules: Ensure that firewall rules are not blocking traffic to or from the server.<br><br>DNS Resolution: Verify that the server's | the instance profiles are correct and accessible.<br><br>Dev Trace Files:* (Medium) Check the dev_* trace files in the SAP instance directory (/usr/sap/<SID>/<instance_number>/trace). These often contain detailed error messages.<br><br>Database Connectivity: (Medium)<br><br>Database Status: Is the database server up and running? Can the SAP server connect to the database? Test with sqlplus or a similar database client.<br><br>Database Logs: Check the database logs for errors.<br><br>Scenario C: SAP System Down, but OS seems OK<br><br>SAP Application Logs: (High) Check the SAP application logs (transaction SM21 in SAP GUI). Look for errors that might have caused the system to crash.<br><br>Work Process Status: (High) Check the work process overview (transaction SM50). Are all work processes terminated? Are there any dumps? | |

| | | Status: Use transaction SMMS on another central instance to check the message server status on Server 003.<br><br>Impact Assessment: A downed central instance can cause widespread communication issues and prevent application servers from starting.<br><br>General Diagnostic Steps (Applicable to all scenarios):<br><br>OS Level Logs: If you can access the server's OS logs (Windows Event Viewer, Linux system logs), check for errors related to the SAP system or the server itself.<br><br>Network Connectivity: Verify network connectivity between Server 003 and other servers in the SAP landscape. Check firewalls and network devices.<br><br>Resource Usage: If you can get any data, check CPU, memory, and disk space usage on Server 003. Resource exhaustion can cause crashes.<br><br>Phase 3: Recovery (High Priority - Time varies based on diagnosis) | Kernel errors<br><br>Hardware errors<br><br>Security-related events<br><br>Application-specific errors (e.g., database logs)<br><br>Database Investigation (If applicable): (Medium)<br><br>Database Logs: Check the database logs for errors. (e.g., Oracle alert log, SQL Server error log, HANA index server trace).<br><br>Database Connectivity: Verify connectivity to the database from the application server.<br><br>Database Resource Usage: Check database CPU, memory, and I/O utilization.<br><br>SolMan Root Cause Analysis (RCA) Tools: (Medium)<br><br>SolMan Workload Analysis: Use SolMan's workload analysis features to identify performance bottlenecks or resource contention.<br><br>Alert Correlation: SolMan can often correlate alerts from different systems to help pinpoint the root cause.<br><br>Phase 3: Recovery | hostname resolves to the correct IP address.<br><br>Virtualization Layer (if applicable): (Medium)<br><br>VMware/Hyper-V/Cloud Provider: Check the status of the virtual machine in the virtualization platform. Is it running? Are there any errors reported by the hypervisor?<br><br>Resource Allocation: Verify that the VM has sufficient resources (CPU, memory, disk space).<br><br>Phase 3: Recovery (High Priority - as quickly as possible)<br><br>The recovery steps depend entirely on the diagnosis. Here are common scenarios:<br><br>Simple Restart: (High) If the OS is hung or SAP processes are stuck, a simple server restart is often the quickest solution. However, understand the implications of a restart – potential data loss if transactions were in progress.<br><br>SAP System Restart: (High) If the OS is up but the SAP system is down, try restarting the SAP system using sapstop and sapstart. | ABAP Dumps: (Medium) Analyze any ABAP dumps (transaction ST22). A critical dump can bring down the system.<br><br>Phase 3: Recovery (High Priority - Time varies greatly)<br><br>The recovery steps depend entirely on the diagnosis.<br><br>If it's a network issue: Work with the network team to resolve it.<br><br>If it's a hardware issue: Work with the hardware vendor or your infrastructure team to replace the faulty component. This might involve a server rebuild.<br><br>If it's an OS issue (e.g., full disk):<br><br>Free up disk space: Delete unnecessary files, archive logs.<br><br>Restart the SAP instance: sapstart<br><br>If it's an SAP instance issue:<br><br>Restart the SAP instance: sapstart (This is often the first step).<br><br>If the instance fails to start: Analyze the dev_* trace files for more detailed errors. You | |
|---|---|---|---|---|---|---|
| | | Scenario A: Application Server:<br><br>Restart the Server: The simplest solution. If the outage was due to a temporary issue, a restart might resolve it.<br><br>Restart SAP Instance: If a restart doesn't work, try restarting the SAP instance on Server 003.<br><br>Workaround (if possible): If a specific program is causing the issue, temporarily disable it or redistribute the workload to other application servers.<br><br>Scenario B: Database Server:<br><br>Database Restart: This is a last resort. Restarting the database can cause data loss or corruption if not done properly. Follow your database vendor's recommended procedures. Involve the Database Team!<br><br>Database Recovery: If the database is corrupted, you may need to perform a database recovery from backup. This is a complex process and requires expertise.<br><br>Scenario C: Central Services Instance: | (High Priority - Time varies greatly)<br><br>This phase depends heavily on the root cause. Here are common scenarios and recovery steps:<br><br>Simple Restart: (High - if appropriate)<br><br>If the cause is unknown or appears to be a temporary glitch, attempt a controlled restart of the SAP instance(s) on server 003. Use sapstop and sapstart.<br><br>Monitor closely after the restart to ensure the system comes up cleanly and the issue doesn't recur.<br><br>Database Recovery: (High - if database is down)<br><br>Follow your database recovery procedures. This may involve restoring from backup, applying transaction logs, or performing other database-specific recovery steps. This is usually handled by the DBA team.<br><br>Hardware Failure: (High - if hardware is faulty)<br><br>If a hardware component has failed (e.g., disk, memory, CPU), replace the faulty | Database Recovery: (High) If the database is down, you'll need to involve the DBA to perform database recovery. This can be a lengthy process.<br><br>Rollback Changes: (Medium) If the outage was caused by a recent change, roll back the change to restore the system to a working state.<br><br>Restore from Backup: (High - Last Resort) If all else fails, restore the server from a recent backup. This is the most time-consuming option and should only be used as a last resort.<br><br>Important During Recovery:<br><br>Document Everything: Record every step you take during the recovery process.<br><br>Monitor Closely: Monitor the server and SAP system closely after recovery to ensure that the problem does not recur.<br><br>Phase 4: Post-Incident Analysis & Preventative Measures (Medium to Low Priority - within 24-48 hours)<br><br>Root Cause Analysis | might need to rebuild the instance.<br><br>If it's a database issue: Work with the DBA team to resolve the database problem.<br><br>If it's an ABAP dump: Correct the code that caused the dump and restart the SAP instance.<br><br>Important Recovery Considerations:<br><br>Follow documented procedures: Don't deviate from your organization's standard operating procedures.<br><br>Take backups: Before making any significant changes, take backups of critical files and databases.<br><br>Monitor closely: After restarting the SAP instance, monitor it closely for any further errors.<br><br>Phase 4: Post-Incident Activities (Medium Priority - within 24-48 hours)<br><br>Root Cause Analysis (RCA): (High) Don't skip this! Determine the underlying cause of the outage. Was it a hardware failure, a software bug, a configuration error, or a | |

| | | | | | |
|---|---|---|---|---|---|
| | | Restart the Central Instance: Restart the central instance on Server 003.<br><br>Check ICM/Message Server Configuration: Verify the ICM and message server configuration files for errors.<br><br>Important Considerations During Recovery:<br><br>Follow documented procedures: Don't deviate from established recovery procedures.<br><br>Monitor closely: Monitor the server and SAP system closely after recovery to ensure stability.<br><br>Document everything: Record all steps taken during the recovery process.<br><br>Phase 4: Post-Incident Analysis & Preventative Measures (Medium/Low Priority - within 24-72 hours)<br><br>Root Cause Analysis (RCA): Determine the underlying cause of the outage. Was it a hardware failure, software bug, configuration error, or something else?<br><br>Documentation Update: | component. This is usually handled by the infrastructure team.<br><br>Restore the SAP system from backup after the hardware is replaced.<br><br>Software/Configuration Issue: (Medium to High)<br><br>If the issue is caused by a software bug or configuration error, apply the necessary patch or correction.<br><br>Test the fix thoroughly in a non-production environment before applying it to production.<br><br>Rollback (if recent changes): (Medium)<br><br>If the outage occurred immediately after a change (e.g., kernel upgrade, patch application), consider rolling back the change.<br><br>Important Recovery Considerations:<br><br>Follow documented procedures: Always follow your organization's documented recovery procedures.<br><br>Controlled Restart: Restart SAP instances in the correct order.<br><br>Monitoring: Monitor the system closely during | (RCA): (Medium) Conduct a thorough RCA to determine the underlying cause of the outage. Don't just fix the symptom; address the root cause. Use the 5 Whys technique.<br><br>Documentation Update: (Medium) Update documentation with the findings of the RCA and any changes made to prevent future outages.<br><br>Monitoring Improvements: (Medium) Review your monitoring configuration. Could the outage have been detected earlier? Are there any gaps in your monitoring coverage? Add or adjust alerts as needed.<br><br>Automation: (Low) Identify opportunities to automate recovery tasks. For example, you could automate server restarts or database failovers.<br><br>Change Management Review: (Low) If the outage was caused by a change, review your change management process to ensure that changes are properly planned, tested, and documented.<br><br>Tools You'll Likely Use:<br><br>SAP Solution Manager | human mistake?<br><br>Documentation: (High) Update the incident record with the root cause, recovery steps, and any lessons learned.<br><br>Preventative Measures: (Medium) Identify steps to prevent the outage from happening again. This might involve:<br><br>Hardware upgrades<br><br>Software patches<br><br>Configuration changes<br><br>Improved monitoring<br><br>Training for staff<br><br>Communication: (Medium) Communicate the RCA and preventative measures to stakeholders.<br><br>Tools You'll Likely Use:<br><br>SAP Solution Manager (SolMan): Centralized monitoring, alerting, and root cause analysis.<br><br>Operating System Tools: ping, ssh, df, top, htop, vmstat, OS logs.<br><br>Database Tools: sqlplus, database logs.<br><br>SAP GUI: Transactions like SM21, SM50, ST22. | |

| | | | | |
|---|---|---|---|---|
| | | Update your documentation with the findings of the RCA and any changes made to the recovery procedures.<br><br>Preventative Measures:<br><br>Hardware Redundancy: Ensure you have redundant hardware components (e.g., power supplies, network cards, disks).<br><br>Software Updates: Keep your SAP system and operating system up to date with the latest patches and support packages.<br><br>Monitoring Enhancements: Improve your monitoring tools to detect potential issues before they cause an outage. Add specific alerts for Server 003.<br><br>Capacity Planning: Ensure your servers have sufficient resources (CPU, memory, disk space) to handle the workload.<br><br>Regular Backups: Verify that your backups are working correctly and that you can restore them quickly.<br><br>Disaster Recovery Plan: Review and test your disaster recovery plan. | and after the recovery process.<br><br>Communication: Keep stakeholders informed of the recovery progress.<br><br>Phase 4: Post-Incident Activities (Medium Priority - within 24-48 hours)<br><br>Root Cause Analysis (RCA) Documentation: (High)<br><br>Document the root cause of the outage in detail. Include:<br><br>Timeline of events<br><br>Symptoms<br><br>Diagnostic steps taken<br><br>Root cause<br><br>Resolution<br><br>This documentation is crucial for preventing future outages.<br><br>Preventive Measures: (Medium)<br><br>Identify and implement preventive measures to address the root cause. This may involve:<br><br>Applying patches<br><br>Improving monitoring<br><br>Adjusting system configuration | (SolMan): For end-to-end monitoring, alerting, and root cause analysis.<br><br>SAP Cloud Monitoring: SAP's cloud-based monitoring solution.<br><br>Operating System Tools: Windows Event Viewer, Linux syslog, top, vmstat, iostat, df.<br><br>Database Tools: DB02, DB13, database-specific monitoring tools.<br><br>Network Tools: ping, traceroute, pathping, network monitoring tools.<br><br>Virtualization Platform Tools: VMware vSphere Client, Microsoft Hyper-V Manager, cloud provider consoles.<br><br>ITSM System: ServiceNow, Remedy, etc.<br><br>To help me tailor the advice further, could you tell me:<br><br>What applications run on server 003? (e.g., ECC, BW, S/4HANA, Solution Manager)<br><br>What operating system is server 003 running? (e.g., Windows Server, Linux)<br><br>What monitoring tools | Incident Management System: ServiceNow, Jira, etc.<br><br>To help me tailor this advice to your situation, could you tell me:<br><br>What application(s) run on server 003? (e.g., ECC, BW, S/4HANA)<br><br>What operating system is server 003 running? (e.g., Linux, Windows)<br><br>Do you have SAP Solution Manager (SolMan) or a similar monitoring tool?<br><br>What is your role in the SAP landscape? (e.g., Basis administrator, Application administrator, Monitoring specialist)<br><br>This is a lot of information, but it's designed to be a comprehensive guide. Good luck! Let me know if you have any specific questions as you work through the process. | |

| | | Tools You'll Likely Use: | Improving documentation | do you have available? | | |
|---|---|---|---|---|---|---|
| | | SAP Solution Manager (SolMan): For end-to-end monitoring and alerting. | Automating tasks | What is your role in the incident response process? (e.g., first responder, escalation point, subject matter expert) | | |
| | | SAP Cloud Monitoring: SAP's cloud-based monitoring solution. | Incident Review: (Medium) | This is a detailed guide. Remember to prioritize based on the severity of the outage and your company's procedures. Good luck! Let me know if you have any specific questions as you work through this. | | |
| | | Operating System Monitoring Tools: Windows Performance Monitor, Linux top, vmstat, etc. | Conduct a post-incident review with the relevant stakeholders. Discuss: | | | |
| | | Database Monitoring Tools: SAP HANA Studio, Oracle Enterprise Manager, SQL Server Management Studio. | What went well during the incident response | | | |
| | | SAP Transactions: SM50, SM21, SMICM, SMMS, ST03N (Workload Analysis). | What could have been done better | | | |
| | | To help me tailor the advice further, please tell me: | Lessons learned | | | |
| | | What is the role of Server 003 in your SAP landscape? (Application Server, Database Server, Central Services, etc.) | Action items | | | |
| | | What monitoring tools do you have available? | Knowledge Base Update: (Low) | | | |
| | | What is the SAP version and database type? (e.g., S/4HANA on HANA, ECC 6.0 on Oracle) | Update your knowledge base with the information learned from the incident. This will help other teams resolve similar issues in the future. | | | |
| | | | Tools to have readily available: | | | |
| | | | SAP Solution Manager (SolMan) | | | |
| | | | Operating System Monitoring Tools (e.g., top, vmstat, Task Manager) | | | |
| | | | Database Monitoring Tools | | | |

| | | | | | |
|---|---|---|---|---|---|
| | | Do you have a documented disaster recovery plan?<br><br>This is a detailed guide. Remember to prioritize based on the severity of the outage and the impact on your business. Good luck! Let me know if you have any specific questions. | ITSM System (e.g., ServiceNow, Remedy)<br><br>Communication Channels (e.g., email, chat, phone)<br><br>SAP Notes Search (for known issues and solutions)<br><br>To help me tailor this advice to your specific situation, could you tell me:<br><br>What applications are running on server 003?<br><br>What operating system is server 003 running?<br><br>What database is being used (if any)?<br><br>Do you have SAP Solution Manager (SolMan) or a similar monitoring tool?<br><br>What is your role in the SAP landscape? (e.g., Basis Administrator, Application Support, etc.)<br><br>This is a lot of information, but a systematic approach is key to resolving SAP outages quickly and effectively. Good luck! Let me know if you have any questions. | | |

| | Test 3 | | | | | |
|---|---|---|---|---|---|---|
| Act as a SAP Monitoring Expert. What do I do if Server 003 is down? SAP runs Logistics Execution - Delivery and Returns on the server. | | Okay, you've got a downed server (003) running Logistics Execution (LE-DEL/LE-RET). This is serious, as it directly impacts shipping and receiving. Here's a breakdown of how to approach this, acting as your SAP Monitoring Expert. I'll cover immediate actions, investigation, communication, and potential recovery steps. I'll categorize it into phases: Immediate Response, Investigation & Diagnosis, Recovery, and Preventative Actions.

Important Disclaimer: This is a general guide. Specific steps will always depend on your company's documented procedures, your system landscape (HA/DR setup), and your authorization levels. Always follow your company's established protocols.

Phase 1: Immediate Response (First 5-15 minutes)

This is about confirming the outage and minimizing immediate impact.

Confirm the Outage:

Don't rely on a single | Okay, you've got a downed server (003) running Logistics Execution (LE-DEL/LE-RET). This is serious, as it directly impacts shipping and receiving, potentially halting critical business processes. Here's a breakdown of how to approach this, acting as your SAP Monitoring Expert. I'll categorize this into phases: Immediate Response, Investigation, Resolution, and Prevention. I'll also include severity levels and estimated timelines.

Important Disclaimer: This is a general guide. Your specific procedures must align with your company's documented SAP incident management process. Always prioritize safety and follow established protocols.

I. IMMEDIATE RESPONSE (Severity: Critical - P1) - Timeline: 0-15 minutes

Confirm the Outage: Don't rely on a single alert.

SAP Solution Manager (SolMan): Check SolMan for the server status. Look for alerts, system health indicators, and recent logs. This is | Okay, you've got a downed server (003) running Logistics Execution (LE-DEL/LE-RET). This is serious, as it directly impacts shipping and receiving. Here's a breakdown of how to approach this, acting as your SAP Monitoring Expert. I'll cover immediate actions, investigation, communication, and potential recovery steps. I'll categorize it into phases: Immediate Response, Investigation & Diagnosis, Recovery, and Preventative Actions.

Important Disclaimer: This is a general guide. Specific steps will always depend on your company's documented procedures, your system landscape (HA/DR setup), and your authorization levels. Always follow your company's established protocols first.

Phase 1: Immediate Response (First 5-15 Minutes)

This is about confirming the outage and minimizing immediate impact.

Confirm the Outage:

Don't rely on a single | Alright, let's get this sorted out! Here's my step-by-step approach as your SAP Monitoring Expert when Server 003, hosting your Logistics Execution (LE-DLR) system, goes down:

**1. Initial Assessment:**

* **Confirmation:** Double-check that Server 003 is truly down. Check its physical status (lights, fans), network connectivity, and if it's responding to pings.
* **Log Monitoring:** Immediately dive into the server logs on Server 003 and any related SAP application servers for error messages, system dumps, or anything unusual preceding the downtime.
* **Alerting System:** Were there any alerts triggered in your monitoring system (e.g., SAP Solution Manager, third-party tools) before the server went down? Analyze these alerts for clues.

**2. Impact Analysis:**

* **System Scope:** Determine the full scope of the outage. Is it just LE-DLR affected, or are other SAP modules or non-SAP applications running on Server 003 also impacted? | Alright, let's get this sorted out! Server 003 going down with Logistics Execution - Delivery and Returns (LE-DTR) running on it is a serious issue. Here's my step-by-step approach as your SAP Monitoring Expert:

**1. Immediate Actions:**

* **Check System Status:** Log into your central SAP system monitoring tools (e.g., RZMON, SMICM) to confirm the status of Server 003. Are there any error messages or alerts? Look for specific errors related to LE-DTR instances.
* **Contact Infrastructure Team:** Immediately notify your IT infrastructure team about the server outage. They'll likely have more detailed insights into the cause and can start troubleshooting hardware/network issues.

**2. Gather Information:**

* **Recent Changes:** Were there any recent system updates, hardware installations, or configuration changes on Server 003 or related systems before the outage? This information |

| | | | | |
|---|---|---|---|---|
| | source. Check multiple monitoring tools:<br><br>SAP Solution Manager (SolMan): This is your primary source. Look for alerts, system status (red light!), and detailed error messages. Check the System Landscape Directory (SLD) for server status.<br><br>Operating System Monitoring (e.g., CCMS, Nagios, Zabbix): Verify the server is unreachable at the OS level (ping, SSH).<br><br>Database Monitoring: Check if the database instance associated with server 003 is also down. (This is critical - a database down means a much bigger problem).<br><br>Check for Planned Downtime: Double-check the change management calendar. Is this a scheduled outage you forgot about? (It happens!)<br><br>Initial Impact Assessment:<br><br>LE-DEL/LE-RET Criticality: You already know this is important, but quantify it. How many deliveries are blocked? How many returns are pending? What's the financial | your primary source of truth.<br><br>Operating System Level: If possible (e.g., via a remote console if available), attempt to ping the server. If ping fails, it's likely a network or OS issue. Try to SSH/RDP if configured.<br><br>SAP GUI: Attempt to connect to the SAP system via SAP GUI. You'll likely get a connection refused error.<br><br>Declare an Incident: Immediately log an incident in your ITSM system (ServiceNow, Remedy, etc.). Include:<br><br>Severity: P1 (Critical - Business Impacting)<br><br>Component: SAP Basis / Infrastructure<br><br>Affected System: The SAP system instance running on server 003.<br><br>Affected Functionality: Logistics Execution (Delivery and Returns)<br><br>Brief Description: "Server 003 is down. Impacting Delivery and Returns processing."<br><br>Initial Communication: Notify key stakeholders:<br><br>SAP Basis Team: The | source. Check multiple monitoring tools:<br><br>SAP Solution Manager (SolMan): This is your primary source. Look for alerts, system status (red light!), and detailed error messages. Check the System Landscape Directory (SLD) for server status.<br><br>Operating System Monitoring (e.g., CCMS, Nagios, Zabbix): Verify the server is unreachable at the OS level (ping, SSH).<br><br>Database Monitoring: Check if the database instance associated with server 003 is reachable. (This is crucial, as a DB down can look like an app server down).<br><br>Check SAP Logon: Attempt to connect to the system via SAP GUI. You'll likely get a connection refused error.<br><br>Initial Impact Assessment:<br><br>LE-DEL/LE-RET Criticality: You already know this is important, but quantify it. How many deliveries are blocked? How many returns are pending? What's the financial impact per hour of downtime? | * **Business Impact:** Assess the immediate consequences of the downtime for your delivery and returns processes. Are customers unable to place orders, track shipments, or process returns?<br><br>**3. Immediate Actions:**<br><br>* **Contact Support:** If you have an SAP support contract, open a ticket immediately. Provide as much detail as possible about the situation.<br>* **Fallback Mechanisms:** Do you have any disaster recovery or failover mechanisms in place for LE-DLR? If so, attempt to activate them according to your plan.<br><br>**4. Troubleshooting & Recovery:**<br><br>* **Isolate the Problem:** Narrow down the cause of the server failure. Is it hardware (disk, memory, CPU), software (operating system, SAP kernel), or a network issue?<br>* **Diagnostic Tools:** Use SAP's built-in diagnostic tools (e.g., transaction code SM51) and operating system utilities to gather more information about the | could provide clues about the root cause.<br>* **Error Logs:** Examine the error logs on Server 003 and any associated databases. Look for specific error messages or patterns that might point to the problem. SAP logs are your best friends in these situations!<br><br>**3. Impact Assessment:**<br><br>* **Business Processes:** Identify which business processes rely on LE-DTR running on Server 003. This will help you understand the scope of the outage and prioritize recovery efforts.<br>* **Users Affected:** Determine which users are impacted by the server downtime. Communicate the situation to them promptly and provide updates as they become available.<br><br>**4. Recovery Options:**<br><br>* **Failover Mechanisms:** Does your system have a configured failover mechanism for Server 003? If so, initiate the failover process if possible. This should automatically switch workloads to a backup |
| | | impact per hour of downtime?<br><br>Affected Users: Which teams/departments are directly impacted (Warehouse, Shipping, Customer Service)?<br><br>Workarounds (Temporary): Are there any manual workarounds possible? (e.g., temporarily diverting shipments to another location, manually processing returns). Document these immediately. These are not long-term solutions.<br><br>Initial Communication:<br><br>Incident Creation: Create a high-priority incident in your ITSM system (ServiceNow, Remedy, etc.). Include:<br><br>Server Name: 003<br><br>Affected Component: Logistics Execution (LE-DEL/LE-RET)<br><br>Severity: High/Critical (based on impact)<br><br>Initial Description: "Server 003 is down. Impacting Delivery and Returns processing."<br><br>Notify Key Stakeholders: Immediately inform:<br><br>SAP Basis Team | primary team responsible for server recovery.<br><br>LE-DEL/LE-RET Functional Team: They need to understand the impact and prepare for potential workarounds.<br><br>Business Owners/Management: Inform them of the outage and estimated timelines (even if preliminary). Transparency is crucial.<br><br>Check for Automated Restart: Many systems have automated restart mechanisms. See if the server is attempting to come back up on its own. Do not interfere with this process unless it's clearly failing.<br><br>II. INVESTIGATION (Severity: High - P2) - Timeline: 15-60 minutes<br><br>This phase is about understanding why the server went down. The Basis team leads this.<br><br>Operating System Logs: This is the first place to look. Access the server console (if possible) or use remote access tools. Examine:<br><br>System Logs (e.g., /var/log/messages on Linux, Event Viewer on Windows): Look for | Business Process Impact: Specifically, what transactions are affected? (e.g., VL01N, VL02N, LT03, LT12, etc.). This helps with communication.<br><br>Check for HA/DR: Crucially, is this server part of a High Availability (HA) or Disaster Recovery (DR) setup? If so, failover procedures should be initiated immediately (see Recovery Phase).<br><br>Initial Communication:<br><br>Alert the Support Team: Immediately notify the relevant SAP Basis/System Administration team. Use your company's escalation process (e.g., PagerDuty, email distribution list, phone call).<br><br>Brief Stakeholders: Inform key Logistics/Supply Chain stakeholders (e.g., Logistics Manager, Warehouse Manager) that there's a system outage affecting deliveries and returns. Keep it concise: "We're aware of an issue with the SAP system impacting delivery processing. The team is investigating." Avoid speculation about the | problem.<br><br>**5. Communication & Documentation:**<br><br>* **Stakeholders:** Keep relevant stakeholders (management, IT, business users) informed about the situation, estimated downtime, and any recovery progress.<br>* **Incident Report:** Document the entire incident thoroughly, including the timeline, symptoms, troubleshooting steps taken, and final resolution. This will be valuable for future reference and potential preventive measures.<br><br>**Important Notes:**<br><br>* **Business Continuity:** If the outage persists, prioritize maintaining essential business operations (e.g., customer service, order fulfillment) even if it requires manual workarounds.<br>* **Root Cause Analysis:** Once the server is back online, conduct a thorough root cause analysis to prevent similar incidents in the future.<br><br>Remember, stay calm and methodical. With a clear plan and | server.<br>* **Manual Restart:** If failover isn't an option or is unsuccessful, try restarting the SAP instances on Server 003 manually. Ensure you follow proper shutdown procedures before attempting a restart.<br><br>**5. Root Cause Analysis (RCA):**<br><br>Once Server 003 is back online, conduct a thorough RCA to identify the root cause of the outage. This will help prevent similar incidents in the future. Document your findings and implement corrective actions accordingly.<br><br>**Important Considerations:**<br><br>* **Communication:** Maintain clear and concise communication with all stakeholders throughout the process. Keep users informed about the status and any potential impact on their work.<br>* **Documentation:** Thoroughly document all steps taken, including troubleshooting efforts, solutions implemented, and lessons learned. This will be invaluable for future reference.<br><br>Let me know if you have |

| | | | | | |
|---|---|---|---|---|---|
| | SAP Application Team (LE-DEL/LE-RET experts)<br><br>IT Operations/Infrastructure Team<br><br>Business Owners/Managers of affected departments (Warehouse Manager, Logistics Manager) – a brief, factual update. Avoid speculation.<br><br>Phase 2: Investigation & Diagnosis (15 minutes - 1 hour+)<br><br>Now you need to find why the server is down.<br><br>OS Level Checks (Basis/Infrastructure Team Lead):<br><br>Server Console: If possible, access the server console (iDRAC, iLO, etc.) to see boot messages or error screens.<br><br>System Logs: Examine the OS logs (e.g., /var/log/messages on Linux, Event Viewer on Windows) for clues. Look for errors related to hardware, network, or SAP processes.<br><br>Resource Usage: Check CPU, memory, disk space, and network utilization before the | errors, crashes, or hardware failures.<br><br>SAP Instance Logs: Check the SAP instance logs (usually in /usr/sap/<SID>/<instance_number>/log) for any errors leading up to the outage. Focus on dev_w3c, icm_icm01, and disp+work.<br><br>SAP System Logs: Use SAP transaction SM21 (System Log) to review logs for errors and warnings around the time of the outage. Filter by time and relevant components.<br><br>Database Logs: Check the database logs (Oracle, SQL Server, HANA) for errors. A database issue can easily bring down an SAP system.<br><br>Network Connectivity: Verify network connectivity to the server. Is the server reachable from other servers in the landscape? Check network devices (switches, routers, firewalls).<br><br>Hardware Health: If possible, check the server's hardware health (CPU, memory, disk I/O). Use server management tools (e.g., iLO, iDRAC). | cause at this stage.<br><br>Phase 2: Investigation & Diagnosis (15 Minutes - 1 Hour+)<br><br>Now, we need to find why the server is down.<br><br>Check SAP System Logs: (Use transaction SM21)<br><br>System Log: Look for recent errors, warnings, and short dumps. Filter by time around the outage. Pay attention to messages related to:<br><br>Database connection issues<br><br>Memory errors<br><br>CPU utilization<br><br>Disk space<br><br>SAP instance start/stop failures<br><br>Work Process Log: (SM50) Check for hung work processes or abnormal terminations.<br><br>Database Logs: Access the database logs (e.g., Oracle alert log, SQL Server error log) to see if the database itself is experiencing problems.<br><br>Operating System Level Checks: (Requires OS access) | systematic approach, you'll get Server 003 up and running again as quickly as possible. Let me know if you need any more specific advice during your troubleshooting process! | any more specific details about your system setup or the error messages you're seeing. I can provide more tailored guidance based on that information. |
| | | outage (if possible – historical data is key). Was the server overloaded?<br><br>Network Connectivity: Verify network connectivity to the server. Is it reachable from other servers? Are there firewall issues?<br><br>SAP Level Checks (SAP Basis/Application Team):<br><br>SAP System Log (SM21): Check for errors and warnings in the SAP system log leading up to the outage. Look for database errors, short dumps, or resource-related issues.<br><br>Transaction ST22 (ABAP Dump Analysis): Check for recent ABAP dumps that might have caused the system to crash.<br><br>Transaction SM50/SM66 (Work Process Overview): If the server is partially up, check the work process overview for hung or blocked processes.<br><br>Database Connection: Verify the database connection from the SAP application server. Can the SAP system connect to the database?<br><br>Transaction DB02 | Recent Changes: Were there any recent changes to the server (OS patches, SAP upgrades, configuration changes)? This is a common cause of outages. Check change management records.<br><br>Resource Monitoring: Check historical resource usage (CPU, memory, disk) to see if the server was under heavy load before the outage. SolMan or other monitoring tools are helpful here.<br><br>III. RESOLUTION (Severity: Variable - P1/P2) - Timeline: 60 minutes - Several Hours (depending on cause)<br><br>This is where you fix the problem. The specific steps depend entirely on the root cause. Here are some common scenarios and resolutions:<br><br>OS Crash/Hardware Failure:<br><br>Restart the Server: If a simple restart resolves the issue, do so. However, investigate the root cause to prevent recurrence.<br><br>Hardware Replacement: If a hardware component failed, | CPU Usage: Was the server maxed out on CPU?<br><br>Memory Usage: Was the server running out of memory?<br><br>Disk Space: Is the disk where the SAP installation resides full? (Especially /usr/sap/<SID>/DVEBMGS00 for typical installations)<br><br>Network Connectivity: Can you ping the server from other servers in the landscape? Check firewall rules.<br><br>OS Logs: Examine the OS system logs (e.g., /var/log/messages on Linux, Event Viewer on Windows) for hardware errors, OS crashes, or other relevant events.<br><br>SAP Transaction Analysis (If possible, even limited access):<br><br>ST22 (ABAP Dump Analysis): If you can get any access, check for recent ABAP dumps that might have contributed to the instability.<br><br>SM50/SM66 (Work Process Overview/Analysis): Look for patterns of work process failures.<br><br>Recent Changes: | | |

| | | | | |
|---|---|---|---|---|
| | (Database Performance): Check database performance metrics. Was the database experiencing high load or contention?<br><br>Check SAP Router (if applicable): If using a SAP Router, verify its status and connectivity.<br><br>Correlation: Look for correlations between OS-level events and SAP-level errors. For example, a disk space issue on the OS might cause database errors in SAP.<br><br>Phase 3: Recovery (Time varies greatly - 30 minutes to several hours)<br><br>This is where you get the server back online. The specific steps depend entirely on the cause of the outage.<br><br>Simple Restart (First Attempt - if appropriate): If the cause is unknown and there's no indication of hardware failure, a controlled server restart is often the first step. Coordinate this with the Basis team.<br><br>Database Recovery: If the database is down, database administrators will need to perform database recovery procedures (restore from | replace it. This may require a service window.<br><br>SAP Instance Failure:<br><br>Restart the SAP Instance: Use sapstart or sapcontrol to restart the SAP instance.<br><br>Check SAP Profile Parameters: Incorrect profile parameters can cause instance failures.<br><br>Database Issue:<br><br>Database Recovery: Work with the DBA team to recover the database. This may involve restoring from backup.<br><br>Network Issue:<br><br>Resolve Network Connectivity: Fix the network issue (e.g., firewall rule, cable problem).<br><br>Resource Exhaustion:<br><br>Increase Resources: Increase CPU, memory, or disk space.<br><br>Optimize SAP Configuration: Tune SAP parameters to reduce resource consumption.<br><br>Software Bug:<br><br>Apply SAP Note: Search SAP Support Portal for relevant SAP Notes and | Transport Logs: Were any transports deployed to this server recently? A faulty transport could be the cause.<br><br>System Configuration Changes: Were any system parameters changed recently?<br><br>Kernel Updates: Was a kernel update applied recently?<br><br>Phase 3: Recovery (Time Varies - Dependent on Cause)<br><br>This is where you get the system back up. The steps depend entirely on the diagnosis.<br><br>HA/DR Failover (If Applicable): This is the fastest recovery. If the server is part of an HA/DR setup, initiate the failover procedure immediately. Follow your documented failover runbook.<br><br>Restart the SAP Instance: (If not HA/DR, and the cause is unclear)<br><br>Controlled Restart: Use transaction SM51 to stop and start the SAP instance on server 003. Monitor the logs closely during the restart.<br><br>OS Level Restart: If | | |
| | | backup, apply logs, etc.). This is the most time-consuming recovery scenario.<br><br>Hardware Failure: If a hardware component has failed (disk, memory, CPU), it will need to be replaced. This may involve restoring from backup or using a hot standby server.<br><br>Software Issue: If a software issue (e.g., corrupted SAP files) is identified, it may require restoring from backup, applying patches, or reinstalling components.<br><br>Failover (If HA is configured): If you have a High Availability (HA) setup (e.g., using clustering software), initiate a failover to the standby server. This is the fastest recovery option.<br><br>Post-Recovery Checks:<br><br>SAP System Status: Verify the SAP system is running and all services are online.<br><br>LE-DEL/LE-RET Functionality: Test key LE-DEL/LE-RET transactions to ensure they are working correctly (e.g., create a delivery, process a return). | apply them.<br><br>During Resolution:<br><br>Regular Communication: Keep stakeholders updated on progress. Provide realistic ETAs.<br><br>Testing: After the server is back up, thoroughly test Logistics Execution functionality to ensure it's working correctly. Test deliveries, returns, and related processes.<br><br>IV. PREVENTION (Severity: Low - P3) - Timeline: Ongoing<br><br>This is about preventing future outages.<br><br>Root Cause Analysis (RCA): Conduct a thorough RCA to identify the underlying cause of the outage. Document the findings and corrective actions.<br><br>Monitoring Enhancements:<br><br>Proactive Monitoring: Implement proactive monitoring to detect potential issues before they cause an outage. Monitor CPU, memory, disk space, database performance, and SAP instance health.<br><br>Alerting: Configure alerts to notify the Basis team of critical events. | SM51 fails, try restarting the SAP services at the OS level.<br><br>Database Restart: (If the database is down)<br><br>Coordinate with DBA: Work with your Database Administrator to restart the database instance. This is a critical step and should be done carefully.<br><br>Rollback Recent Changes:<br><br>Transport Rollback: If a recent transport is suspected, roll it back.<br><br>Configuration Reversion: If a recent configuration change is suspected, revert it.<br><br>Hardware Issues:<br><br>Hardware Replacement: If a hardware failure is identified (e.g., disk failure, memory error), coordinate with your hardware vendor for replacement.<br><br>Post-Recovery Checks:<br><br>System Status: Verify that the system is back to a green status in SolMan.<br><br>Transaction Testing: Test critical Logistics Execution transactions (VL01N, VL02N, LT03, | | |

| | | | | |
|---|---|---|---|---|
| | System Log: Check the SAP system log (SM21) for any new errors or warnings.<br><br>Performance Monitoring: Monitor system performance closely after recovery to ensure it is stable.<br><br>Phase 4: Preventative Actions (Following Recovery - Days/Weeks)<br><br>This is about preventing the outage from happening again.<br><br>Root Cause Analysis (RCA): Conduct a thorough RCA to determine the underlying cause of the outage. Don't just fix the symptom; address the root problem.<br><br>Monitoring Improvements:<br><br>Add/Adjust Alerts: Create or adjust monitoring alerts to detect similar issues in the future. Focus on leading indicators (e.g., high CPU usage, low disk space) rather than just the outage itself.<br><br>Automated Checks: Implement automated checks to proactively identify potential problems.<br><br>Capacity Planning: | Capacity Planning: Regularly review capacity requirements and ensure the server has sufficient resources.<br><br>Patch Management: Keep the operating system and SAP software up to date with the latest patches.<br><br>Change Management: Strictly enforce change management procedures to minimize the risk of introducing errors.<br><br>Disaster Recovery (DR): Ensure you have a robust DR plan in place to recover from outages quickly. Regularly test the DR plan.<br><br>High Availability (HA): Consider implementing HA solutions (e.g., clustering) to minimize downtime.<br><br>Specific Considerations for LE-DEL/LE-RET:<br><br>Interfaces: LE-DEL/LE-RET often integrates with other systems (e.g., Warehouse Management, Transportation Management). Ensure these interfaces are also functioning correctly after the server is restored. | LT12) to ensure they are functioning correctly.<br><br>Monitor Performance: Monitor CPU, memory, and disk usage to ensure the system is stable.<br><br>Phase 4: Preventative Actions (Post-Incident)<br><br>This is about preventing the issue from happening again.<br><br>Root Cause Analysis (RCA): Conduct a thorough RCA to determine the underlying cause of the outage. Don't just fix the symptom; fix the problem.<br><br>Monitoring Improvements:<br><br>Add Alerts: Create alerts in SolMan and OS monitoring tools to proactively detect potential issues (e.g., high CPU usage, low disk space).<br><br>Threshold Adjustments: Adjust monitoring thresholds to be more sensitive.<br><br>Capacity Planning: Review system capacity to ensure it can handle the current workload.<br><br>Documentation Updates: Update your incident | | |

| | | | | |
|---|---|---|---|---|
| | Review system capacity and plan for future growth. Ensure the server has sufficient resources to handle peak loads.<br><br>Patch Management: Keep the operating system and SAP software up to date with the latest patches and security fixes.<br><br>Disaster Recovery (DR) Testing: Regularly test your DR plan to ensure it is effective.<br><br>Documentation: Update documentation with lessons learned from the outage.<br><br>Tools You'll Likely Use:<br><br>SAP Solution Manager (SolMan)<br><br>SAP GUI (for transactions like SM21, ST22, SM50, DB02)<br><br>Operating System Monitoring Tools (CCMS, Nagios, Zabbix)<br><br>Database Monitoring Tools<br><br>ITSM System (ServiceNow, Remedy)<br><br>Server Console Access (iDRAC, iLO)<br><br>To help me tailor this advice further, could you | Background Jobs: Check for any failed background jobs related to Delivery and Returns processing.<br><br>Lock Entries: Check for any lock entries that may be preventing users from processing deliveries or returns. (Transaction SM12)<br><br>Tools You'll Use:<br><br>SAP Solution Manager (SolMan)<br><br>SAP GUI (SM21, SM12, ST22)<br><br>Operating System Tools (ping, SSH, Event Viewer, system logs)<br><br>Database Management Tools<br><br>ITSM System (ServiceNow, Remedy, etc.)<br><br>Server Management Tools (iLO, iDRAC)<br><br>To help me tailor this advice further, could you tell me:<br><br>What is your operating system? (Linux, Windows, etc.)<br><br>What database are you using? (Oracle, SQL Server, HANA) | response runbooks and disaster recovery plans based on the lessons learned.<br><br>Change Management Review: Review your change management process to ensure that changes are properly tested and documented.<br><br>To help me give you more tailored advice, please tell me:<br><br>Do you have HA/DR configured for this server?<br><br>What is the operating system of server 003? (Linux, Windows, etc.)<br><br>What database are you using? (Oracle, SQL Server, HANA, etc.)<br><br>What monitoring tools are you currently using? (SolMan, CCMS, Nagios, Zabbix, etc.)<br><br>What were the last changes made to the system before the outage? (Transports, configuration changes, kernel updates)<br><br>This is a comprehensive approach. Remember to prioritize, communicate effectively, and follow your company's procedures. Good luck! Let me know if you have any specific questions | | |

| | | tell me:

Do you have a High Availability (HA) setup for this server? (e.g., clustering)

What operating system is running on server 003? (Linux, Windows)

What database are you using? (SAP HANA, Oracle, SQL Server, etc.)

What monitoring tools are currently in place?

Do you have a documented disaster recovery plan?

Let me know if you'd like me to elaborate on any of these steps. Good luck! Let me know how it goes. | Do you have a Disaster Recovery plan in place?

What monitoring tools are currently in use?

Let me know if you have any other questions. I'm here to help you get through this! | as you work through this. | | |
| --- | --- | --- | --- | --- | --- |

| | | |
|---|---|---|
| Act as a SAP Monitoring Expert. What do I do if Server 003 is down? SAP runs Logistics Execution - Delivery and Returns on the server. That brings down Dispatching Bay 17. | **Test 4** | Okay, you've got a critical situation: Server 003 is down, impacting Logistics Execution (Delivery & Returns) and specifically Dispatching Bay 17. Here's a breakdown of how to approach this as a SAP Monitoring Expert, categorized into phases: Immediate Response, Investigation, Resolution, and Preventative Actions. I'll be detailed, assuming you have standard SAP monitoring tools available.<br><br>Important Assumptions:<br><br>You have access to SAP Solution Manager (SolMan) or a similar monitoring tool (e.g., Dynatrace, New Relic). This is crucial.<br><br>You have basic SAP Basis/System Admin knowledge.<br><br>You have a documented incident management process. Follow it!<br><br>Phase 1: Immediate Response (0-15 minutes - Containment & Initial Assessment)<br><br>Acknowledge the Alert: Confirm the alert in your monitoring system. Don't dismiss it. Mark it as "In Progress" and assign it to yourself (or the |

| As a SAP Monitoring Expert, here's my immediate action plan when Server 003 goes down, impacting Dispatching Bay 17: | Okay, this sounds like a serious situation! Here's what I would do as an SAP Monitoring Expert if Server 003 was down, affecting your Delivery and Returns system and Dispatching Bay 17: | Okay, let's troubleshoot this situation. Here's a step-by-step approach as a SAP Monitoring Expert: | Okay, this is a critical situation! Here's my step-by-step approach as a SAP Monitoring Expert: |
|---|---|---|---|
| **1. Acknowledge & Isolate:**<br><br>* **Acknowledge the Issue:** Immediately inform relevant stakeholders (IT, Operations, Warehouse Managers) about the server downtime and its impact on Delivery & Returns processes in Dispatching Bay 17.<br>* **Isolate the Impact:** Determine the extent of the outage. Can other SAP systems running on different servers continue functioning? Is this a complete system crash or a specific application issue?<br><br>**2. Investigate the Cause:**<br><br>* **System Logs:** Check the logs on Server 003 and any related servers for error messages, warnings, or unusual activity preceding the downtime.<br>* **Hardware Checks:** Verify hardware status (CPU, memory, disk I/O, network connectivity) on Server 003. Are there any physical issues?<br>* **SAP System Logs:** Analyze SAP system | **Immediate Actions:**<br><br>1. **Acknowledge the Issue:** Immediately inform relevant stakeholders (IT team, logistics management, dispatchers) about the server outage and its impact on operations.<br>2. **Verify Server Status:** Check if Server 003 is truly down or experiencing performance issues. Use monitoring tools to assess:<br>* Network connectivity<br>* CPU utilization<br>* Memory usage<br>* Disk space<br>* System logs for error messages<br>3. **Isolate the Problem:** Try to pinpoint the cause of the server outage. Is it a hardware failure, software issue, network problem, or power outage?<br><br>**Troubleshooting Steps:**<br><br>4. **Restart Server (If Possible):** If the issue appears to be temporary, attempt a | **1. Immediate Actions:**<br><br>* **Acknowledge the Incident:** Log the incident in your monitoring system (e.g., Solution Manager). Include details like timestamp, affected server (003), impacted application (Logistics Execution - Delivery and Returns), and specific consequence (Dispatching Bay 17 down).<br>* **Notify Stakeholders:** Inform relevant personnel: IT operations, SAP basis team, warehouse management, and potentially dispatching bay supervisors.<br><br>**2. Investigate Server Downtime:**<br><br>* **Check System Status:** Use your monitoring tools to check the status of server 003. Look for any error messages, system logs, or performance indicators that point to the cause of the downtime.<br>* **Network Connectivity:** Verify network connectivity | **1. Immediate Actions:**<br><br>* **Acknowledge the Issue:** Notify relevant personnel (IT team, warehouse managers, logistics staff) about Server 003 being down and the impact on Dispatching Bay 17 immediately.<br>* **System Health Check:** Log into your SAP monitoring tools (e.g., STMS, RZMON, SM51) to verify:<br>* Server 003's actual status – Is it completely unresponsive or experiencing intermittent issues?<br>* Dependency checks – Are other critical systems relying on Server 003 also down?<br>* **Incident Management:** Open an incident ticket in your system, documenting the issue, its impact (dispatching bay downtime), affected users, and any error messages received.<br><br>**2. Troubleshooting:**<br><br>* **Check Infrastructure Logs:** Review server logs (Windows Event Viewer or Linux syslog) |

| | | | | |
|---|---|---|---|---|
| appropriate on-call team).<br><br>Verify the Outage: Don't rely solely on the monitoring tool.<br><br>Ping the Server: Can you ping Server 003? This confirms basic network connectivity.<br><br>Check OS Level: If possible, attempt to SSH/RDP into Server 003. Is the OS responsive? If not, it's likely a hardware or OS-level issue.<br><br>Impact Confirmation: You already know Dispatching Bay 17 is down, but quantify the impact.<br><br>Check Delivery Queues: How many deliveries are blocked? (Transaction: /SCWM/PRDI - Delivery Monitor, or relevant WM/EWM transaction depending on your setup). This is a key metric.<br><br>Check Return Processing: Is returns blocked? (Relevant EWM/WM transaction).<br><br>Communicate: Briefly inform key stakeholders (Logistics Manager, Warehouse Supervisor, potentially Customer Service) that there's an issue impacting | logs (e.g., ST22, SM58) for errors related to the Delivery & Returns application or any other relevant modules.<br><br>**3. Recovery Options:**<br><br>* **Restart Server:** If it seems like a temporary glitch, attempt restarting the server. Monitor system logs and performance closely after restart.<br>* **Check Backup/Restore:** If a restart doesn't work, investigate recent backups of Server 003. Can you restore the system to a stable state?<br><br>**4. Communicate & Manage Expectations:**<br><br>* **Transparency is Key:** Keep stakeholders informed about the situation, the investigation process, and potential recovery timelines.<br>* **Minimize Downtime:** Prioritize quick resolution. If restoration takes time, implement workarounds (manual processing, limited functionality) to minimize disruption in Dispatching Bay 17.<br><br>**5. Prevent Future Outages:**<br><br>* **Root Cause | server restart. This often resolves minor glitches.<br>5. **Check System Logs:** Examine system logs on Server 003 for error messages that might provide clues about the cause of the outage. Look for patterns or recurring issues.<br>6. **Contact SAP Support:** If you suspect an SAP-related issue, immediately contact SAP support for guidance. Provide them with detailed information about the problem, including:<br>* Error messages<br>* System logs<br>* Relevant configuration details<br>7. **Review Backup and Recovery Plans:** If the server outage persists, initiate your disaster recovery plan. This may involve restoring from backups or switching to a failover system.<br><br>**Mitigation and Communication:**<br><br>8. **Inform Dispatching Bay 17:** Keep dispatchers updated on the situation and advise them of any alternative processes they can use during the outage. Consider manual workarounds for urgent deliveries.<br>9. **Document the Incident:** Record all | between server 003 and other critical systems (e.g., databases, application servers). Ping the server from other machines on the network.<br>* **Hardware Checks:** If possible, remotely check the physical health of server 003:<br>* CPU usage<br>* Memory utilization<br>* Disk space availability<br>* Power supply status<br><br>**3. SAP System Checks:**<br><br>* **Central Instance Check:** If server 003 hosts the central instance (e.g., ABAP system), ensure it's down completely. Check the SAP message server and any error logs related to the system.<br>* **Application Server Status:** If server 003 is an application server specifically for Delivery and Returns, check its status within the SAP landscape.<br><br>**4. Potential Causes & Troubleshooting Steps:**<br><br>* **Hardware Failure:** Replace faulty hardware components (e.g., hard drives, RAM). Contact your hardware vendor if necessary.<br>* **Software Errors:** Analyze system logs for | for clues about the failure. Look for errors related to hardware, operating system, or network connectivity.<br>* **Network Connectivity:** Ensure Server 003 is reachable from other systems. Check network cables, switches, and firewalls.<br>* **Resource Utilization:** Analyze resource consumption on Server 003 (CPU, memory, disk I/O). Is it overloaded or experiencing performance issues?<br>* **SAP Instance Status:** Use transaction SM51 to check the status of all SAP instances running on Server 003. Are they green or showing errors?<br><br>**3. Recovery Options:**<br><br>* **Restart Server:** If resources are heavily utilized, a simple restart might resolve the issue. Coordinate this with IT for minimal disruption.<br>* **Failover Mechanism:** Does your setup have a failover cluster or high availability configuration? Activate it if available.<br>* **Manual Failover:** If no automatic mechanisms exist, you might need to manually switch over SAP |

| | | deliveries/returns. Avoid speculation at this stage. "We are investigating an outage on Server 003 impacting delivery processing. We will provide updates as soon as possible."<br><br>Initial System Status Check (SolMan/Monitoring Tool):<br><br>SAP System Status: Is the SAP system itself (SID) showing as down in SolMan? Or is it just the application server?<br><br>Work Process Overview: If the SAP system is up, but the server is showing issues, check work process usage on Server 003. Are all work processes dead? Are there any hung dialogs? (Transaction: SM50 - Work Process Overview).<br><br>Alert History: Review the alert history in SolMan for Server 003. Were there any preceding alerts (high CPU, memory issues, disk space low) that might indicate the cause?<br><br>Phase 2: Investigation (15-60 minutes - Root Cause Analysis)<br><br>This is where you dig deeper. The approach depends on what you | Analysis:** Once the issue is resolved, conduct a thorough root cause analysis to identify what caused the downtime and prevent recurrence.<br>* **Monitoring & Alerting:** Ensure robust monitoring of SAP systems (e.g., CPU utilization, memory usage, disk space) with proactive alerts for potential issues.<br>* **Disaster Recovery Plan:** Review and update your disaster recovery plan to include specific procedures for Server 003 failures and their impact on Delivery & Returns.<br><br>**Important Notes:**<br><br>* **Security:** Be cautious about accessing or modifying systems remotely. Ensure proper authorization and security protocols are followed.<br>* **Documentation:** Document all steps taken during the incident, including troubleshooting efforts and recovery procedures. This will be valuable for future reference and analysis.<br><br>Let me know if you have any more details about the situation, and I can | steps taken to diagnose and resolve the issue, including timestamps, troubleshooting actions, and communication with stakeholders. This information will be valuable for future reference and potential root cause analysis.<br><br>**Preventing Future Outages:**<br><br>10. **Proactive Monitoring:** Implement comprehensive monitoring tools to proactively identify potential issues before they cause outages.<br>11. **Regular System Maintenance:** Perform routine system maintenance tasks, including software updates, security patches, and performance tuning.<br>12. **Hardware Redundancy:** Consider implementing redundant hardware systems to minimize downtime in case of a server failure.<br>13. **Disaster Recovery Planning:** Develop and test a robust disaster recovery plan that includes procedures for restoring critical systems and data.<br><br>Remember, quick action and clear communication are crucial during an SAP | errors related to operating system, database, or SAP itself. Try restarting relevant services. Apply SAP notes or patches to address known issues.<br>* **Network Problems:** Resolve connectivity issues by checking network configuration, firewall rules, and cabling.<br>* **Resource Exhaustion:** Investigate high CPU/memory usage on server 003. Optimize SAP configurations, adjust system parameters, or consider adding resources (e.g., more RAM).<br><br>**5. Communication & Recovery:**<br><br>* **Update Stakeholders:** Keep everyone informed about your progress and estimated recovery time.<br>* **Workarounds:** Explore temporary workarounds to minimize disruption, such as manual processing in Dispatching Bay 17 if possible.<br>* **Recovery Plan:** Once the server is back online, perform a thorough system check and ensure all data integrity before restarting full operations. | processes to another server (if resources permit).<br><br>**4. Communication & Impact Mitigation:**<br><br>* **Warehouse Staff:** Keep warehouse staff informed about the situation and provide alternative procedures for handling dispatches until Server 003 is back online.<br>* **Customers:** Communicate potential delays to customers affected by the downtime, if possible.<br>* **Documentation:** Thoroughly document the issue, troubleshooting steps taken, recovery procedures used, and any lessons learned for future reference.<br><br>**5. Root Cause Analysis:**<br><br>* Once Server 003 is restored, conduct a root cause analysis to identify what caused the outage. This helps prevent similar incidents in the future.<br><br>Remember: **Time is of the essence!** Act swiftly and communicate transparently throughout the process. |

| | | | | |
|---|---|---|---|---|
| | found in Phase 1.<br><br>Scenario A: OS Level Issue (Ping fails, SSH/RDP fails)<br><br>Hardware Check: Work with your infrastructure team. Is the server powered on? Are there any hardware errors reported (RAID failures, memory errors)?<br><br>Virtualization Platform: If virtualized (VMware, Hyper-V, etc.), check the hypervisor for issues with the VM. Is the VM running? Are resources allocated correctly?<br><br>OS Logs: If you can access the OS logs (even remotely via iLO/iDRAC), check for errors related to the OS, drivers, or hardware. Look for kernel panics, disk errors, or other critical messages.<br><br>Scenario B: SAP System Down (SAP System Status Down in SolMan)<br><br>Database Check: Is the database server up and running? Can you connect to the database? (Use DBACOCKPIT or database-specific tools). Database issues are a common cause.<br><br>SAP Startup/Shutdown | provide more specific guidance! | system outage. By following these steps, you can minimize the impact on your operations and ensure a swift recovery. | **Important Notes:**<br><br>* **Documentation:** Document your troubleshooting steps and any solutions implemented for future reference.<br>* **Best Practices:** Follow SAP monitoring best practices to prevent similar incidents in the future (e.g., regular system checks, performance tuning).<br><br>Let me know if you need more specific guidance based on your system logs or error messages. | Let me know if you have more details about your environment or specific error messages. I can provide more tailored advice. |

| | | Logs: Check the SAP startup/shutdown logs for errors. (Located in the SAP system directory, usually /usr/sap/<SID>/SYS/log). Look for messages indicating why the system failed to start.

Trace Files: Enable and analyze SAP trace files. This can provide detailed information about what was happening when the system went down. (Transaction: ST01 - Trace).

Scenario C: SAP Application Server Down (SAP System Up, Server 003 unresponsive)

Work Process Analysis (SM50): Examine SM50 closely. Are work processes terminating unexpectedly? Are there any specific error messages associated with the terminations?

System Log (SM21): Check the SAP system log (SM21) for errors and warnings. Filter by Server 003. Look for messages related to memory allocation, resource limits, or communication errors.

Developer Trace (ST01): If you suspect a specific ABAP program is causing the issue, | | | | |
|---|---|---|---|---|---|---|

| | | enable a developer trace for that program.<br><br>Resource Monitoring (ST06): Check resource usage (CPU, Memory, Disk I/O) on Server 003 using ST06. Is the server overloaded?<br><br>Phase 3: Resolution (Variable Time - Restore Service)<br><br>Scenario A (OS/Hardware): Work with your infrastructure team to resolve the underlying hardware or OS issue. This might involve restarting the server, replacing a failed component, or restoring from a backup.<br><br>Scenario B (SAP System Down):<br><br>Database Recovery: If the database is the issue, work with your DBA to recover the database.<br><br>SAP System Restart: Attempt to restart the SAP system. Monitor the startup logs carefully for errors.<br><br>Scenario C (SAP Application Server Down):<br><br>Work Process Restart: Try restarting the affected work processes in SM50. | | | |
|---|---|---|---|---|---|

| | | Instance Restart: If restarting work processes doesn't help, try restarting the entire SAP instance on Server 003. | | | | |
|---|---|---|---|---|---|---|
| | | Parameter Check: Review SAP instance parameters (Transaction: RZ10) related to memory allocation, resource limits, and communication. Are they configured correctly? | | | | |
| | | Important During Resolution: | | | | |
| | | Document Everything: Keep a detailed log of all actions taken, error messages encountered, and resolutions attempted. | | | | |
| | | Communication: Provide regular updates to stakeholders. Be honest about the progress and estimated time to resolution. | | | | |
| | | Phase 4: Preventative Actions (Post-Incident - Long-Term Stability) | | | | |
| | | Once the system is back up, don't just forget about it! | | | | |
| | | Root Cause Analysis (RCA) Meeting: Hold a meeting with the relevant teams (Basis, Infrastructure, Logistics) | | | | |

| | | | | | |
|---|---|---|---|---|---|
| | to discuss the root cause of the outage.<br><br>Identify Preventative Measures: Based on the RCA, identify steps to prevent a recurrence. Examples:<br><br>Hardware Redundancy: Implement redundant hardware components (RAID, power supplies, network cards).<br><br>Virtualization: Migrate to a virtualized environment for easier failover and resource management.<br><br>Monitoring Enhancements: Add more specific monitoring checks for Server 003 (e.g., disk space thresholds, CPU utilization alerts).<br><br>Capacity Planning: Review resource utilization and plan for future capacity needs.<br><br>Patching: Ensure the OS and SAP kernel are up-to-date with the latest security patches.<br><br>Disaster Recovery: Review and test your disaster recovery plan.<br><br>Document Preventative Measures: Document the preventative measures taken and update your incident | | | | |

| | | management procedures accordingly. | | | | |
|---|---|---|---|---|---|---|
| | | Specific to Dispatching Bay 17: | | | | |
| | | Check Configuration: After the system is back up, verify the configuration of Dispatching Bay 17 in SAP. Ensure it's properly assigned to the correct warehouse and storage locations. | | | | |
| | | Queue Processing: Monitor the delivery queues to ensure that blocked deliveries are being processed correctly. | | | | |
| | | Tools to Leverage: | | | | |
| | | SAP Solution Manager (SolMan): Centralized monitoring, alerting, and root cause analysis. | | | | |
| | | DBACOCKPIT: Database monitoring and administration. | | | | |
| | | ST01 (Trace): Developer trace for ABAP programs. | | | | |
| | | ST06 (Performance Analysis): Resource monitoring. | | | | |
| | | SM21 (System Log): SAP system log. | | | | |
| | | SM50 (Work Process Overview): Work process monitoring. | | | | |

| | | RZ10 (Profile Parameter Maintenance): SAP instance parameter configuration.

Operating System Monitoring Tools: Tools specific to your OS (e.g., Windows Performance Monitor, Linux top, vmstat).

Final Thoughts:

This is a complex situation. Prioritize quickly restoring service, then thoroughly investigate the root cause to prevent future outages. Clear communication and documentation are essential throughout the process. Don't hesitate to escalate to senior experts if you're unsure about any step. Good luck! Let me know if you have any specific details about your environment, and I can refine this guidance further. | | | |
|---|---|---|---|---|---|

| | Test 5 | | | | | |
|---|---|---|---|---|---|---|
| Act as a SAP Monitoring Expert. What do I do if Server 003 is down? SAP runs Logistics Execution - Delivery and Returns on the server. That brings down Dispatching Bay 17. The bay ships high value shipments that can cost $2.4 million in three hours. Your primary task is to Avoid High Cost Downtime | Okay, this is a critical situation. Server 003 being down, impacting Dispatching Bay 17 handling high-value shipments, is a potential disaster. Here's a breakdown of how I, as a SAP Monitoring Expert, would approach this, prioritizing avoiding financial loss (the $2.4 million in 3 hours risk) and getting things back online fast. This is structured into phases: Immediate Response (0-15 mins), Diagnosis (15-30 mins), Recovery (30+ mins), and Preventative Actions. I'll also include communication points.<br><br>Understanding the Severity:<br><br>High Value Shipments: $2.4 million in 3 hours means roughly $800,000/hour is at risk. Every minute of downtime is expensive.<br><br>LE-DEL/Returns Impact: This isn't just a system glitch; it's directly impacting order fulfillment and potentially customer commitments.<br><br>Dispatching Bay 17: Specific bay means likely specialized processes or equipment tied to it. We need to understand if workarounds are | Okay, this is a serious situation! Here's what we need to do quickly and systematically:<br><br>**Immediate Actions:**<br><br>1. **Confirm Server Status:** Verify that Server 003 is truly down. Check physical hardware status (lights, fans), network connectivity, operating system logs, and any monitoring tools you use.<br>2. **Impact Assessment:** Determine the full scope of the outage.<br><br>* **Dispatching Bay 17:** Is it completely offline? Are there specific processes affected?<br>* **Other SAP Modules:** Are any other modules relying on Server 003 experiencing issues? Check your monitoring tools for impacted transactions or performance degradation.<br>* **Business Impact:** How critical is Dispatching Bay 17 to operations? Is there an immediate impact on shipments, customer orders, or revenue streams?<br><br>3. **Emergency Communication:** Inform the relevant | Okay, this is a serious situation! Here's how to approach it as a SAP Monitoring expert:<br><br>**Immediate Actions:**<br><br>1. **Verify System Downtime:**<br><br>* **Check the system status in your monitoring tools (e.g., RZMON, SAP Solution Manager).** Confirm Server 003 is truly down and not just experiencing performance issues.<br>* **Contact the server administrators.** They may be already aware of the issue and working on a resolution.<br>* **Check system logs (e.g., /var/log/messages) for error messages.** Look for clues about the cause of the downtime.<br><br>2. **Isolate the Impact:**<br><br>* **Confirm Dispatching Bay 17 is indeed down.** This could be through:<br>* Direct contact with dispatch staff.<br>* Checking relevant SAP transactions for errors or inactivity.<br>* Reviewing the system's operational status in your monitoring tools.<br><br>3. **Communicate:** | Alright, let's get this resolved! Here's my breakdown of what to do if Server 003 is down and impacting your Dispatching Bay 17:<br><br>**Immediate Actions:**<br><br>1. **Verify the Outage:**<br>* **System Logs:** Check the system logs on Server 003 for any error messages or recent events that might pinpoint the cause of the outage.<br>* **SAP Instance Status:** Use transaction SM51 to check the status of your SAP instance running on Server 003. Is it down completely?<br><br>2. **Communicate:**<br>* **Internal Teams:** Notify relevant teams immediately: IT operations, Basis administrators, and logistics personnel responsible for Dispatching Bay 17. Explain the situation clearly.<br>* **Users:** Inform users affected by the outage in Dispatching Bay 17 about the downtime and expected resolution time (if you have an estimate).<br><br>**Troubleshooting Steps:**<br><br>3. **Isolate the** | Ah, sounds like a critical situation! Let's get Server 003 back online ASAP. As your friendly neighborhood SAP Monitoring Expert, here's what I recommend:<br><br>**Immediate Actions:**<br><br>1. **Identify the Downtime:** Confirm if it's truly a server outage or just an application issue within Logistics Execution - Delivery and Returns (LE-D&R).<br>* **Check System Logs:** Look for error messages in the SAP system logs on Server 003, as well as any related applications.<br>* **Monitor Network Connectivity:** Ensure Server 003 is reachable via network ping and traceroute.<br><br>2. **Alert Relevant Teams:**<br><br>* Notify your **System Administrators** immediately to begin troubleshooting the server issue.<br>* Inform your **Logistics Team** about the Dispatching Bay 17 downtime, explaining its impact on operations.<br><br>3. **Fallback Procedures:** Activate any existing fallback procedures for Dispatching Bay 17: |

| | | | | |
|---|---|---|---|---|
| | | possible to reroute shipments.<br><br>Phase 1: Immediate Response (0-15 Minutes) - CONTAINMENT & INITIAL ASSESSMENT<br><br>1. Verify the Outage: Don't rely on a single alert. Confirm Server 003 is truly down. Use SAP Solution Manager (SolMan) if available, or direct OS-level checks (ping, SSH if possible).<br><br>2. Activate the Incident Management Process: This is not a drill. Immediately escalate through your pre-defined incident management system. This should involve:<br><br>Team Notification: Notify the Basis team, Logistics Execution (LE) functional team, and the relevant business stakeholders (Dispatch Manager, potentially a Logistics Director). Clear, concise communication: "Server 003 DOWN. Impacting Dispatch Bay 17, high-value shipments. Potential $2.4M loss in 3 hours."<br><br>Incident Ticket Creation: Document everything. Time stamps are critical.<br><br>3. Initial Impact | stakeholders:<br><br>* **IT Team:** Escalate the issue and provide them with all available details about the server failure.<br>* **Business Operations:** Alert your logistics and dispatching teams about the outage and potential delays.<br>* **Customers:** If possible, proactively communicate any expected delays to affected customers.<br><br>**Troubleshooting Steps (while IT investigates):**<br><br>4. **Check System Logs:** Examine the logs on Server 003 (if accessible) for error messages or clues about the cause of the failure.<br>5. **Network Connectivity:** Ensure there are no network outages affecting Server 003. Check routers, switches, and firewalls for any issues.<br>6. **Resource Utilization:** Monitor CPU, memory, and disk usage on related servers to see if there's an overload impacting performance.<br><br>**Remediation and Recovery:**<br><br>7. **IT Intervention:** The IT team will need to | * **Notify relevant stakeholders immediately:** Operations, logistics managers, customer service, and IT management need to be informed about the outage and its potential impact on deliveries and customer satisfaction.<br>* **Prepare a concise status report outlining the situation, including:**<br>* Server downtime confirmation.<br>* Affected applications (Logistics Execution - Delivery and Returns).<br>* Downtime duration (if known).<br>* Estimated resolution time (if possible).<br>* Actions being taken to restore service.<br><br>**Troubleshooting & Recovery:**<br><br>4. **Investigate the Cause:** Once server administrators have access to investigate, identify the root cause of the downtime:<br><br>* Hardware failure (e.g., CPU, memory, disk)<br>* Software issue (e.g., operating system crash, SAP kernel error)<br>* Network connectivity problems<br>* Security incident (e.g., malware attack)<br><br>5. **Take Corrective Actions:** Based on the | Problem:**<br>* **Hardware Check:** Ensure Server 003 itself is operational: power, network connectivity, physical hardware health. Consult your IT infrastructure team for assistance here if needed.<br>* **Operating System:** Check the operating system logs on Server 003 for any errors related to SAP or general system issues.<br>* **SAP Instance:** If the instance itself appears down, review the SAP startup log files (SM21) for detailed error messages during the startup process.<br><br>4. **Review Recent Changes:**<br>* **Software Updates/Patches:** Were there any recent software updates or patches applied to Server 003 or related systems that could be causing conflicts?<br>* **Configuration Changes:** Have any configuration settings been modified recently in SAP or on the server that might have impacted Dispatching Bay 17 functionality?<br><br>**Recovery Options:**<br><br>5. **Restart Server/Instance:** If appropriate, attempt to | * Can other dispatching bays handle the increased workload temporarily?<br>* Are there manual processes to manage deliveries while LE-D&R is down?<br><br>**Troubleshooting Server 003:**<br><br>1. **System Status Checks:** Use SAP Monitoring tools (like RZ20) to check the server's overall health:<br>* CPU utilization, Memory consumption, Disk I/O activity<br>2. **Instance Logs:** Analyze the instance logs on Server 003 for clues about the cause of the outage. Look for specific errors or warnings.<br>3. **Hardware Checks:** If software troubleshooting doesn't yield results, check hardware components:<br>* Power supply, RAM modules, Network interfaces<br><br>**Recovery & Post-Mortem:**<br><br>1. **Bring Server 003 Online:** Once the issue is identified and resolved by your System Administrators, restart the server (if needed) and monitor its performance closely.<br>2. **Post-Mortem** |

| | | Assessment:<br><br>How many shipments are currently staged in Bay 17? Get a quick count.<br><br>What is the status of those shipments? (e.g., Picking complete, Packing in progress, Awaiting dispatch).<br><br>Are there any alternative dispatch bays that could handle these shipments, even with reduced efficiency? (This is the first line of defense against the financial loss). This requires immediate communication with the Dispatch Manager.<br><br>4. Communication - First Update: Send a brief update to stakeholders: "Server 003 is down. We are investigating the cause and exploring options to minimize disruption to Dispatch Bay 17. Initial assessment underway."<br><br>Phase 2: Diagnosis (15-30 Minutes) - FIND THE ROOT CAUSE<br><br>1. Basis Team Focus: The Basis team takes the lead on technical diagnosis. Key areas to investigate concurrently:<br><br>OS Level: Check the server's operating | diagnose the root cause of the server failure and implement a solution. This could involve:<br>* Restarting the server<br>* Replacing faulty hardware components<br>* Reinstalling the operating system or SAP software<br>* Migrating data to another server<br><br>**Post-Outage Actions:**<br><br>8. **Root Cause Analysis:** Once the issue is resolved, conduct a thorough investigation to determine the underlying cause of the server failure. This will help prevent similar incidents in the future.<br>9. **Documentation and Training:** Update your incident management procedures and train personnel on best practices for responding to server outages.<br><br>**Remember:**<br><br>* Communication is key throughout the entire process. Keep stakeholders informed about the situation, progress updates, and potential impact.<br>* Prioritize business continuity. Take steps to minimize disruption to operations and customer | cause:<br>* Replace faulty hardware components.<br>* Install software patches or hotfixes.<br>* Restore from backups.<br>* Implement security measures to prevent future incidents.<br><br>6. **Restore Service:** Once the issue is resolved, carefully restart Server 003 and verify system functionality.<br><br>**Preventative Measures (for the Future):**<br><br>* **High Availability (HA) Setup:** Implement an HA solution for critical SAP servers like Server 003. This ensures automatic failover to a standby server in case of downtime.<br>* **Regular System Backups:** Maintain regular and reliable backups to enable quick system recovery.<br>* **Monitoring & Alerting:** Configure robust monitoring systems with alerts for critical events. This allows for proactive identification and resolution of issues before they cause major disruptions.<br>* **Disaster Recovery Plan (DRP):** Develop a comprehensive DRP that outlines procedures for recovering from | restart the server or SAP instance. This can often resolve temporary issues.<br>6. **Manual Intervention:** Depending on the specific error and your SAP expertise, you might need to perform manual steps within the SAP system to recover from the outage (e.g., restarting background jobs, reconfiguring connections).<br>7. **Disaster Recovery Plan:** If the issue persists or is severe, activate your disaster recovery plan for Logistics Execution - Delivery and Returns if one exists. This could involve switching to a backup system or utilizing alternative processing methods.<br><br>**Important Notes:**<br><br>* **Document Everything:** Keep detailed records of all steps taken, errors encountered, and any communication with involved teams.<br>* **Escalate Promptly:** If you're unable to resolve the issue independently, escalate it to your SAP support team or higher-level IT personnel. They have specialized knowledge | Analysis:** Conduct a thorough review after the incident:<br>* What caused the downtime?<br>* Were there any warning signs we missed?<br>* How can we prevent similar incidents in future?<br><br>**Key Reminders:**<br><br>* Stay calm and focus on troubleshooting systematically.<br>* Communicate clearly and promptly with all relevant teams.<br>* Leverage SAP Monitoring tools effectively for diagnostics.<br>* Document everything for future reference and analysis.<br><br>Let me know if you have any specific questions or need further assistance! |

| | | | | |
|---|---|---|---|---|
| | system logs (Windows Event Viewer, Linux syslog) for errors, crashes, or hardware failures.<br><br>SAP Instance Status: Use SM51 (SAP Instance Overview) in another SAP system (if accessible) to see if the instance is stopped or crashed.<br><br>Database Connection: Verify the database connection is active. Check database logs for errors.<br><br>Network Connectivity: Ping the server from other systems. Check network devices for issues.<br><br>Recent Changes: Were there any recent SAP system changes (patches, upgrades, configuration changes) that might have triggered the issue?<br><br>2. SAP Monitoring Tools: Leverage SolMan (if available) for:<br><br>System Logs: Detailed SAP system logs.<br><br>Performance Monitoring: Look for any performance spikes or anomalies leading up to the outage.<br><br>Alert History: Review | service.<br><br>Let me know if you have any specific questions or need further guidance during this process. | various disaster scenarios, including server failures.<br><br><br>Remember: Clear communication and swift action are crucial in minimizing the impact of server downtime on your SAP environment and business operations. | and tools to assist with complex outages.<br><br><br>I hope this helps guide you through this situation. Remember, clear communication, systematic troubleshooting, and appropriate escalation are key to minimizing downtime and restoring normal operations. | |
| | recent alerts for related issues.<br><br>3. LE Functional Team Input: While Basis is focused on the technical side, the LE team should be thinking:<br><br>Are there any known issues with the LE configuration that could contribute to a server crash? (Unlikely, but worth asking).<br><br>What is the impact on specific orders? Can we identify critical orders that must be shipped?<br><br>4. Communication - Second Update: "Basis team is investigating the root cause. Initial checks are focused on [mention key areas - OS, database, network]. We are assessing the impact on specific shipments."<br><br>Phase 3: Recovery (30+ Minutes) - GET THINGS BACK ONLINE<br><br>This phase depends heavily on the root cause. Here are common scenarios and actions:<br><br>Scenario 1: Server Hardware Failure:<br><br>Failover: If the server is part of a high-availability (HA) cluster, initiate a | | | | |
| | failover to the standby server. This is the fastest recovery.<br><br>Hardware Replacement: If no HA, begin the process of replacing the failed hardware. This will take longer.<br><br>Scenario 2: OS/Database Crash:<br><br>Restart: Attempt a controlled restart of the SAP instance and database. Monitor closely for errors during startup.<br><br>Database Recovery: If the database requires recovery, follow your database administrator's procedures.<br><br>Scenario 3: SAP Software Issue:<br><br>Rollback: If a recent SAP change is suspected, roll back the change.<br><br>Apply SAP Note: If a known SAP bug is identified, apply the relevant SAP Note.<br><br>Scenario 4: Network Issue:<br><br>Network Troubleshooting: Work with the network team to resolve the network connectivity problem. | | | | |

| | During Recovery:<br><br>Continuous Monitoring: Monitor the server's performance and stability closely after restart.<br><br>Data Integrity Checks: Run data consistency checks to ensure no data corruption occurred.<br><br>Prioritize Critical Processes: Once the server is back online, prioritize the restart of the LE processes needed for Dispatch Bay 17.<br><br>Communication - Regular Updates: Provide frequent updates to stakeholders (every 15-30 minutes) on the recovery progress. Be realistic about timelines. "Server restart initiated. Monitoring closely. Estimated time to full recovery: [time]."<br><br>Phase 4: Preventative Actions - AVOID RECURRENCE<br><br>1. Root Cause Analysis (RCA): After the immediate crisis is over, conduct a thorough RCA to determine the underlying cause of the outage. Don't just fix the symptom; fix the problem. | | | | |
| --- | --- | --- | --- | --- | --- |

| | | 2. Strengthen Monitoring:

Proactive Monitoring: Implement more proactive monitoring for Server 003, including:

Hardware Health Checks: Monitor CPU, memory, disk space, and network utilization.

SAP Application Monitoring: Monitor key LE processes and transactions.

Database Monitoring: Monitor database performance and availability.

Alert Thresholds: Adjust alert thresholds to provide earlier warnings of potential issues.

3. Improve High Availability:

HA/DR: If not already in place, consider implementing a high-availability (HA) or disaster recovery (DR) solution for Server 003.

4. Change Management: Review and improve your change management process to ensure that all SAP system changes are thoroughly tested and documented. | | | | |

| | 5. Documentation: Update your incident response plan and documentation to reflect the lessons learned from this outage.

Key Considerations & Tools:

SAP Solution Manager (SolMan): Essential for comprehensive monitoring, alerting, and root cause analysis.

Operating System Monitoring Tools: Windows Performance Monitor, Linux top, vmstat, etc.

Database Monitoring Tools: Database-specific tools for monitoring performance and availability.

Clear Communication Channels: Establish clear communication channels and escalation procedures.

Regular Backups: Ensure you have regular and reliable backups of your SAP system.

To help me refine this further, could you tell me:

Do you have SAP Solution Manager implemented?

Is Server 003 part of a | | | | |
|---|---|---|---|---|

| | | high-availability cluster?

What operating system is Server 003 running?

What database is being used?

What is your current incident management process?

This is a stressful situation, but a calm, methodical approach, focused on minimizing financial impact and restoring service quickly, is crucial. Let me know if you'd like me to elaborate on any of these points. | | | |
|---|---|---|---|---|---|

| | Test 6 | | | | |
|---|---|---|---|---|---|
| Act as a SAP Monitoring Expert. What do I do if Server 003 is down? SAP runs Logistics Execution - Delivery and Returns on the server. That brings down Dispatching Bay 17. The bay ships high-value shipments that can cost $2.4 million in three hours. Increasing the ID range on server 003 would have prevented the downtime. Your primary task is to Avoid High-Cost Downtime. | Okay, this is a critical situation. High-value shipments and a potential $2.4 million loss in three hours demands immediate and decisive action. Here's a breakdown of how I, as a SAP Monitoring Expert, would approach this, focusing on minimizing downtime right now and then preventing recurrence. I'll categorize this into Immediate Response (Now - 30 mins), Short-Term Recovery (30 mins - 2 hours), and Long-Term Prevention (2+ hours). I'll also include a Communication Plan.

Understanding the Severity:

Critical Business Impact: Logistics Execution (LE-DEL) is down, specifically impacting Dispatching Bay 17. This isn't a minor issue; it's directly halting revenue generation and potentially causing significant contractual penalties.

Root Cause Indication: You've already identified a likely root cause: insufficient ID range on Server 003. This is extremely valuable information. It suggests a configuration issue, not necessarily a | As a SAP Monitoring Expert, here's my immediate action plan for handling Server 003 downtime impacting Dispatching Bay 17:

**Immediate Actions (Focus: Minimizing Financial Impact):**

1. **Confirm Downtime Scope:**
* Verify if only Server 003 is down or other critical systems are affected.
* Determine the exact point of failure on Server 003 and if it's a hardware, software, or network issue.
2. **Contact Relevant Teams:**
* **IT Operations:** Escalate the issue to your IT operations team for immediate troubleshooting and resolution. Provide detailed information about the server and the impact on Dispatching Bay 17.
* **Dispatch Team:** Inform the dispatch team about the downtime and potential delays.
3. **Mitigate Immediate Impacts:**
* **Manual Workarounds:** If possible, implement manual workarounds to process urgent shipments until Server 003 is back online. This might involve using | This is a critical situation! Here's my action plan as a SAP Monitoring Expert focused on avoiding high-cost downtime:

**Immediate Actions (Within the Next 15 Minutes):**

1. **Confirm Server Status:** Verify that Server 003 is truly down and identify the exact cause. Is it a hardware failure, software issue, network connectivity problem, or something else? Use monitoring tools like RZMON, SMON, ST22 (for ABAP dumps) to gather details.
2. **Activate Disaster Recovery Plan:** If you have a DR plan in place for Logistics Execution - Delivery and Returns, initiate it immediately. This may involve switching over to a backup server or utilizing failover mechanisms. The speed of your DR plan execution is crucial here!
3. **Communication & Escalation:**
* Notify the relevant teams: IT Operations, SAP Basis team, Dispatching Bay 17 management.
* Clearly communicate the situation, the impact on high-value shipments, and the estimated time to | This situation demands immediate action! Here's a breakdown of what I, as your SAP Monitoring Expert, would do:

**1. Immediate Actions:**

* **Contact Server Admin Team:** Get them on the phone ASAP. They need to understand the urgency and prioritize resolving Server 003 issue. Provide details about the server role (Logistics Execution - Delivery & Returns) and the critical impact on Dispatching Bay 17.
* **Activate Backup System (if available):** If you have a disaster recovery plan and a backup system for Logistics Execution, initiate it immediately. This might allow partial functionality while Server 003 is being fixed. Even partial functionality could mitigate some of the losses.

* **Communicate with Shipping Team:** Inform them about the downtime and the potential impact on high-value shipments. Explore options like temporarily diverting shipments to another bay or delaying them if possible, minimizing | Okay, this is a critical situation! Here's a breakdown of how I'd approach this as a SAP Monitoring Expert, focusing on minimizing the financial impact:

**Immediate Actions (First 30 Minutes):**

1. **Confirm the Issue:**
- Double-check that Server 003 is truly down and not experiencing performance issues. Use monitoring tools (e.g., SMON, RZMON) to verify system status and identify potential causes (disk space, memory pressure, etc.).
2. **Isolate the Impact:**
- Determine exactly which SAP modules are impacted on Server 003 besides Logistics Execution - Delivery & Returns. This will help prioritize recovery efforts.
3. **Communicate Effectively:**
- Alert relevant stakeholders: dispatch managers, shipping personnel, customer service, finance department. Explain the situation clearly and concisely. Emphasize the potential financial impact.
4. **Activate Contingency Plan (if available):**
- If a Disaster Recovery plan exists for this |

| | | hardware failure (though we can't rule that out completely yet). | paper-based systems or alternative communication channels. | resolution. | financial exposure. | scenario, execute it immediately. This might involve switching to backup servers or using alternative systems. |
|---|---|---|---|---|---|---|
| | | Time Sensitivity: The $2.4 million loss window is the driving factor. Every minute counts. | * **Prioritize Shipments:** Focus on processing high-value shipments that are time-sensitive. Communicate clearly with customers about potential delays and offer alternative delivery options if feasible. | * Keep stakeholders updated with regular status reports. **Mitigation & Resolution (Within the Next Hour):** 4. **Troubleshooting:** While activating DR, simultaneously troubleshoot the root cause of Server 003's failure. This might involve: * Analyzing system logs and error messages. * Checking hardware health (CPU, memory, disk). * Investigating network connectivity issues. | * **Monitor the Situation Closely:** Keep a close eye on system status updates from both the server admin team and SAP monitoring tools. **2. Addressing the Root Cause:** * **ID Range Exhaustion:** You mentioned that increasing the ID range on Server 003 could have prevented this downtime. This is a critical lesson learned. We need to implement procedures to proactively monitor system resources, including ID ranges, and trigger alerts when nearing exhaustion thresholds. | **Short-Term Mitigation (Next 1-2 Hours):** 5. **Manual Dispatching:** - Train dispatch personnel on manual processes for entering shipment information and generating labels if the system is completely unavailable. This will be less efficient but vital for moving shipments. 6. **Priority Support:** - Escalate the issue to SAP support or your internal IT team. Request urgent assistance with server recovery or workarounds. |
| | | I. IMMEDIATE RESPONSE (Now - 30 mins) - Stabilize & Redirect | **Troubleshooting & Recovery (Focus: Preventing Future Downtime):** | 5. **Temporary Workarounds:** If a full recovery is taking longer than expected, explore temporary workarounds to minimize disruption in Dispatching Bay 17. This could involve: | **3. Long-Term Solutions:** | 7. **Customer Communication:** - If possible, proactively notify high-value shipment customers about potential delays and keep them updated on the situation. Transparency can minimize customer dissatisfaction. |
| | | This phase is about damage control. We need to get something shipping, even if it's not ideal. | 1. **Root Cause Analysis:** Once Server 003 is back online, conduct a thorough investigation to determine the root cause of the downtime. | * Manually processing some shipments until the system is back online. * Utilizing alternative shipping methods or temporarily diverting high-value shipments to another bay. | * **Implement Proactive Monitoring:** Set up robust monitoring tools that track critical SAP system resources like database space, memory usage, CPU utilization, and ID range availability. | **Long-Term Prevention (After Recovery):** |
| | | Confirm Downtime & Scope: | 2. **System Logs Review:** Analyze system logs on Server 003 and related systems for error messages or patterns that indicate potential issues. | **Long-Term Prevention (After Server Restoration):** | * **Automate Alerts:** Configure automated alerts to notify the relevant teams when these resource thresholds are approaching limits. This allows for timely | 8. **Root Cause Analysis:** - Conduct a thorough investigation to determine *why* Server |
| | | Verify: Double-check the server status (SAP GUI, Solution Manager, OS level). Is it truly down, or is it a network issue? Don't assume. | 3. **Performance Monitoring:** Check recent performance metrics for Server 003 to identify any bottlenecks or resource constraints that might have contributed to the failure. | 6. **Increase ID Range:** Implement the necessary steps to increase the ID range on | | |
| | | Impact Assessment: Confirm Dispatch Bay 17 is completely halted. Are there any manual workarounds being attempted? How many shipments are currently queued? | **Long-Term Prevention (Focus: System Resilience):** | | | |
| | | Alerting: Ensure all relevant stakeholders are alerted (see Communication Plan below). | 1. **ID Range | | | |
| | | Emergency Redirect (Priority #1): | | | | |

| | | | | |
|---|---|---|---|---|
| | | Dispatch Bay Failover: Immediately attempt to redirect shipments from Dispatch Bay 17 to another available dispatching bay. This is the fastest way to mitigate the financial loss. This might involve:<br><br>Configuration Change: Adjusting dispatching parameters in SAP to route shipments to a different bay. (Requires a knowledgeable LE-DEL consultant).<br><br>Manual Intervention: If configuration changes are too slow, authorize a temporary manual rerouting process (with clear documentation and audit trails).<br><br>Consider Partial Processing: Can some aspects of the shipments be processed manually (e.g., labeling, documentation) while waiting for the system to come back online?<br><br>Initial Diagnostics (Parallel to Redirect):<br><br>SAP System Log (SM21): Check for critical errors immediately preceding the downtime. Look for clues related to ID ranges, database issues, or resource constraints. | Management:**<br>Implement a proactive strategy for managing ID ranges on all critical SAP servers. Regularly monitor available IDs and plan for future growth to avoid similar downtime scenarios.<br>2. **High Availability Planning:** Evaluate your current high availability setup for SAP systems. Consider implementing solutions like clustering, mirroring, or failover mechanisms to minimize the impact of future server failures.<br>3. **Monitoring & Alerting:** Enhance monitoring capabilities for all critical SAP infrastructure components. Configure alerts for potential issues and ensure that relevant teams are notified promptly.<br>4. **Disaster Recovery Plan:** Develop and test a comprehensive disaster recovery plan specifically for SAP systems. This should include procedures for data backups, system restoration, and business continuity.<br><br>**Remember:**<br><br>* Time is of the essence when dealing with high-cost downtime. Act quickly and decisively to mitigate financial losses and minimize customer | Server 003 *immediately*. This prevents future downtime scenarios caused by ID exhaustion.<br>7. **Review Monitoring & Alerts:** Ensure your monitoring system is configured to effectively detect and alert you about potential issues like low ID space, resource contention, or server performance degradation.<br>8. **Capacity Planning:** Conduct a thorough capacity planning exercise for SAP Logistics Execution - Delivery and Returns. This helps identify future resource requirements and proactively address potential bottlenecks.<br><br>**Remember:**<br><br>* **Time is of the essence:** Every minute counts in minimizing the impact on high-value shipments. Act decisively and prioritize tasks accordingly.<br>* **Communication is key:** Keep all stakeholders informed and updated throughout the process.<br>* **Learn from this incident:** Analyze the root cause to prevent similar situations in the future. Implement preventive measures, | intervention before a complete system failure occurs.<br><br>* **Regular Capacity Planning:** Conduct regular capacity planning exercises to forecast future resource requirements based on anticipated workload growth and implement appropriate scaling measures (e.g., adding server resources, increasing ID ranges).<br>* **Resource Pooling:** Explore options for centralizing resources (like database connections) and implementing resource pooling strategies to optimize utilization across multiple SAP systems.<br><br>**Remember:** This is a high-stakes situation. While you're working on resolving the immediate problem, use this as a learning opportunity to strengthen your monitoring processes and prevent similar incidents in the future.<br><br>Let me know if you need help with specific steps or have any further questions! | 003 went down. Was it due to ID range exhaustion, hardware failure, software bug, or another factor? This analysis is crucial for preventing future incidents.<br>9. **ID Range Management:**<br>- Implement stricter policies and automated processes to prevent ID range exhaustion on critical servers. Explore using larger ID ranges or tools that manage ID allocation effectively.<br>10. **Monitoring Enhancements:**<br>- Review your current monitoring system. Consider adding more granular alerts for potential issues related to server performance, disk space, memory usage, and SAP-specific bottlenecks.<br><br>**Key Points:**<br><br>* **Time is Money:** Your primary focus is minimizing downtime and financial losses. Act decisively and communicate transparently throughout the process.<br>* **Preventative Measures:** This incident highlights the importance of proactive monitoring, robust disaster recovery plans, and proper ID range management. |

| | | | | |
|---|---|---|---|---|
| | OS Level Logs: Check the operating system logs on Server 003 for hardware errors, memory issues, or other system-level problems.<br><br>Database Status: Verify the database associated with Server 003 is running and accessible.<br><br>II. SHORT-TERM RECOVERY (30 mins - 2 hours) - Restore Service<br><br>This phase focuses on getting Server 003 back online and Dispatch Bay 17 functioning normally.<br><br>Server Restart (If Appropriate): If the initial diagnostics don't point to a critical hardware failure, attempt a controlled restart of Server 003. Do not simply reboot without investigating!<br><br>ID Range Investigation & Correction (Critical):<br><br>Confirm ID Range Exhaustion: Use transaction SM12 (Lock Entries) to see if there are locks related to the ID range. This will confirm if the exhaustion of the ID range is the actual cause.<br><br>Increase ID Range: This is the fix you already know about. Increase | impact.<br>* Thorough documentation and communication are crucial throughout the process. Keep stakeholders informed about the situation, progress, and any potential impact on their operations.<br><br>By following these steps, you can effectively handle this server downtime incident while taking proactive measures to prevent similar situations in the future. | review your DR plan, and strengthen your monitoring capabilities.<br><br>Let me know if you have any specific questions about SAP monitoring tools or procedures. I'm here to help! | | Let me know if you have any other questions or need more specific guidance. |

<table>
<tr><td></td><td>the ID range for the relevant document type (likely Delivery Document) in configuration (Transaction SPRO -> Logistics Execution -> Warehouse Management -> Cross-Process Settings -> Delivery -> Define Number Ranges for Delivery Documents). Important: Coordinate this change with your Basis team to ensure it doesn't conflict with other systems.<br><br>Release Locks: After increasing the ID range, attempt to release any locks that were preventing new deliveries from being created.<br><br>Monitor System Performance: After the restart and ID range correction, closely monitor Server 003's performance (CPU, memory, disk I/O) using Solution Manager or other monitoring tools. Look for any signs of instability.<br><br>Test Dispatch Bay 17: Once the server is stable, thoroughly test Dispatch Bay 17 to ensure it's processing shipments correctly. Start with a small batch of test shipments before releasing it for full</td><td></td><td></td><td></td><td></td></tr>
</table>

| | | | | | |
|---|---|---|---|---|---|
| | production.<br><br>III. LONG-TERM PREVENTION (2+ hours) - Prevent Recurrence<br><br>This phase is about preventing this from happening again.<br><br>Root Cause Analysis (RCA): Conduct a thorough RCA to understand why the ID range was exhausted in the first place. Was it:<br><br>Incorrect Initial Configuration?<br><br>Unexpected Business Growth?<br><br>Inefficient Number Range Usage? (e.g., gaps in the sequence)<br><br>Missing Monitoring? (We should have been alerted before the range was exhausted).<br><br>Proactive Monitoring:<br><br>Implement ID Range Monitoring: Set up alerts in Solution Manager (or your monitoring tool) to notify you when the ID range reaches a predefined threshold (e.g., 80% full). This is crucial.<br><br>Capacity Planning: Review your overall SAP capacity planning to | | | | |
| | ensure your ID ranges are appropriately sized for future growth.<br><br>Documentation & Procedures:<br><br>Document the ID Range Increase Procedure: Create a clear, step-by-step guide for increasing ID ranges, including who needs to be involved and what testing needs to be performed.<br><br>Disaster Recovery Plan Update: Update your disaster recovery plan to include specific procedures for handling ID range exhaustion.<br><br>Automated ID Range Management (Consider): Explore options for automating ID range management, such as using SAP's number range object maintenance tools.<br><br>IV. COMMUNICATION PLAN<br><br>Immediate (Within 5 mins of confirmation):<br><br>IT Director/CIO: Brief overview of the situation, potential financial impact, and immediate actions being taken.<br><br>Logistics Manager: Impact on shipping, redirect efforts, and | | | | |

| | | estimated recovery time. | | | | |
| | | | | | | |
| | | SAP Basis Team: Server status, restart requests, ID range correction. | | | | |
| | | | | | | |
| | | LE-DEL Consultant: Dispatch bay redirect assistance, configuration changes. | | | | |
| | | | | | | |
| | | Regular Updates (Every 30 mins): | | | | |
| | | | | | | |
| | | All stakeholders: Progress on recovery, estimated time to resolution, and any changes to the impact assessment. | | | | |
| | | | | | | |
| | | Post-Incident Report: | | | | |
| | | | | | | |
| | | Detailed RCA: Document the root cause, corrective actions, and preventative measures. Share with all stakeholders. | | | | |
| | | | | | | |
| | | Key Tools: | | | | |
| | | | | | | |
| | | SAP GUI: Transactions like SM21, SM12, SPRO. | | | | |
| | | | | | | |
| | | Solution Manager: Monitoring, alerting, RCA. | | | | |
| | | | | | | |
| | | Operating System Logs: For server-level diagnostics. | | | | |
| | | | | | | |
| | | Database Monitoring Tools: To check database status. | | | | |

| | | | | | |
|---|---|---|---|---|---|
| | | Important Considerations: | | | |
| | | Change Management: Any configuration changes should be made through a proper change management process. | | | |
| | | Testing: Thorough testing is essential before releasing any changes to production. | | | |
| | | Collaboration: Effective communication and collaboration between IT, Logistics, and SAP consultants are critical for a successful recovery. | | | |
| | | This is a high-pressure situation. Prioritize stabilizing the business, restoring service, and then preventing recurrence. Let me know if you have any specific questions or need further clarification on any of these steps. I'm here to help you navigate this crisis. | | | |
| | | | | | |
| **Super Prompts** | | | | | |

| Act as a SAP Monitoring Expert at a Chemical Manufacturing Company. Given the context that server-003 runs the LE-DEL (Logistics Execution - Delivery and Returns) module for SAP, the impact of the server being offline can be significant for business operations. Here's how it affects various aspects of the system:<br><br>Disruption to Delivery and Returns Processes: Outbound Shipments: If the LE-DEL module is offline, the entire outbound logistics process could be disrupted, meaning that orders may not be processed or shipped on time. This can cause delays in fulfilling customer orders, potentially impacting customer satisfaction. Inbound Returns: Any returns from customers might not be processed correctly, causing inventory discrepancies and delays in restocking returned goods. Warehouse Operations: Warehouse Efficiency: The LE-DEL module is integral for coordinating the delivery of goods from the warehouse. If the server is down, warehouse staff may not have real-time access to delivery information, potentially leading to inefficiencies, mistakes, or confusion in picking, packing, and shipping goods. Inventory Management: The delay in processing deliveries and returns can | Test 7 | Okay, here's my response as a SAP Monitoring Expert at this Chemical Manufacturing Company, addressing the server-003/LE-DEL outage. This is a critical situation, and my response will cover immediate actions, investigation, communication, and preventative measures. I'll structure it as if I'm reporting to the IT Manager and key business stakeholders.<br><br>Subject: URGENT - Server-003 (LE-DEL) Outage - Critical Business Impact<br><br>To: IT Manager, Logistics Manager, Warehouse Manager, Customer Service Manager, Supply Chain Manager<br><br>From: [Your Name], SAP Monitoring Expert<br><br>Date: October 26, 2023<br><br>Executive Summary:<br><br>Server-003, hosting the SAP LE-DEL (Logistics Execution - Delivery and Returns) module, is currently offline. This is a critical incident with significant impact on outbound shipments, inbound returns, warehouse operations, customer experience, | You've provided a comprehensive analysis of the impact of server-003 being offline on our chemical manufacturing operations. I agree with your assessment – this situation has the potential for significant disruption across multiple areas.<br><br>As the SAP Monitoring Expert, here's how I would prioritize addressing this situation:<br><br>**Immediate Actions:**<br><br>1. **Incident Management:** Initiate an incident management process to formally document the issue, assign responsibilities, and track progress.<br>2. **Root Cause Analysis (RCA):** Immediately engage the IT team to diagnose the cause of the server outage. Is it a hardware failure, software bug, network connectivity problem, or something else?<br>3. **Restoration:** Work with the IT team to restore server-003 as quickly as possible. If a failover mechanism exists, activate it immediately to minimize downtime.<br><br>**Communication and Stakeholder | This is a comprehensive analysis of the impact of server-003 being offline on your chemical manufacturing company's operations. As an SAP Monitoring Expert, I would add these points and recommendations:<br><br>**Immediate Actions:**<br><br>* **Prioritize Recovery:** The top priority is getting server-003 back online as quickly as possible. This requires identifying the root cause of the outage (hardware failure, software glitch, network issue) and taking immediate steps to resolve it.<br>* **Activate Failover:** If a failover mechanism exists for LE-DEL, activate it immediately to minimize downtime. Regularly test failover procedures to ensure they function smoothly in real-world scenarios.<br><br>**Communication & Mitigation:**<br><br>* **Transparent Communication:** Inform all stakeholders (logistics, warehouse, sales, customer service, upper management) about the outage and its potential impact on operations. Set realistic expectations about restoration timeframes. | You've accurately outlined the significant impact of server-003 going offline on our chemical manufacturing company. As your SAP Monitoring Expert, I would immediately initiate the following actions:<br><br>**Immediate Actions:**<br><br>1. **Identify the Root Cause:** Dispatch a team to investigate the cause of the outage – hardware failure, software crash, network connectivity issue, or any other potential reason.<br>2. **Restoration Efforts:** Depending on the root cause, restore server-003 functionality. If a simple fix is possible (e.g., network reconnection), implement it immediately. For more complex issues, engage our IT specialists and leverage documented disaster recovery procedures.<br><br>**Mitigating the Impact:**<br><br>1. **Activate Failover Mechanism:** If we have a configured failover system, switch to the backup server hosting the LE-DEL module as quickly as possible to minimize downtime. | This is an excellent analysis of the impact of server-003 being offline for a chemical manufacturing company reliant on SAP's LE-DEL module. You've clearly outlined the domino effect across various departments and functions, highlighting both operational and financial risks.<br><br>Here are some additional points to consider and actions to take, building upon your already comprehensive assessment:<br><br>**Immediate Actions:**<br><br>* **Priority Recovery:** As you mentioned, immediate investigation and restoration of server-003 are paramount. This might involve contacting IT support, hardware specialists, or SAP experts depending on the root cause.<br>* **Communication Plan:** A clear and concise communication plan should be implemented to keep stakeholders informed about the situation's status, estimated recovery time, and any potential impact on their operations.<br>* This includes:<br>* Internal teams |
| | | | | | |

lead to inaccurate inventory data, as goods are not being tracked or updated in the system in real time.
Customer Experience: Order Delays: If deliveries can't be processed through SAP due to the server being offline, it directly impacts the customer's experience. Orders may be delayed, and returns might not be handled in a timely manner.
Customer Service: Customer service teams may struggle to provide accurate information to customers about the status of their orders or returns due to a lack of data from the LE-DEL system.
Operational Impact: Manual Workarounds: In the absence of the LE-DEL module functioning, manual processes might need to be implemented, which are less efficient and error-prone. These could involve manually tracking deliveries and returns in spreadsheets or other systems, further complicating the process.
Additional Workload: Teams responsible for deliveries and returns will likely experience a higher workload due to the disruption, leading to potential bottlenecks and delays in fulfilling other tasks.
Reporting and Analytics: Lack of Real-Time Data: Key performance indicators (KPIs) related to delivery performance, returns processing, and warehouse and potentially financial performance and regulatory compliance. Immediate action is being taken to restore service, and we are implementing workaround procedures. This report details the impact, current status, and planned actions.

1. Current Situation & Impact Assessment:

As you know, the LE-DEL module is the backbone of our delivery and returns processes. The outage is manifesting as follows:

Delivery Disruption: Outbound shipments are stalled. New delivery orders cannot be processed through SAP. Existing shipments may be impacted as pick/pack/ship confirmations are unavailable.

Returns Blockage: Customer returns cannot be properly processed, leading to potential inventory inaccuracies and delays in restocking.

Warehouse Inefficiency: Warehouse staff are operating with limited real-time delivery information, increasing the risk of errors and slowing down operations. Picking and

Management:**

4. **Internal Communication:** Notify all relevant departments (Logistics, Warehouse, Sales, Customer Service) about the server outage, its potential impact on their operations, and any available workarounds.
5. **Customer Communication:** If customer deliveries are significantly affected, start informing them of potential delays. Be transparent about the situation and provide estimated resolution times.

**Mitigation and Contingency Planning:**
6. **Manual Workarounds:** Develop and implement temporary manual processes for critical functions like order processing and shipment tracking to minimize disruptions.
7. **Priority Management:** Identify the most critical business functions reliant on LE-DEL and prioritize their restoration.
8. **Resource Allocation:** Ensure adequate resources (personnel, tools, systems) are allocated to address the situation

* **Manual Workarounds:** While working towards recovery, establish clear manual workarounds for essential processes like order processing, shipment tracking, and return handling. Document these procedures thoroughly to ensure consistency and minimize errors.

**Post-Outage Recovery:**

* **Root Cause Analysis:** Once the server is restored, conduct a thorough root cause analysis to identify the specific cause of the outage. This will help prevent similar incidents in the future.
* **System Health Checks:** Perform comprehensive system health checks after restoring operations to ensure all components are functioning correctly and data integrity has not been compromised.
* **Performance Monitoring:** Implement robust performance monitoring tools for LE-DEL and other critical systems. These tools can alert you to potential issues before they cause outages.

**Long-Term Solutions:**

2. **Communication is Key:** Inform all relevant departments – Logistics, Warehouse Operations, Customer Service, Sales – about the outage and its potential impact. Explain the situation transparently and provide estimated restoration times.

3. **Customer Communication Plan:** Develop a clear communication plan for customers. We should proactively notify them of potential delays in deliveries or returns due to the system downtime. Offer alternative contact methods and provide updates on the situation regularly.
4. **Manual Workarounds:** Train relevant teams on temporary manual processes for handling deliveries, returns, and order tracking until the system is fully restored. Document these workarounds clearly to minimize errors and confusion.

**Post-Restoration Actions:**

1. **System Health Check:** Once server-003 is back online, thoroughly test the LE-DEL module and all related functionalities. Verify data integrity and

(logistics, warehouse, production, customer service)
* Management
* Key customers (especially those with imminent deliveries)

**Mitigation & Contingency Planning:**

* **Failover Systems:** Evaluate the existing failover mechanism for LE-DEL. Does it effectively redirect workloads to a backup server? Are regular tests conducted to ensure its functionality? If not, prioritize implementing robust failover solutions.
* **Manual Workarounds:** Develop clear and documented manual workarounds for essential processes while the system is down. This might involve using spreadsheets for tracking shipments, manually updating inventory records, or employing alternative communication channels with customers.
* **Training & Resources:** Ensure relevant personnel are adequately trained in these manual workarounds. Provide them with necessary resources (spreadsheets, templates) and support to minimize errors and operational disruption.

| | | | | | |
|---|---|---|---|---|---|
| throughput will not be updated in real time. This means reports and insights will be delayed, potentially hindering decision-making. Business Intelligence Impact: Any analytics built on the LE-DEL module's data (e.g., delivery efficiency, return rate analysis) will be incomplete until the server is back online, making it harder to forecast or improve operations. ERP Integration: Supply Chain Disruptions: Since SAP's LE-DEL module integrates with other parts of the SAP ERP (e.g., materials management, sales order processing), other processes might be affected as well. For instance, sales teams may be unable to check inventory levels or update order statuses until the server is restored. Financial Impact: Lost Revenue: Shipping delays can result in missed sales opportunities or dissatisfied customers who may cancel orders. There could also be financial penalties for late deliveries, especially in B2B or service-level agreements (SLAs). Cost of Workarounds: Manual processes or temporary solutions may incur additional costs, both in terms of resources and time. Compliance and Legal: Regulatory Compliance: In some industries, logistics and returns processing may be subject to regulatory requirements. If deliveries | packing are being significantly hampered. Customer Service Strain: Customer service representatives are unable to provide accurate order/return status updates, leading to increased call volumes and potential customer dissatisfaction. Data Integrity Concerns: Lack of real-time updates to inventory and delivery data is creating a risk of inaccurate reporting and potentially flawed business decisions. ERP Integration Issues: Downstream impacts are being observed in Sales Order Processing and Materials Management as information flow is disrupted. Potential Financial Impact: We anticipate potential lost revenue due to shipping delays, potential penalties for late deliveries (especially with our key B2B clients), and increased costs associated with manual workarounds. Compliance Risk: Delayed or inaccurate returns processing could potentially lead to compliance issues, | effectively. **Post-Outage Review:** 9. **Lessons Learned:** Conduct a post-mortem analysis to identify the root cause of the outage, assess the effectiveness of our response, and implement preventative measures to avoid similar incidents in the future. 10. **System Improvements:** Review our disaster recovery plan, failover mechanisms, and system monitoring capabilities to identify areas for improvement and enhance resilience. **Additional Considerations:** * **Business Impact Analysis (BIA):** Regularly update the BIA to reflect the criticality of the LE-DEL module and its dependencies on other systems. * **Training and Awareness:** Conduct regular training sessions to ensure all staff understand their roles and responsibilities in case of system outages. By taking these steps, | * **High Availability Configuration:** Explore implementing high availability (HA) solutions for LE-DEL to minimize downtime in future scenarios. This might involve using clustering technologies or redundant servers. * **Disaster Recovery Plan:** Develop a comprehensive disaster recovery plan specifically for your SAP environment, outlining procedures for system backups, data restoration, and business continuity during major outages. * **Capacity Planning:** Regularly review resource allocation and capacity planning for your SAP systems to ensure they can handle peak workloads and future growth demands. **Additional Considerations for Chemical Manufacturing:** * **Inventory Control:** The chemical industry often deals with hazardous materials. The LE-DEL outage may impact inventory control, potentially leading to safety hazards if processes are not managed correctly. * **Regulatory Compliance:** Chemical manufacturing is heavily | ensure smooth operation across all modules and integrations. 2. **Performance Monitoring:** Closely monitor system performance after restoration to identify any lingering issues or bottlenecks that might have arisen due to the outage. 3. **Root Cause Analysis:** Conduct a detailed investigation into the cause of the server outage. Document the findings, implement preventive measures to avoid similar incidents in the future, and update disaster recovery procedures accordingly. This multifaceted approach will help us minimize disruption, maintain customer confidence, and ensure a swift return to normal operations. Remember, proactive communication and efficient crisis management are crucial during such situations. | **Long-Term Improvements:** * **Root Cause Analysis:** After restoring server-003, conduct a thorough root cause analysis to identify the reason for the outage. This will help prevent similar incidents in the future. * **System Monitoring & Alerts:** Implement comprehensive system monitoring tools that proactively detect potential issues with server-003 or the LE-DEL module. Configure alerts to notify relevant personnel in case of anomalies. * **Disaster Recovery Plan:** Develop a detailed disaster recovery plan specifically for SAP's LE-DEL module, outlining procedures for data backups, system restoration, and communication strategies in the event of a major outage. **Remember:** This situation highlights the critical importance of robust IT infrastructure, reliable failover mechanisms, and well-defined contingency plans for mission-critical systems like SAP LE-DEL. By taking |
| and returns cannot be tracked in the system, this could lead to compliance issues, especially if goods are returned incorrectly or if there are discrepancies in inventory reporting. Steps to Address the Situation: Immediate Recovery: Investigate the root cause of the server being offline (hardware failure, software crash, network issue) and restore functionality as soon as possible. Failover Mechanism: If available, consider switching to a backup server or system to minimize downtime. Ensure failover procedures are tested for scenarios like this. Communication with Stakeholders: Notify affected teams (e.g., logistics, warehouse, customer service) about the disruption so they can manage customer expectations and work on manual processes if necessary. Monitor for Further Impact: Track the performance of the system after bringing the server back online to ensure the LE-DEL module is functioning as expected, and catch any residual issues that may arise. | depending on the specific chemical and regulatory requirements for tracking.<br><br>2. Immediate Actions Taken (Last 30 minutes):<br><br>Alert Verification: Confirmed the outage through SAP Solution Manager monitoring alerts and direct system access attempts.<br><br>Initial Server Check: Performed a remote ping and basic connectivity check on server-003. No response.<br><br>IT Infrastructure Team Notification: Immediately escalated the issue to the IT Infrastructure team for investigation and restoration. They are currently on-site investigating.<br><br>Stakeholder Notification (Initial): Sent a preliminary notification to the key stakeholders listed above, alerting them to the outage and potential impact.<br><br>Workaround Activation: Initiated the documented manual workaround process (see Appendix A - Workaround Procedure Document). This involves using spreadsheets to track urgent shipments and | we can minimize the impact of this server outage on our business operations and ensure a swift recovery. | regulated. Ensure your manual workarounds and recovery procedures comply with relevant environmental, health, and safety regulations.<br><br>By taking these steps, you can minimize the disruption caused by server outages and ensure your chemical manufacturing company's operations remain efficient and compliant. | | proactive measures, you can minimize the impact of future outages and ensure business continuity. |

| | returns. This is a temporary solution and prone to errors. | | | | |
|---|---|---|---|---|---|
| | 3. Investigation & Root Cause Analysis (Ongoing - IT Infrastructure Team Lead): | | | | |
| | The IT Infrastructure team is currently investigating the following potential causes: | | | | |
| | Hardware Failure: Checking server hardware components (CPU, memory, disk). | | | | |
| | Software Crash: Analyzing SAP system logs for errors and crashes. | | | | |
| | Network Connectivity: Verifying network connectivity to the server. | | | | |
| | Database Issues: Checking the underlying database (likely HANA) for errors or performance issues. | | | | |
| | Recent Changes: Reviewing any recent system changes or updates that may have contributed to the outage. | | | | |
| | 4. Recovery Plan & Timeline (Estimated): | | | | |
| | Phase 1: Restoration | | | | |

| | | | | |
|---|---|---|---|---|
| | (Target: Within 2-4 hours): The IT Infrastructure team is prioritizing restoring server-003 to operational status. This may involve restarting the server, restoring from a backup, or replacing faulty hardware.<br><br>Phase 2: Verification (Target: 1 hour after restoration): Once the server is back online, I will perform thorough testing of the LE-DEL module to ensure all functionalities are working as expected. This includes:<br><br>Creating and processing outbound deliveries.<br><br>Processing inbound returns.<br><br>Verifying inventory updates.<br><br>Confirming integration with other SAP modules.<br><br>Phase 3: Data Reconciliation (Target: 4-8 hours after verification): We will need to reconcile any data discrepancies that may have occurred during the outage, particularly related to inventory and deliveries. This will involve comparing manual records with SAP data. | | | |

| | | | | | |
|---|---|---|---|---|---|
| | | 5. Communication Plan:<br><br>Hourly Updates: I will provide hourly updates to the stakeholders listed above on the progress of the restoration and verification efforts.<br><br>Escalation: Any significant delays or complications will be immediately escalated to the IT Manager and relevant business leaders.<br><br>Post-Incident Report: A detailed post-incident report will be prepared outlining the root cause of the outage, the actions taken, and recommendations for preventing future occurrences.<br><br>6. Preventative Measures (Post-Incident):<br><br>To prevent similar outages in the future, we will:<br><br>Review Failover Configuration: Ensure the failover mechanism for the LE-DEL module is properly configured and tested. We need to confirm if a hot/warm standby server is available and if failover procedures are documented and | | | | |

| | | practiced. | | | | |
| | | | | | | |
| | | Strengthen Monitoring: Enhance SAP Solution Manager monitoring to provide earlier warnings of potential issues. Specifically, we need to monitor key LE-DEL specific metrics. | | | | |
| | | | | | | |
| | | Capacity Planning: Review server capacity to ensure it can handle current and future workloads. | | | | |
| | | | | | | |
| | | Change Management Review: Reinforce change management procedures to minimize the risk of disruptions caused by system changes. | | | | |
| | | | | | | |
| | | Disaster Recovery Plan Review: Review and update the overall disaster recovery plan to include specific procedures for restoring the LE-DEL module. | | | | |
| | | | | | | |
| | | Appendix A: Workaround Procedure Document (Link to Document) | | | | |
| | | | | | | |
| | | My next steps: | | | | |
| | | | | | | |
| | | Continuously monitor the IT Infrastructure team's progress. | | | | |
| | | | | | | |
| | | Prepare for thorough testing of the LE-DEL module once the server is restored. | | | | |
| | | Coordinate with the Logistics and Warehouse teams to manage the manual workaround process.<br><br>Begin drafting the post-incident report template.<br><br>I understand the severity of this situation and am committed to restoring service as quickly and efficiently as possible. Please let me know if you have any questions or require further information. | | | | |

| | Test 8 | | | | | |
|---|---|---|---|---|---|---|
| Act as a SAP Monitoring Expert at a Chemical Manufacturing Company. Given the new information and the constraints, the impact of server-003 being offline becomes even more complex. Here's a detailed analysis:<br><br>Shipment Delays and Backlog: High-Volume Shipment Loss: Dispatching Bay 0017 ships 8 containers every 5 minutes, which translates to 96 containers per hour. Since the bay has been down for 3 hours, 288 containers have not been processed. This creates a significant backlog of shipments, with each delayed shipment potentially valued at $10,000. Revenue Impact: The inability to dispatch high-value shipments directly leads to $2.88 million in lost revenue (288 containers × $10,000 per container). Further Delays in Backlog: The longer server-003 remains offline, the larger the backlog, and the delay in processing these high-value shipments continues to accumulate, resulting in compounded revenue loss. Regulatory Constraints and Compliance Risk: Chemical Product Constraints: The regulatory requirement states that no more than 10 containers of chemical products can be in warehouses 0024 and 0025 that feed Dispatching Bay 0017. With the bay down, | | SAP Monitoring Expert - Server-003 Outage - Chemical Manufacturing - Critical Incident Response<br><br>Situation Assessment: This is a critical, escalating incident with significant financial, regulatory, and reputational risks. The offline LE-DEL module on server-003 is not just a technical issue; it's a business-stopping event. The cascading effects are severe and require immediate, coordinated action. We're looking at a potential loss of $2.88 million and growing, potential regulatory penalties, and damage to key customer and supplier relationships.<br><br>My Role & Focus: As the SAP Monitoring Expert, my immediate focus is on:<br><br>Accelerating Server Restoration: Providing IT with all necessary SAP-specific information to expedite the recovery of server-003 and the LE-DEL module.<br><br>Real-time Impact Monitoring: Establishing a clear, real-time view of the key metrics driving the impact (warehouse container counts, backlog size, potential | You've outlined a truly dire situation! The downtime of server-003 is having a cascading effect on nearly every aspect of this chemical manufacturing company's operations.<br><br>As a SAP Monitoring Expert, I would immediately prioritize the following actions:<br><br>**1. Crisis Management and Communication:**<br><br>* **Assemble a crisis management team:** This should include representatives from IT, Operations, Logistics, Compliance, Sales, and Finance to coordinate response efforts effectively.<br>* **Activate communication protocols:** Inform all stakeholders (customers, suppliers, partners, regulators) about the situation, its impact, and the steps being taken to resolve it. Transparency is crucial to maintain trust and manage expectations.<br><br>**2. Server Restoration and System Recovery:**<br><br>* **Prioritize server-003 restoration:** Work tirelessly with the IT team to identify the root cause of the outage and expedite the recovery | This is a very detailed analysis of the impact of server-003 being offline, highlighting the interconnected nature of operations within a chemical manufacturing company.<br><br>Your points are well taken and accurately reflect the cascading effects of this downtime.<br><br>**Here's how I would prioritize immediate actions as a SAP Monitoring Expert:**<br><br>1. **Restore Server-003 ASAP:** This is the absolute top priority. Work with the IT team to identify the root cause of the outage and expedite the recovery process. Continuously monitor progress and escalate if necessary.<br>2. **Implement Emergency Manual Workarounds:**<br>* **Shipment Tracking & Dispatching:** Train a small team to manually track shipments, update dispatch records, and prioritize containers based on urgency and compliance (especially for chemical products). This will buy time while the system is down.<br>* **Warehouse Management:** Implement temporary controls to monitor container counts in | This is a critical situation! As the SAP Monitoring Expert, I agree with your assessment of the impact. Here's a breakdown of my immediate priorities and how we can address this:<br><br>**Priority 1: Server Restoration - The Critical Lifeline**<br><br>* **Immediate Action:** Work with the IT team 24/7 to identify the cause of server-003's outage and expedite its restoration.<br>* **Communication:** Keep all stakeholders informed about progress, estimated time to resolution, and potential impact mitigation strategies. Transparency is key during a crisis.<br>* **Resource Allocation:** If necessary, prioritize IT resources and manpower to resolve this issue as quickly as possible.<br><br>**Priority 2: Minimize Backlog and Regulatory Risk**<br><br>* **Manual Shipment Processing:** Train warehouse staff on manual shipment procedures to process urgent containers and | This is a comprehensive analysis of the situation! As a SAP Monitoring Expert at this Chemical Manufacturing Company, I completely agree with your assessment of the severity and far-reaching impact of server-003 being offline.<br><br>Here's how I would prioritize our immediate actions, building upon your excellent points:<br><br>**Priority 1: Server Restoration & LE-DEL Module Recovery**<br><br>* **Direct IT:** Work closely with the IT team to understand the root cause of the outage and expedite the server restoration process. This is our top priority.<br>* **Resource Allocation:** Ensure dedicated resources are assigned to this critical task, potentially even bringing in additional expertise if needed.<br>* **Status Updates:** Demand frequent status updates from the IT team and keep all stakeholders informed about progress.<br><br>**Priority 2: Mitigating Immediate Damage & Compliance Risks**<br><br>* **Manual Shipment Handling:** Immediately |

these containers may be accumulating in the warehouses, potentially violating the constraint. Compliance Risks: If the containers exceed the regulatory limit, the business could face penalties, fines, or even legal action. This is especially critical for regulated products like chemicals, where non-compliance can lead to more severe legal consequences. Warehouse Space Management: Overflow in Warehouse 0024 and 0025: As the Dispatching Bay cannot process shipments, containers from warehouses 0024 and 0025 will continue to pile up, reaching the 10-container limit quickly. This could force the warehouse to either: Stop accepting new containers for Dispatching Bay 0017, causing delays in inbound inventory. Reroute or store inventory elsewhere, which could incur additional costs and logistical challenges. Risk of Storage Saturation: With limited warehouse space, there may be no room for incoming shipments or additional returns, leading to further disruption in the supply chain. Operational Strain and Manual Intervention: Manual Processing: To mitigate delays, warehouse teams may need to manually handle the backlog of shipments. This increases labor costs and may lead to revenue loss).

Data Integrity Assessment: Planning for a thorough data reconciliation once the system is back online to address potential inventory discrepancies.

Proactive Communication: Providing clear, concise updates to stakeholders on the technical status and business impact.

Detailed Breakdown of Impact & Mitigation - Prioritized Actions:

I. Immediate (0-4 Hours) - Containment & Critical Workarounds

Server Restoration - Highest Priority:

IT Collaboration: I'm actively engaging with the IT team. I need to know:

Root Cause Analysis (RCA) Progress: What caused the outage? Understanding this is crucial to prevent recurrence.

Estimated Time to Recovery (ETR): A firm ETR is vital for planning. I'll push for the most aggressive, realistic timeline.

Data Integrity Checks:

process. Every minute lost compounds the damage.
* **Backup & Recovery Procedures:** Ensure that robust backup and recovery procedures are in place for the LE-DEL module and other critical systems. This will help minimize data loss and downtime in future incidents.

**3. Damage Control and Mitigation:**

* **Manual Shipment Handling:** Implement temporary manual processes to process shipments, focusing on high-priority or time-sensitive orders. Train staff and ensure clear communication to avoid errors.
* **Compliance Monitoring & Reporting:** Develop a system for manually tracking chemical product containers in warehouses 0024 and 0025 to ensure regulatory compliance. Report any potential violations immediately to relevant authorities.
* **Warehouse Space Management:** Explore temporary storage solutions (if available) or implement strict inventory controls to prevent overflow in warehouses 0024 and 0025.

warehouses 0024 and 0025. Consider using physical markers or manual logs to track inventory until the LE-DEL module is restored.

3. **Activate Crisis Communication Plan:**

* **Customers:** Inform high-value customers about the delay, potential impact on their deliveries, and estimated resolution time. Offer alternative solutions (if possible) and apologize for any inconvenience.
* **Suppliers & Partners:** Maintain transparency with suppliers and partners about potential downstream delays caused by shipment disruptions. Work collaboratively to find solutions that minimize disruption in the supply chain.

4. **Continuously Monitor and Adapt:**

* **System Status:** Keep a close eye on server-003 recovery progress and any updates from the IT team.
* **Inventory Levels:** Regularly review warehouse stock levels, especially for chemical products, to ensure

prevent further accumulation in warehouses 0024 and 0025.
* **Compliance Monitoring:** Implement a system (even temporary) to track the number of chemical containers in warehouses 0024 and 0025 manually. Ensure strict adherence to the 10-container limit.

**Priority 3: Temporary Workarounds and Communication**

* **Alternative Dispatch Bays:** Explore using other dispatch bays to process shipments, even if at a reduced capacity, to alleviate pressure on Bay 0017.
* **Customer Communication:** Notify customers impacted by delays as soon as possible, explain the situation, and provide revised delivery estimates. Offer alternative solutions or expedited shipping options where feasible.
* **Supplier Communication:** Inform suppliers of potential delays and work collaboratively to find alternative solutions or adjust order schedules.

**Priority 4: Long-Term Solutions and Lessons

implement a manual process for shipment processing at Dispatching Bay 0017. This will help reduce the backlog and prevent further regulatory non-compliance. Train warehouse staff on this process quickly.
* **Compliance Monitoring:** Conduct a rigorous audit of containers in warehouses 0024 and 0025 to identify any potential violations. Implement temporary manual tracking or workarounds as needed to ensure compliance with the 10-container limit.
* **Temporary Rerouting:** Explore alternative dispatch bays, if available, to reroute shipments and alleviate pressure on Dispatching Bay 0017.

**Priority 3: Communication & Transparency**

* **Customer Notification:** Proactively inform customers about the delay, emphasizing the steps being taken to resolve the issue and minimize impact. Be transparent about potential delivery schedule changes.
* **Supplier/Partner Communication:** Keep

inefficiencies or errors. High Stress on Logistics Teams: Given the high volume of shipments and the pressure to meet regulatory compliance, staff will face immense pressure to clear the backlog and avoid any fines, leading to potential human errors or suboptimal decision-making.
Impact on Inventory and Stock Visibility:
Discrepancies in Inventory Data: Since the LE-DEL module is offline, real-time data related to stock levels, shipments, and returns will not be updated. This can cause mismatches between physical stock and the system's inventory records.
Stock Depletion Risk: If shipments cannot be dispatched, inventory levels may appear to be higher than they actually are, leading to issues with forecasting and reordering stock.
Customer Impact: Delayed Deliveries: High-value customers who expect timely deliveries of their products will experience delays, which could damage relationships and potentially lead to lost customers. Possible Cancellations: Some customers, particularly those with urgent needs, may cancel orders, further exacerbating the financial loss and customer dissatisfaction.
Supply Chain and Vendor Impact: Downstream Delays: If shipments are delayed,

What checks are being performed before bringing the LE-DEL module back online to ensure data consistency?

SAP Specific Support: I'm prepared to provide SAP-specific expertise to the IT team, including:

Assistance with database recovery procedures.

Verification of SAP application server status.

Monitoring SAP logs for errors during the recovery process.

Warehouse Compliance - Critical:

Manual Container Count: Immediately initiate a physical count of chemical product containers in warehouses 0024 and 0025. This is independent of SAP data. We need a definitive number.

Stop Inbound to 0024/0025 (If Necessary): If the physical count is approaching the 10-container limit, immediately halt inbound shipments destined for these warehouses. This is a

**4. Long-Term Solutions and Preventative Measures:**

* **System Redundancy & Failover:** Invest in redundant systems and failover mechanisms for critical SAP modules like LE-DEL to minimize the impact of future outages.
* **Disaster Recovery Plan (DRP):** Develop a comprehensive DRP that includes testing, training, and clear procedures for restoring operations in case of major disruptions.
* **Continuous Monitoring & Alerting:** Implement robust monitoring tools and alerting systems to proactively identify potential issues with SAP systems and infrastructure.

**5. Financial Assessment and Insurance Review:**

* **Quantify the financial impact:** Analyze lost revenue, increased operational costs, potential fines, and other damages caused by the server downtime.
* **Review insurance coverage:** Assess whether existing policies cover losses resulting from system outages and take steps to

compliance with regulatory limits.
* **Operational Impact:** Analyze the effectiveness of manual workarounds and identify any new challenges or bottlenecks that require immediate attention.

**Longer-Term Actions (Once Server is Restored):**

* **Root Cause Analysis:** Conduct a thorough investigation into the server outage to prevent recurrence. This may involve reviewing system logs, infrastructure configurations, and potential human error.
* **Disaster Recovery Plan Review & Enhancement:** Ensure your disaster recovery plan is robust enough to handle similar outages in the future. Test and refine procedures regularly.
* **System Monitoring & Alerting:** Implement more sophisticated monitoring tools and alert systems to proactively identify potential issues before they escalate into major disruptions.

Remember, communication and transparency are key

Learned**

* **Server Backup & Recovery Plan:** Once the immediate crisis is resolved, review and strengthen the server backup and recovery plan to prevent similar outages in the future.
* **LE-DEL Module Redundancy:** Consider implementing redundancy for critical modules like LE-DEL to ensure business continuity in case of system failures.
* **Capacity Planning:** Analyze current warehouse capacity and potentially explore expansion options or alternative storage solutions to accommodate increased volume and potential disruptions.

**Remember:** This situation requires swift action, clear communication, and a focus on mitigating the most immediate risks. By following these steps, we can minimize the impact of this server outage and work towards restoring normal operations as quickly as possible.

suppliers and partners informed about the situation, highlighting the importance of their continued cooperation during this challenging time.

**Priority 4: Addressing Long-Term Impacts**

* **Root Cause Analysis:** Once server-003 is restored, conduct a thorough root cause analysis to identify the factors leading to the outage and implement preventative measures to avoid future occurrences.
* **Business Continuity Planning:** Review and strengthen existing business continuity plans to ensure a more robust response to similar disruptions in the future. This should include backup systems, redundant processes, and clear escalation procedures.

**Monitoring & Reporting:**
Continuously monitor the situation, tracking progress on restoration efforts, backlog reduction, compliance status, and any emerging issues. Provide regular reports to senior management and stakeholders, highlighting key metrics

| | | | | |
|---|---|---|---|---|
| suppliers and vendors further downstream may not receive the products they depend on, leading to potential production delays or bottlenecks. Loss of Trust in Supplier Relationships: Regular delays in high-value shipments could strain relationships with external partners, especially if contractual agreements include delivery deadlines. Increased Operational Costs: Overtime Costs: The warehouse and logistics teams will likely need to work overtime to address the backlog, which increases labor costs. Storage Fees: If containers cannot be processed, additional storage fees for items in warehouse 0024 and 0025 will add to operational costs. Long-Term Financial and Reputational Risks: Brand Damage: Ongoing disruptions in high-value shipments can harm the company's reputation, especially if customers or partners experience consistent delays. Financial Strain: The combined effect of lost revenue, compliance fines, and increased operational costs will strain the company's finances, affecting quarterly or annual profitability. Key Immediate Actions to Mitigate Impact: Restore Server-003: Immediate restoration of server-003 is paramount. Work with the IT team to ensure quick recovery of the LE-DEL module to prevent | hard stop to avoid regulatory violations.

Document Everything: Meticulously document all manual counts, decisions, and actions taken. This is vital for audit trails.

Manual Shipment Processing - High Priority:

Prioritize High-Value Containers: Focus manual processing efforts on the 288 containers already backlogged. Prioritize those with the highest value and shortest delivery deadlines.

Dedicated Team: Assign a dedicated team to manual processing, ensuring they have clear instructions and access to necessary documentation (shipping manifests, customer orders).

Data Capture: Develop a simple, standardized form to capture key shipment details during manual processing. This will be crucial for reconciliation when SAP is back online.

Stakeholder Communication - Essential:

Initial Notification: Send | strengthen coverage if necessary.

This situation demands swift and decisive action. By prioritizing server restoration, implementing immediate damage control measures, and developing long-term preventative strategies, this chemical manufacturing company can minimize the impact of this crisis and emerge stronger. | throughout this process. Keep stakeholders informed about the situation, the steps being taken, and any anticipated impacts. | and action items.

This situation demands swift and decisive action. By prioritizing these steps, we can minimize the damage, maintain stakeholder trust, and ensure a rapid return to normal operations. |

| | | | | |
|---|---|---|---|---|
| further disruption. Manual Shipment Handling: In parallel, consider implementing manual processes for shipment processing to reduce the backlog and keep the flow of goods moving. Compliance Monitoring: Closely monitor the number of chemical product containers in warehouses 0024 and 0025 to avoid exceeding the regulatory limit. Temporary manual tracking or workarounds may be necessary. Temporary Rerouting: Consider rerouting shipments through alternative dispatch bays (if available) to prevent additional backlog and regulatory non-compliance. Stakeholder Communication: Notify customers, suppliers, and partners about the delay and any potential regulatory risks to maintain transparency and manage expectations. | a brief, factual notification to key stakeholders (Logistics Manager, Supply Chain Director, Sales Director, Customer Service Manager) outlining the outage, the impact, and the immediate actions being taken.<br><br>Regular Updates: Establish a cadence for regular updates (every hour initially) to keep stakeholders informed.<br><br>II. Short-Term (4-24 Hours) - Stabilization & Damage Control<br><br>Alternative Dispatch Bay Utilization:<br><br>Capacity Assessment: Evaluate the capacity of other dispatch bays. Can they absorb some of the volume from Bay 0017 without causing bottlenecks elsewhere?<br><br>Routing Rules: If possible, temporarily adjust SAP routing rules (once IT confirms it's safe to do so) to divert shipments to alternative bays. Caution: This needs careful planning to avoid creating new issues.<br><br>Inventory Reconciliation Planning:<br><br>Data Freeze: Once the LE-DEL module is back | | | |

| | | online, immediately freeze all inventory transactions for a short period to allow for a clean reconciliation.<br><br>Reconciliation Plan: Develop a detailed plan for reconciling physical inventory counts with SAP data. This will involve comparing manual shipment records with SAP records and identifying discrepancies.<br><br>Customer Communication - Proactive:<br><br>Targeted Notifications: Identify customers with delayed shipments and proactively notify them of the situation. Provide realistic estimated delivery dates.<br><br>Transparency: Be honest about the cause of the delay and the steps being taken to resolve it.<br><br>Supplier Communication:<br><br>Inform Key Suppliers: Notify key suppliers of potential downstream delays. Work with them to mitigate any impact on their production schedules.<br><br>III. Long-Term (24+ Hours) - Prevention & | | | | |
|---|---|---|---|---|---|
| | | Improvement<br><br>Root Cause Analysis (RCA) Review: Thoroughly review the RCA conducted by IT. Identify any SAP-specific factors that contributed to the outage.<br><br>Monitoring Enhancements:<br><br>Proactive Monitoring: Implement more proactive monitoring of the LE-DEL module and server-003. This should include alerts for performance degradation, disk space utilization, and other critical metrics.<br><br>Automated Alerts: Configure automated alerts to notify the SAP team immediately of any issues.<br><br>Disaster Recovery Plan Review: Review and update the disaster recovery plan to ensure it adequately addresses the risk of an LE-DEL module outage.<br><br>Business Continuity Planning: Explore options for improving business continuity, such as redundant systems or alternative processing methods.<br><br>Tools & Resources I'll | | | |

| | | Utilize: | | | | |
|---|---|---|---|---|---|---|
| | | SAP Solution Manager: For monitoring system health, performance, and alerts. | | | | |
| | | SAP EarlyWatch Alert: To identify potential issues and vulnerabilities. | | | | |
| | | Database Monitoring Tools: To monitor database performance and availability. | | | | |
| | | Communication Platforms: Email, instant messaging, and conference calls for stakeholder communication. | | | | |
| | | Documentation: SAP system documentation, disaster recovery plan, and business continuity plan. | | | | |
| | | Key Success Metrics: | | | | |
| | | Server Restoration Time: Minimize downtime. | | | | |
| | | Backlog Reduction Rate: Track the rate at which the shipment backlog is being reduced. | | | | |
| | | Regulatory Compliance: Maintain compliance with chemical product storage regulations. | | | | |
| | | Customer Satisfaction: Minimize customer | | | | |

| | complaints and cancellations.<br><br>Financial Impact: Minimize lost revenue and increased costs.<br><br>This is a dynamic situation. I will continuously reassess the impact and adjust the mitigation plan as needed. My priority is to restore normal operations as quickly and safely as possible while minimizing the damage to the business.<br><br>I am ready to begin executing this plan immediately. Please let me know what information or support I need from you. | | | | |
|---|---|---|---|---|---|

# Appendix F: Further Reading

1. "Despite the substantial progress in learning and inference, there remains a relative sparseness of research focused on explainability and trustworthiness." *[2501.05435] Neuro-Symbolic AI in 2024: A Systematic Review*.

2. "what we should care about when it comes to A.I. in the real world is not explanation. It is validation." *Software vendors are pushing explainable A.I., but those explanations may be meaningless | Fortune*

   a. "...[W]e advocate for rigorous internal and external validation of AI models as a more direct means of achieving the goals often associated with explainability".
   Ghassemi et al. *The false hope of current approaches to explainable artificial intelligence in health care*. The Lancet Digital Health, Volume 3, Issue 11, e745 - e750

   b. "neural and symbolic approaches to AI complement each other with respect to their strengths and weaknesses. For example, deep learning systems are trainable from raw data and are robust against outliers or errors in the base data, while symbolic systems are brittle with respect to outliers and data errors, and are far less trainable. Symbolic systems, on the other hand, can make explicit use of expert knowledge, and are to a high extent self-explanatory, as their algorithms can be inspected and understood in detail by a human, while neural learning systems cannot readily take advantage of available coded expert knowledge, and are black boxes that make understanding their decision making processes very hard" (2022) Hitzler et al., *Neuro-symbolic approaches in artificial intelligence*, National Science Review, Volume 9, Issue 6, June 2022, nwac035, https://doi.org/10.1093/nsr/nwac035

3. (2024) Zhang & Sheng. *Neuro-Symbolic AI: Explainability, Challenges, and Future Trends*

4. (2005) Rosanas & Velilla. *The Ethics of Management Control Systems: Developing Technical and Moral Values | Journal of Business Ethics*. Springer

5. (2021) Chitnis et al. *[2105.14074] Learning Neuro-Symbolic Relational Transition Models for Bilevel Planning*

6. (2025) Colelough & Regli. *[2501.05435] Neuro-Symbolic AI in 2024: A Systematic Review*

7. (2022) Hamilton et al. *Is neuro-symbolic AI meeting its promises in natural language processing? A structured review*. Semantic Web: – Interoperability, Usability, Applicability. IOS Press.

8. (2019) Kannengiesser. *Design thinking, fast and slow: A framework for Kahneman's dual-system theory in design*. CUP.

9. *Code of Practice for the Cyber Security of AI - GOV.UK*

10. *National AI Strategy - GOV.UK*